\documentclass{article}
\usepackage[eandd, preprint]{neurips_2026}
\usepackage[utf8]{inputenc} 
\usepackage[T1]{fontenc}    
\usepackage{hyperref}       
\usepackage{url}            
\usepackage{booktabs}     
\usepackage{amsfonts}       
\usepackage{nicefrac}     
\usepackage{microtype}    
\usepackage{graphicx}
\usepackage{siunitx} 
\usepackage{graphicx} 
\usepackage{titletoc}
\usepackage{xcolor}
\definecolor{darkgreen}{rgb}{0,0.5,0}
\definecolor{mnotecolor}{rgb}{.4,0,.6}
\usepackage{array}
\usepackage{amsmath}
\usepackage[normalem]{ulem}
\usepackage{hyperref}
\usepackage{caption}
\usepackage{booktabs,xcolor,colortbl,makecell}
\hypersetup{
    colorlinks=true,
    linkcolor=blue,
    filecolor=magenta,      
    urlcolor=cyan,
    pdftitle={Overleaf Example},
    pdfpagemode=FullScreen,
    }
\usepackage{natbib}
\usepackage[normalem]{ulem}
\usepackage{listings}
\usepackage{subcaption}
\usepackage{algorithm}
\usepackage{algpseudocode}
\usepackage[T1]{fontenc}
\usepackage{inconsolata}

\usepackage{float}
\usepackage[edges]{forest}
\usepackage{array}
\usepackage{comment}
\usepackage{soul}

\usepackage[table]{xcolor}
\usepackage{colortbl}

\newcommand{\mnote}[1]{}

\renewcommand{\mnote}[1]{\textcolor{mnotecolor}{\textbf{[MU: #1]}}}

\usepackage{graphicx}
\usepackage[table]{xcolor}
\usepackage{booktabs}
\usepackage{array}
\usepackage{placeins}

\title{\includegraphics[height=1.3em]{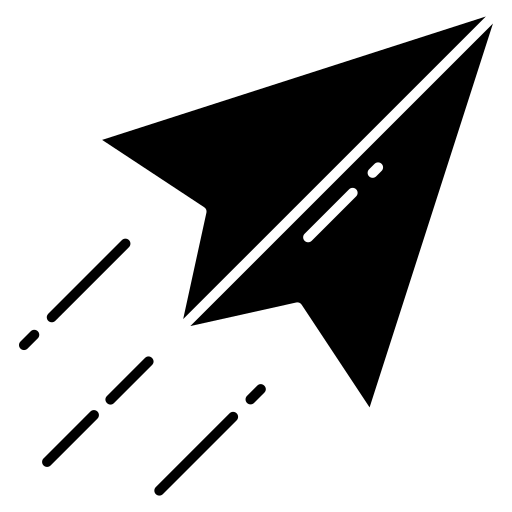} ShapeBench: A Scalable Benchmark and Diagnostic Suite for Standardized Evaluation in Aerodynamic Shape Optimization}

\author{
  Shaghayegh Fazliani\footnotemark[1]\kern0.45em\footnotemark[3] \\
  \texttt{fazliani@stanford.edu} \\
  \And
  Krissh Chawla\footnotemark[1]\kern0.45em\footnotemark[2]\kern0.45em\footnotemark[3] \\
  \texttt{krissh@stanford.edu} \\
  \And
  Jack Guo\footnotemark[2]\kern0.45em\footnotemark[3] \\
  \texttt{jack@spinozalabs.ai} \\
  \AND
  Yiren Shen\footnotemark[1]\kern0.45em\footnotemark[3] \\
  \texttt{yrshen@stanford.edu} \\
  \And
  Matthias Ihme\footnotemark[1] \\
  \texttt{mihme@stanford.edu} \\
  \And
  Madeleine Udell\footnotemark[1] \\
  \texttt{udell@stanford.edu} \\
}

\begin{document}

\maketitle
\begingroup
\renewcommand{\thefootnote}{\fnsymbol{footnote}}
\footnotetext[1]{Stanford University.}
\footnotetext[2]{Spinoza Labs.}
\footnotetext[3]{Equal Contribution.}
\endgroup

\vspace{-2 em}
\begin{center}
\small
\setlength{\parskip}{0pt}
\textbf{ShapeBench:}\quad \url{https://github.com/ShapeBench/ShapeBench}\\[0.15em]
\textbf{Dataset:}\quad \url{https://huggingface.co/datasets/ShapeBench}
\end{center}
\vspace{2em}

\begin{abstract}
    Rapid progress in aerodynamic shape optimization (ASO) has outpaced currently-available standardized evaluation frameworks. Fair comparison  requires a unified benchmark spanning diverse shape classes, objective formulations, and matched-budget state-of-the-art baselines. We introduce \textbf{ShapeBench}, an open-source ASO benchmark with a unified API spanning 103 tasks across eight shape categories and multiple optimization regimes. Each ShapeBench task includes a validated surrogate for fast search; when feasible, a high-fidelity Computational Fluid Dynamics (CFD) pipeline for final verification is available, enabling systematic fidelity-gap analysis. ShapeBench provides a reproducible protocol with well-configured baselines to compare fairly using a consistent budget metric, allowing for comparison among both classical and LLM-driven methods, including general-purpose optimizers and a new domain-specialized evolutionary LLM baseline, \textbf{ShapeEvolve}.
    Results on ShapeBench demonstrate substantial variance in optimizer rankings across shape categories and problem formulations, with mean pairwise Spearman $\rho = 0.013$, 
    so single-task conclusions do not reliably generalize across problem classes. The benchmark is also far from saturation; classical methods are rarely applicable across all shape categories and tasks, further highlighting the need for more general-purpose approaches.
\end{abstract}

\section{Introduction}

Aerodynamic shape optimization (ASO) is a core problem in aerospace and automotive design, with direct impact on efficiency, emissions, and performance~\citep{jameson1988aerodynamic,drela1998pros}. Recent progress has expanded the optimizer set used for ASO, including gradient-based methods~\citep{wachter2006implementation,kingma2014adam,loshchilov2017decoupled}, evolutionary algorithms~\citep{hansen2001completely,salimans2017evolution,zhou2024evolutionary}, Bayesian optimization~\citep{giannakoglou2002design,balandat2020botorch}, and Large Language Model (LLM)-guided approaches~\citep{openevolve,lange2025shinkaevolveopenendedsampleefficientprogram}. Current ASO benchmarks and evaluation setups are often narrow (e.g. two-dimensional) and lack diversity (e.g. cover a single problem setup), which weakens fair cross-method comparison and hides how well algorithms generalize to new ASO setups. As optimizers become more capable and new Machine Learning (ML)-based methods emerge , ASO needs broader and more challenging benchmark suites with shared evaluation protocols to measure generalization, expose failure modes, and push for progress beyond narrow two-dimensional settings.

\textbf{Task diversity.} Aerodynamic design problems vary considerably in geometry --- 2D airfoils ~\citep{sharpe2025neuralfoilairfoilaerodynamicsanalysis, aerosandbox_phd_thesis}, 3D wings ~\citep{shen2025aso, yang2025superwingcomprehensivetransonicwing}, blended-wing bodies ~\citep{Sung_2025}, full aircraft configurations ~\citep{Martins2013MultidisciplinaryDO}, automotive external aerodynamics ~\citep{qiu2025drivaerstar, elrefaie2025drivaernetlargescalemultimodalcar, ashton2024drivaerml} , etc. --- and must perform well across a range of flight conditions (multi-point mission robustness), across several metrics (multi-objective trade-offs), and with mixed-variable design spaces ~\citep{Saves_2022}. 
Conclusions drawn from isolated case studies often fail to generalize~\citep{wolpert2002no}.
A comprehensive benchmark should include a collection of heterogeneous tasks spanning a spectrum of possible design needs. 

\textbf{ASO needs surrogates, but robust benchmarking is essential.} Balancing the computational cost and analysis fidelity is critical for ASO automation and convergence as running high fidelity analyses, e.g. Computational Fluid Dynamics (CFD), in an ASO loop is prohibitively expensive. Surrogate models are instead used to build a reduced-complexity model to approximate CFD solvers. They speed up ASO but raise three serious challenges for reliable evaluation: (1) they introduce approximation error and potential optimizer bias~\citep{queipo2005surrogate, forrester2008engineering}; (2) surrogate exploitation: optimization can push designs outside the surrogate’s valid region, yielding unstable predictions near the edge of the training distribution~\citep{forrester2008engineering}; (3) infrastructure fragmentation: surrogates are released with model-specific interfaces, data conventions, and output formats ---  this creates substantial integration overhead and weakens reproducibility across groups~\citep{pineau2020improvingreproducibilitymachinelearning}, discouraging broad cross-task evaluation.

\textbf{LLM-based evaluation gap.} General-purpose LLM-optimization frameworks are increasingly used in scientific fields~\citep{lewkowycz2022solvingquantitativereasoningproblems, romera2024mathematical, abhyankar2025accelerating}, yet ASO has limited standardized evidence on how such methods compare against established optimizers across diverse aerodynamic tasks~\citep{zhang2024usinglargelanguagemodels}.  Without a strong, carefully tuned domain-specific LLM reference baseline evaluated under the same protocols and compute budgets, claims about LLM advantages or limitations in ASO remain difficult to interpret.

Classical ASO suites such as the AIAA Aerodynamic Design Optimization Discussion Group (ADODG) cases~\citep{anderson2015aerodynamic, kenway2016aerodynamic, ledoux2015study} provide strong CFD-based benchmarks for airfoil and wing optimization. Recent resources like AFBench~\citep{liu2024afbench} focuse on large-scale airfoil design and generation, and engineering benchmarks such as EngiBench~\citep{zhou2025engibenchbenchmarkevaluatinglarge} include airfoil tasks through a shared API. These benchmarks do not simultaneously support (i) multi-geometry aerodynamic tasks, (ii) surrogate-to-CFD fidelity checks, (iii) matched-budget comparisons across optimizers, and (iv) validity diagnostics. ShapeBench addresses these needs with a unified, reproducible benchmark and diagnostic suite for classical and LLM-driven ASO methods.

\begin{figure}[h!]
    \centering
    \includegraphics[width=0.83\linewidth]{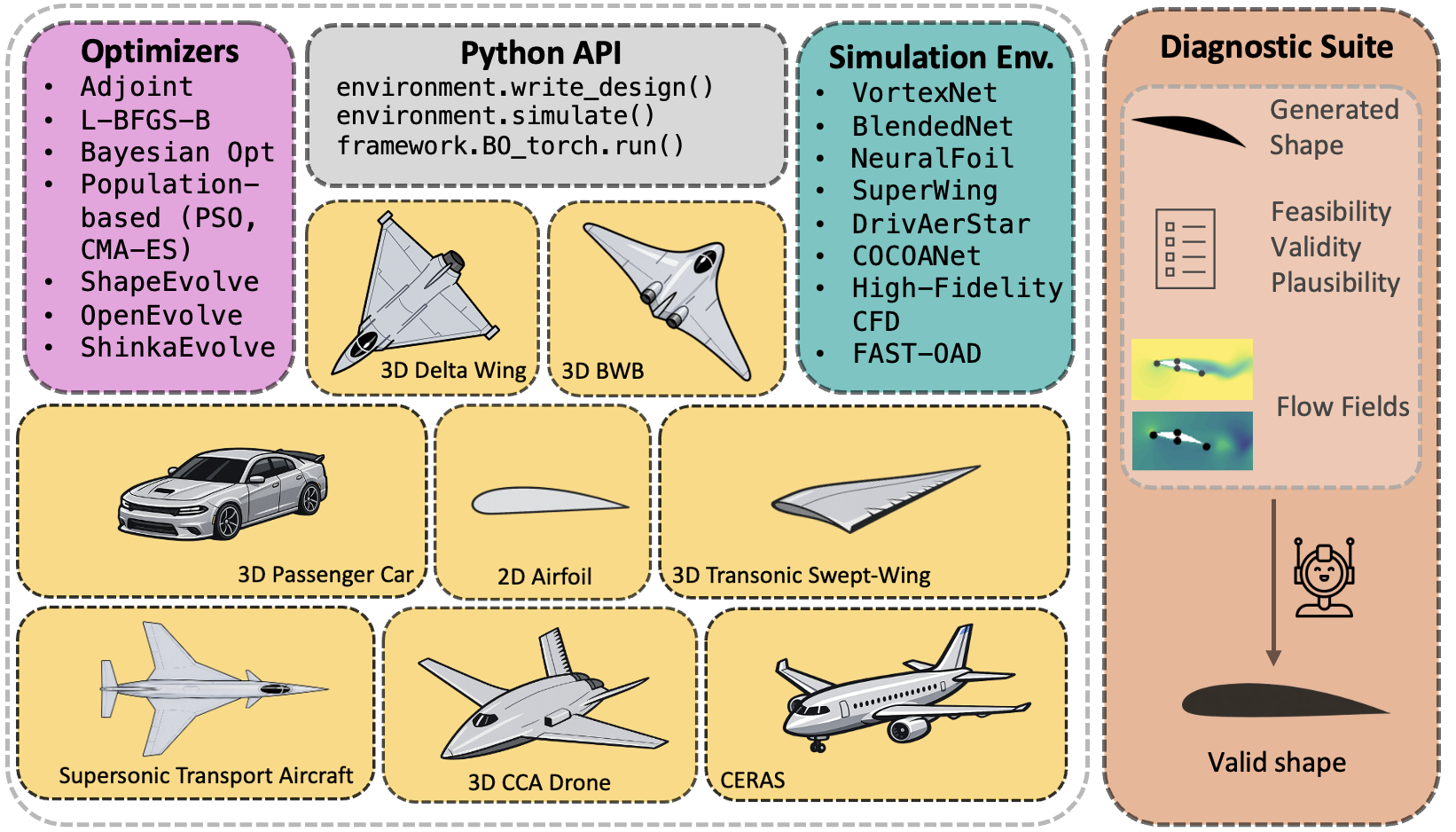}
    \caption{An overview of ShapeBench: covered geometries, optimizers, simulation environments, an example of the Python API, and the Diagnostic Suite }
    \label{shapebench overview}
\end{figure}

\textbf{Contributions.}
Key contributions of this paper are summarized as follows:
\begin{itemize}
\item \textbf{A cross-domain ASO evaluation suite.}
  We release \textbf{ShapeBench}: a collection of 103 instantiated ASO tasks spanning eight aerodynamic shape categories and multiple optimization regimes (single-/multi-objective, multi-point, and mixed-variable), enabling systematic cross-task comparison of optimizers.

\item \textbf{Empirical evidence that single-task experiments do not generalize.}
  Using ShapeBench, we document substantial instability of method rankings across shape categories and formulations: mean pairwise Spearman $\rho \approx 0.013$ across environment-level rankings.

\item \textbf{Paired surrogate solvers for each task.}  ShapeBench pairs each task with a fast, pretrained surrogate and a comparable high-fidelity where available. 
To supplement existing surrogates, we develop 
\texttt{COCOANet} for the Collaborative Combat Aircraft (CCA) tasks. COCOANet is based on Transolver~\citep{wu2024transolver}, a physics attention neural architecture that achieves high accuracy on irregular‑mesh PDE surrogate benchmarks and is trained on our open dataset of 3,570 high fidelity aerodynamic simulations. \texttt{COCOANet} and our CCA dataset cover a setting with limited public training data.

\item \textbf{Diagnostic suite.}
ShapeBench includes a \textbf{Diagnostic Suite} to flag high-reward but infeasible or implausible designs and identify other characteristics of surrogate exploitation, supporting valid interpretations and suggesting appropriate fixes.
  \item \textbf{Standardized and specialized baselines.}
We provide a reproducible evaluation protocol and baseline results for established classical optimizers (adjoint, L-BFGS-B, PSO, CMA-ES, Bayesian optimization) and general-purpose LLM optimizers (OpenEvolve, ShinkaEvolve).
We also introduce \textbf{ShapeEvolve}, 
our ASO-specialized LLM evolutionary method, which serves as a rigorous reference for emerging LLM-driven optimizers.
\end{itemize}

We release tasks, surrogate model and training data, baseline runs, evaluation scripts, generated designs, and a reference CCA design developed by \texttt{nTop}\footnote{\texttt{nTopology} (\texttt{nTop}) is a commercial implicit-modeling CAD platform used for lattice structures, lightweighting, and design-for-manufacturing workflows in aerospace and related industries.} to support reproducible ASO research.

\section{ASO Evaluation Landscape}

Let $x\in\mathcal{X}$ denote design parameters controlling the aerodynamic shape geometry, and let $u$ denote the flow state satisfying the governing equations
(residual) $\mathcal{R}(u;x)=0$. ASO solves
\begin{equation}
\label{eq:aso}
\begin{aligned}
\mbox{minimize} \quad & J(u,x) \\
\mbox{subject to} \quad & \mathcal{R}(u;x)=0, \\
& g_i(u,x)\le 0,\quad i=1,\ldots,I,\\
& h_j(u,x)=0,\quad j=1,\ldots,J,
\end{aligned}
\end{equation}
over $x \in \mathcal X$, where $J$ is a scalar aerodynamic performance objective, and
$\{g_i\},\{h_j\}$ encode engineering constraints (forces/moments, geometric feasibility,
trim, etc.). For multi-point missions, let $\{\omega_k\}_{k=1}^{K}$ denote operating
conditions (e.g. Mach number, Reynolds number, etc.) and let $u_k$ satisfy
$\mathcal{R}(u_k;x)=0$ for each $k$. Then objective $J$ scalarizes the per-point
objectives $\{J_k(u_k,x)\}_{k=1}^{K}$, for example, as the sum or maximum,
under the same constraint structure.

\textbf{Optimizers in ASO.} \textbf{Gradient-based} optimizers
exploit derivatives of a reduced objective to improve the design (i.e. minimizing $J(u,x)$ subject to $\mathcal{R}(u;x)=0$ using derivatives
of the reduced objective $\mathcal{J}(x):=J\bigl(u(x),x\bigr)$, where $u(x)$ solves
$\mathcal{R}\bigl(u(x);x\bigr)=0$).
Adjoint methods~\citep{giles2000introduction, jameson1988aerodynamic} compute $\nabla_x \mathcal{J}(x)$ with a cost that depends only weakly on the design dimension, working well when
differentiable forward models and consistent discrete linearizations are available. Quasi-Newton methods (e.g., L-BFGS-B~\citep{zhu1997algorithm}) use finite-difference gradient estimates.
In contrast, \textbf{gradient-free}
methods rely on simulated objectives/constraints, treating $\mathcal{J}(x)$ as a black-box function and constructing an internal probabilistic model of the objective. This class includes population-based search (e.g. evolutionary methods~\citep{hansen2001completely}, Particle Swarm Optimization (PSO)~\citep{kennedy1995particle}); direct pattern
search~\citep{hooke1961direct}; surrogate-acquisition methods such as Bayesian optimization~\citep{frazier2018tutorial}, and the emerging class of LLM-driven frameworks~\citep{lange2025shinkaevolveopenendedsampleefficientprogram, openevolve}. 

\textbf{Benchmarking in ASO.} The AIAA ADODG includes a widely used set of reference CFD optimization cases for two-dimensional transonic airfoil
problems and three-dimensional wing and wing--body configurations~\citep{anderson2015aerodynamic, kenway2016aerodynamic, ledoux2015study}. ADODG-style suites are valuable for gradient-based adjoint optimization pipelines, yet focus on high-fidelity CFD-driven optimization over a small set of canonical geometries limits their broader benchmarking utility. Recent ML resources have substantially improved data access and evaluation speed for ASO.
This includes airfoil dataets/benchmarks (e.g., AFBench~\citep{liu2024afbench}) and surrogate models~\citep{qiu2025drivaerstar, shen2025aso, yang2025superwingcomprehensivetransonicwing}. Recent broader benchmarks such as EngiBench~\citep{zhou2025engibenchbenchmarkevaluatinglarge}
provide general engineering problems and tools aimed at
evaluating LLMs on engineering reasoning and problem solving across many
subdomains, including airfoil design, which improves reproducibility for ``engineering-as-code'' workflows, but
they lack coverage on advanced problems studied in ASO.

\section{ShapeBench: A Unified Benchmark for ASO Evaluation}
This section describes ShapeBench's features, as indicated in Figure~\ref{shapebench overview}, in more detail.

\begin{figure}[h!]
    \centering
    \includegraphics[width=0.7\linewidth]{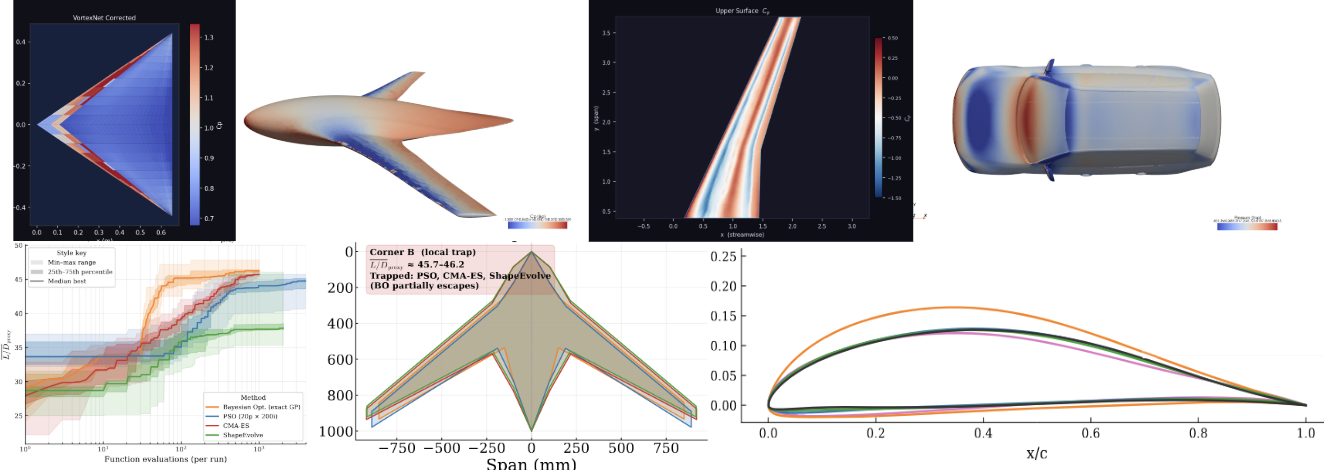}
    \caption{ShapeBench generates design- and run-level visuals (e.g., geometry/field plots and optimization trajectories) to support qualitative debugging and consistent cross-method analysis.}
    \label{fig: visuals}
\end{figure}

\textbf{Versioning, Reproducibility, Problem Metadata \& Visualizations.} ShapeBench versions problem implementations and associated assets (task definitions, surrogate checkpoints, and evaluation settings) and records every run with a fully resolved configuration file, so results remain traceable and reproducible as the benchmark evolves. Each task exposes structured metadata through a common interface: variable types and bounds, operating conditions, objectives, and constraints. This lets optimizers switch across tasks without task-specific parsing. Figure \ref{fig: visuals} demonstrates different visualizations supported by ShapeBench.

\textbf{Diagnostic Suite.}
To evaluate design credibility beyond reward values, diagnostics run as a two-stage pipeline: (i) deterministic evidence checks and (ii) an LLM integrative judge. The deterministic stage produces a tiered evidence bundle as shown in Table~\ref{tab:diagnostic_suite} (Appendix~\ref{app: diagnostic suite}). The LLM stage converts this bundle (plus design plots, flow field visualizations, and run context) into an auditable diagnostic report (example in Section \ref{analysis and insights}) with controlled failure-mechanism and  mitigation actions.

\subsection{Problem Environments \&  Simulation Pipelines}

ShapeBench problem environments are listed in Table \ref{problems list}. These problems reflect the diversity of aerodynamic shape categories, ranging from 2D airfoil to different types of 3D aircraft and automotive design. ShapeBench covers single- and multi-objective optimization settings as well as multi-point missions. In addition to continuous-variable ASO, two mixed-variable conceptual design problems, CEntral Reference
Aircraft System (CERAS) and Supersonic Transport
Aircraft (STA), are provided, which represent the design choices made in aircraft conceptual design (typically referred to as overall aircraft design (OAD)~\citep{saves2022bayesian, Saves_2022, saporito2023robust}). ASO relies on multi-fidelity and surrogate-based methods due to simulation cost. ShapeBench provides a range of simulation fidelity levels for these problem environments (FAST-OAD~\citep{david2021fast} for the mixed-variable conceptual design, fast surrogate models, and high-fidelity solver implementations). Details of each shape category, its importance in ASO, and its corresponding surrogate appear in Appendices  \ref{ShapeBench Task Setups and Additional Plots} and \ref{app: surrogate models} respectively. Key properties of each problem --- dimension, geometry parametrization, and simulation times --- are summarized in Table~\ref{problems list}.

\begin{table}[h!]
\centering
\caption{ASO Problems currently implemented in ShapeBench. All shape categories come with surrogates. For the mixed-variable problems, CERAS and STA, the designs are evaluated through FAST-OAD. High fidelity CFD simulation environments are implemented for the rest (except Passenger Car). Simulation times (Simu. Time) given per evaluation. CPU is a single core of an Intel Xeon Platinum 8275CL, 3.00 GHz.}
\vspace{0.5em}
\label{problems list}
\renewcommand{\arraystretch}{1.25}

\resizebox{\textwidth}{!}{
\begin{tabular}{>{\raggedright\arraybackslash}p{0.23\textwidth}
                >{\raggedright\arraybackslash}p{0.12\textwidth}
                >{\raggedright\arraybackslash}p{0.23\textwidth}
                >{\centering\arraybackslash}p{0.18\textwidth}
                >{\centering\arraybackslash}p{0.23\textwidth}
                >{\centering\arraybackslash}p{0.2\textwidth}}
\toprule
\shortstack[l]{\rule{0pt}{3.1ex}\textbf{ShapeBench}\\\textbf{Design Category}} &
\shortstack[l]{\rule{0pt}{3.1ex}\textbf{Shape}\\\textbf{Dimension}} &
\shortstack[l]{\rule{0pt}{3.1ex}\textbf{Geometry}\\\textbf{Parameterization}} &
\shortstack[c]{\rule{0pt}{3.1ex}\textbf{ShapeBench}\\\textbf{Surrogate}} &
\shortstack[c]{\rule{0pt}{3.1ex}\textbf{Simu. Time}\\\textbf{(Surrogate)}} &
\shortstack[c]{\rule{0pt}{3.1ex}\textbf{Simu. Time}\\\textbf{(CFD/Other)}} \\
\midrule
Delta Wing & 3D & 2 planform vars + \mbox{fixed root chord}  & \texttt{VortexNet}~\citep{shen2025vortexnet} & $0.1 \pm 0.01$ GPU-s & $ 450 \pm 25$ CPU-hr \\
\hline
Blended Wing Body (BWB) & 3D & 9 planform vars + \mbox{fixed airfoil section} & \texttt{BlendedNet}~\citep{Sung_2025} & $0.3 \pm 0.2$ CPU-s & $3 \pm 0.1$ GPU-h \\
\hline
Airfoil & 2D & 18 Kulfan/CST vars & \texttt{NeuralFoil}~\citep{sharpe2025neuralfoilairfoilaerodynamicsanalysis} & $1.2 \pm 0.1 $ CPU-ms& $508 \pm 83$ CPU-ms\\
\hline
Transonic \mbox{Swept-Wing} & 3D & 20 Baseline CST airfoil + \mbox{18 wing-level} vars & \texttt{SuperWing}~\citep{yang2025superwingcomprehensivetransonicwing} & $0.137 \pm 0.012$ CPU-s & $0.836 \pm 0.19$ GPU-h (H100)\\
\hline
Passenger Car & 3D & 20 FFD/CAD \mbox{deformation vars} & \texttt{DrivAerStar}~\citep{qiu2025drivaerstar}& $ 0.02 \pm 0.001$ CPU-h & $ 90 \pm 15$ CPU-h \citep{qiu2025drivaerstar}  \\ 
\hline
Collaborative Combat Aircraft (CCA) & 3D & 16 vars & \texttt{COCOANet} (Ours) [Appendix \ref{app: cca surrogate}] & $1.207 \pm .044$ CPU-s & $1.13 \pm 0.15$  GPU-h (H100)\\
\hline
CEntral Reference
Aircraft System (CERAS) & 3D & 10 mixed-type vars (6 cont, 2 discrete, 2 categorical) & \texttt{CERASNet} [Appendix \ref{app: CERAS Surrogate}] & $533.2 \pm 20.6$ CPU-$\mu$s & $88 \pm 15$ CPU-s\\
\hline
Supersonic Transport Aircraft (STA) & 3D & 9 mixed-type vars (6 cont, 3 categorical)& \texttt{STANet} [Appendix \ref{app: STA Surrogate}] & $69.3 \pm 29.2$ CPU-ms & $3.52 \pm 0.76$ CPU-s\\
\bottomrule
\end{tabular}
}
\end{table}

\subsection{Implemented Optimization Methods/Algorithms} \label{Implemented Algorithms}
 ShapeBench includes eight optimizers: adjoint can be used when the simulation is differentiable (here, the airfoil), while the remaining seven optimizers (Table \ref{tab:optimizers_dense}) are tested across all benchmark settings.
Further details on each method appear in Appendix~\ref{Optimization_Method_Descriptions_and_Implementation_Details}. Here, we provide a brief description on \textbf{ShapeEvolve}, our proposed LLM-driven optimizer specialized for ASO.

\begin{figure}[h!]
    \centering
\includegraphics[width=0.9\linewidth]{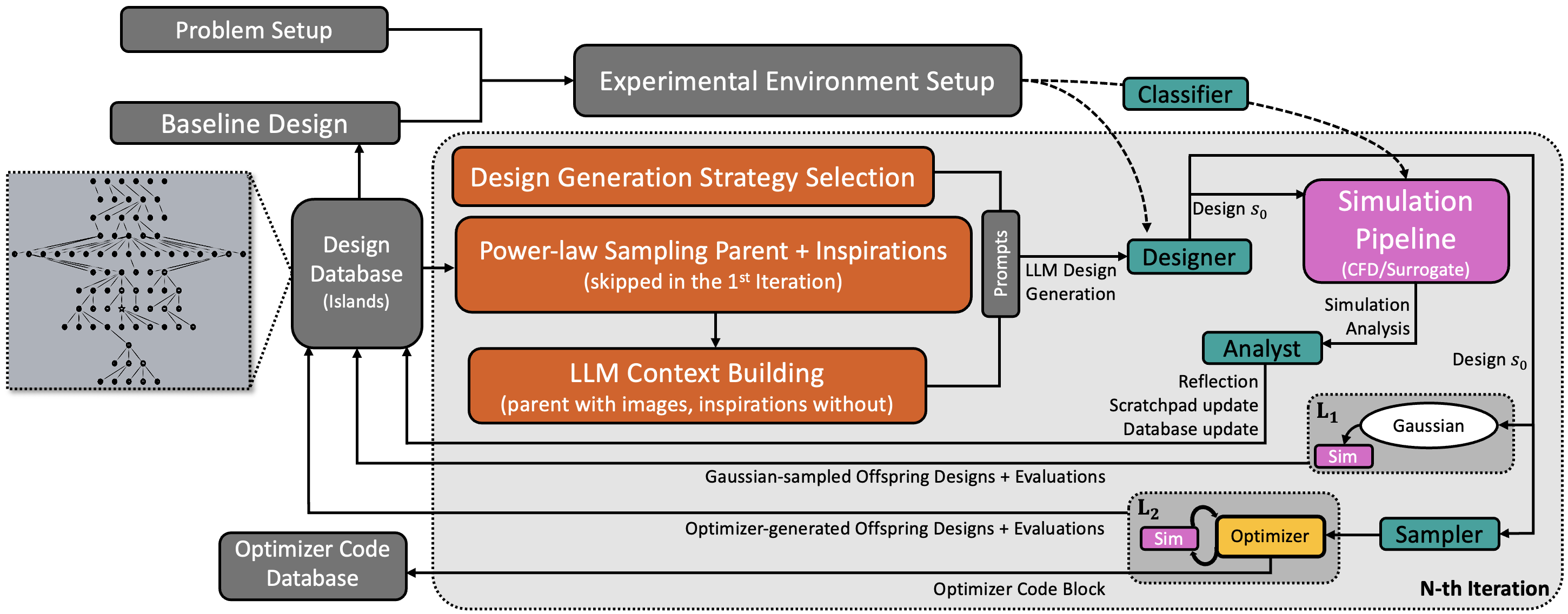}
    \caption{Overview of the ShapeEvolve pipeline.}
    \label{pipeline v3}
\end{figure}

\begin{table*}[h!]
\centering
\caption{ShapeBench provides optimizers with a shared interface and robust default parameters.}
\scriptsize
\setlength{\tabcolsep}{3.5pt}
\resizebox{\textwidth}{!}{
\begin{tabular}{lllc}
\toprule
\textbf{Optimizer} & \textbf{Category} & \textbf{Core Mechanism} & \textbf{Implementation Reference} \\
\midrule
\textbf{Adjoint (IPOPT)}
& Gradient-based
& Adjoint gradients
& \citep{giles2000introduction,jameson1988aerodynamic} \\

\textbf{L-BFGS-B}
& Gradient-based
& Finite-diff gradients
& \citep{zhu1997algorithm} \\

\textbf{BO}
& Surrogate-based
& GP + LogEI
& \citep{frazier2018tutorial} \\

\textbf{PSO}
& Derivative-free pop.
& Objective-only swarm
& \citep{kennedy1995particle} \\

\textbf{CMA-ES}
& Derivative-free pop.
& Objective-only adaptive Gaussian
& \citep{hansen2001completely} \\

\textbf{OpenEvolve}
& LLM evolutionary
& LLM propose/eval/refine (MAP-Elites + islands)
& \citep{openevolve} \\

\textbf{ShinkaEvolve}
& LLM evolutionary
& LLM propose/eval/refine (bandit-guided operators)
& \citep{lange2025shinkaevolveopenendedsampleefficientprogram} \\

\textbf{ShapeEvolve}
& LLM evolutionary 
& ASO prompts/flow feedback + LLM code for final design sampling
& Ours \\
\bottomrule
\end{tabular}
}
\label{tab:optimizers_dense}
\end{table*}
\textbf{ShapeEvolve} follows the same LLM propose-evaluate-refine loop as other evolutionary LLM frameworks, but uses different search-feedback strategies. Its ASO-specialization is in the optimization interface: ASO-structured prompts, aerodynamic state feedback, and geometry and flow-field visual contexts are injected at every iteration, with persistent reflection memory to accumulate design knowledge over time.
Figure~\ref{pipeline v3} summarizes ShapeEvolve's pipeline. Each outer iteration follows an island-based evolutionary loop with power-law parent selection. A \textbf{designer} LLM agent first proposes a center design, which is simulated and logged, and the resulting geometry/flow feedback is folded into the search context (including a persistent reflection scratchpad).
Another distinction from generic LLM optimizers is a second stage: a separate LLM (\textbf{sampler}) emits a Python optimizer script that does gradient-free local search, coordinate updates, or other adaptive strategies.
Successful sampler programs are tracked in a small code archive for reuse and mutation across iterations, improving search efficiency and coverage of the design space.

\section{Proof-Of-Concept Numerical Experiments}

In this section, we illustrate the breadth of experiments and analyses  that ShapeBench supports. 

\subsection{Cross-category Study \& Performance Variability} \label{main: cross-category}
\begin{figure}[h!]
    \centering

    \begin{subfigure}{0.55\textwidth}
        \centering
        \includegraphics[width=\linewidth]{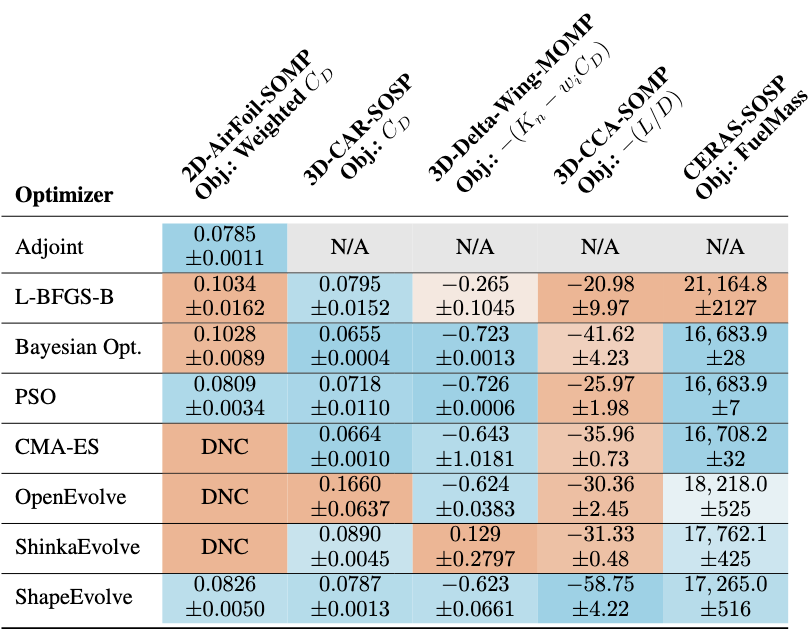}
        \caption{}
        \label{variance table}
    \end{subfigure}
    \hfill
    \begin{subfigure}{0.43\textwidth}
        \centering
        \includegraphics[width=\linewidth]{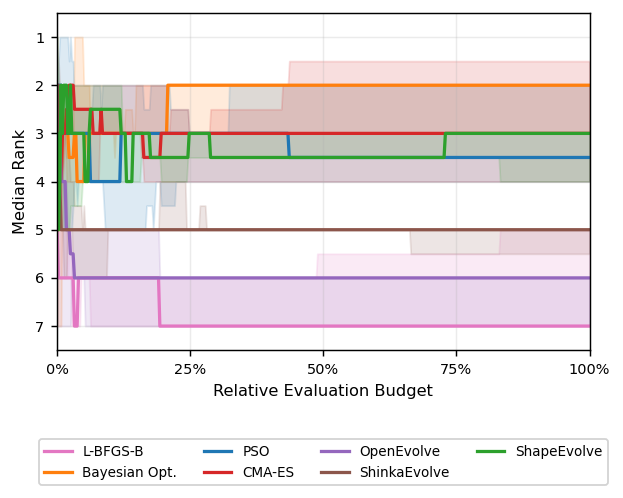}
        \caption{}
        \label{rank vs budget}
    \end{subfigure}

    \caption{(a) Final median objective values for each task per optimizer.  Colors are column-wise log-normalized per task to highlight within-task performance variability (lower = darker blue = better) Abbreviations: SO for single-objective, MO for multi-objective SP for single-point, MP for multi-point, DNC for ``Did Not Complete" (no feasible design was found), N/A for ``Not Applicable" for that specific combination of task and optimizer method. See Appendix \ref{ShapeBench Task Setups and Additional Plots} for details. (b) Median rank of each optimizer over relative evaluation budget. Rank 1 denotes the best-performing method and shaded regions indicate the IQR bands across environments. }
    \label{plot and table}
\end{figure}

 ShapeBench makes cross-category experiments easy to run by swapping \texttt{problem\_id} and optimizer name to explore different ASO categories and optimizers.
Results are directly comparable, as all methods share one evaluation interface and one budget protocol, 
New optimizers can be added once and benchmarked against all baselines. Figure~\ref{variance table} shows evaluation across diverse shape categories and task setups can surface optimizer behaviors that are not visible in single-task studies --- that is, performance is strongly task-dependent. For instance, Bayesian optimization and PSO perform best on the CERAS and delta-wing tasks, but rank among the weakest methods on the CCA task, while LLM-driven methods exhibit the opposite trend on the CCA environment. More broadly, substantial variation is observed in optimizer rankings across aerodynamic settings in the benchmark: conclusions from a single ASO tasks do not always generalize to other problem classes. Figure~\ref{rank vs budget} further shows that optimizer rankings are not only task-dependent, but also budget-dependent. The relative ordering of methods changes over the course of optimization. No method stays at rank 1 across the aggregate curve, and the wide interquartile bands indicate that each optimizer wins only on particular subsets of environments. ShapeBench exposes when different methods are competitive under a fixed evaluation budget.

\subsection{\texttt{COCOANet} \& the CCA Benchmark}
We include a 3D Collaborative Combat Aircraft (CCA) task to represent an ASO setting with higher-dimensional aircraft geometry and limited public data. In ShapeBench, the CCA vehicle is parameterized by $16$ design variables and its lift-to-drag ratio ($L/D$) is optimized under a prescribed flight mission (see Appendix~\ref{app: cca tasks}). \texttt{COCOANet} provides a $\sim 1{,}600\times$ speedup (up to constant overheads) over direct CFD evaluation (details in Appendix~\ref{app: cca surrogate}).  \texttt{COCOANet} significantly reduces the computational cost, allowing comparisons on the CCA setup with different optimizers (Figure~\ref{variance table}).

\begin{figure}[h!]
    \centering
    \includegraphics[width=1\linewidth]{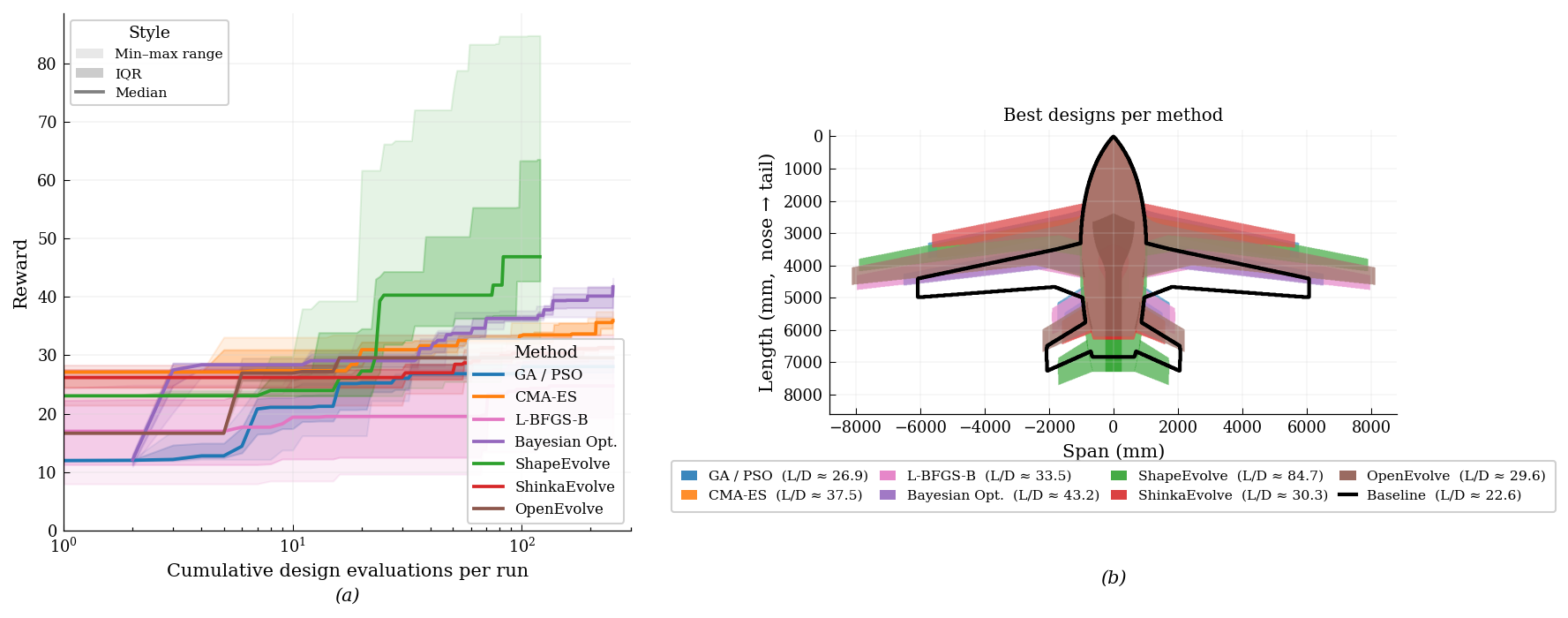}
    \caption{3D CCA experiment for $L/D$ objective. (a) Reward vs. evaluations (b) best shape per optimization method + \texttt{nTop} baseline design}
    \label{cca plot}
\end{figure}

Figure~\ref{cca plot} shows a representative CCA experiment and illustrates ShapeBench's cross-optimizer visualization tools. Panel~\ref{cca plot}(b) also compares the best designs from each optimizer against the reference geometry produced by \texttt{nTop}\cite{ntop2025}. Optimizers behave differently on this task than on many others from Figure~\ref{variance table}: ShapeEvolve converges to the best $L/D$, significantly outperforming classical methods. The source of this distinction, and its role in ShapeEvolve's performance (and the stronger performance of LLM-driven methods more broadly in this task), requires further study, but a potential explanation is the much larger design space of the CCA case (as discussed in Appendix~\ref{app: cca surrogate}), making ASO for this case particularly challenging. No one optimizer dominates the ASO tasks in ShapeBench: instead, observed variation in performance provides grist for improved methodologies. 

\subsection{Analysis \& Insights} \label{analysis and insights}

\begin{figure}[!htbp]
    \centering

    \begin{subfigure}{0.48\textwidth}
        \centering
        \includegraphics[width=\linewidth]{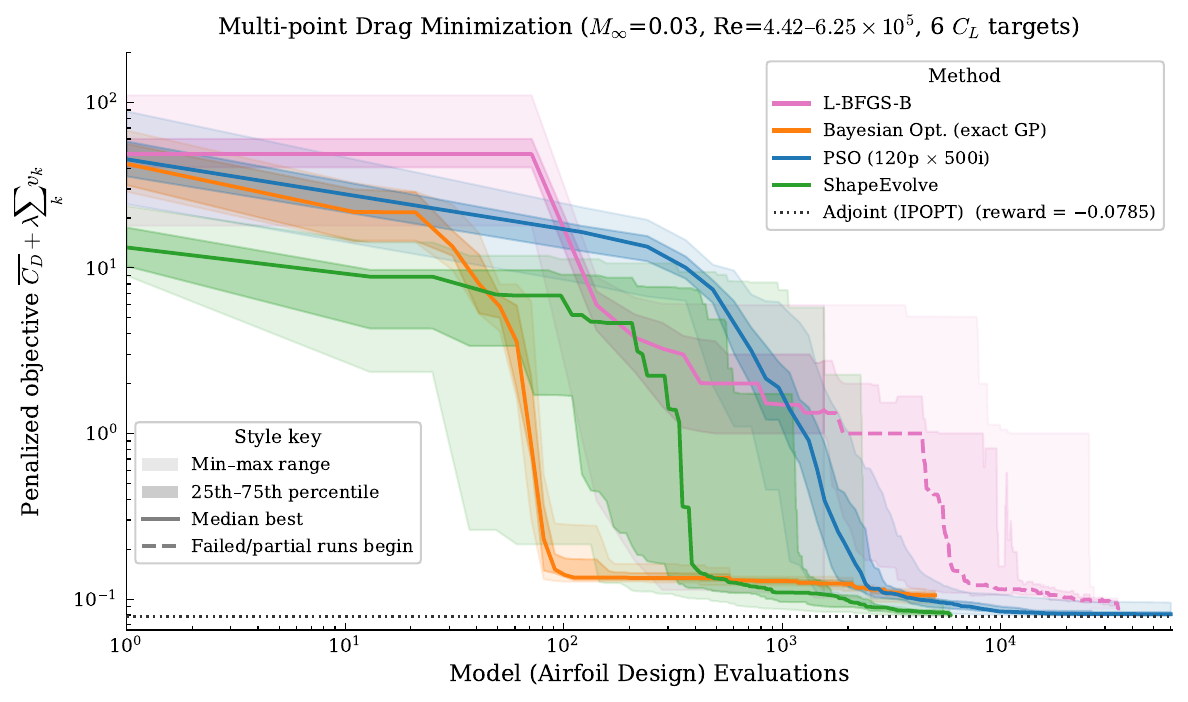}
        \caption{}
        \label{fig:neuralfoil-multipoint-objective-vs-iterations}
    \end{subfigure}
    \hfill
    \begin{subfigure}{0.48\textwidth}
        \centering
        \includegraphics[width=\linewidth]{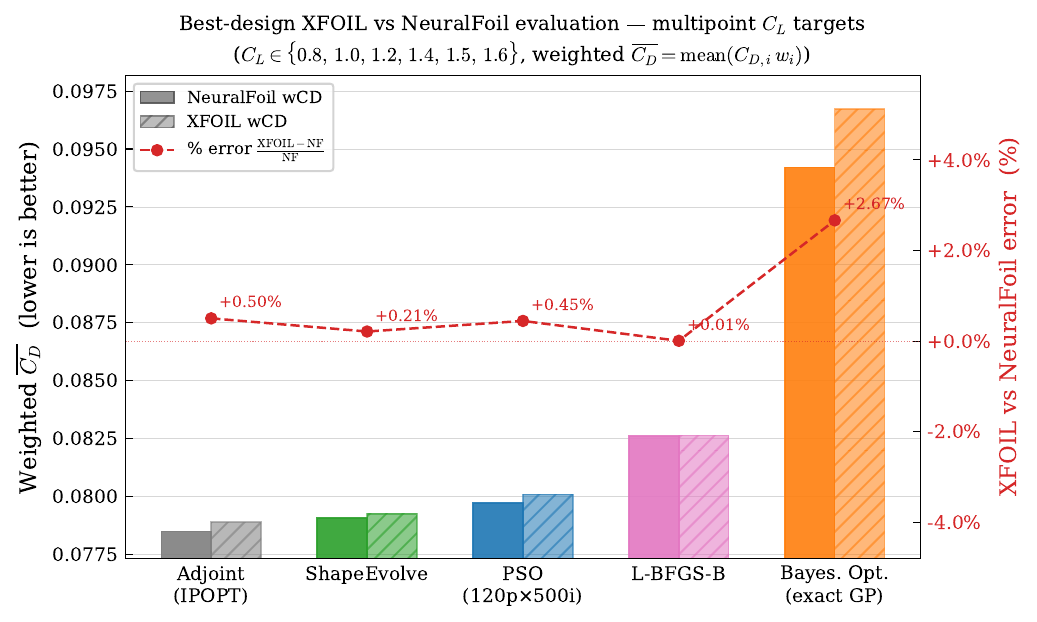}
        \caption{}
        \label{fig:neuralfoil-multipoint-xfoil-vs-nf-fidelity}
    \end{subfigure}

    \caption{(a) Convergence plot (objective vs. evaluations) plot for the 2D airfoil multi-point drag minimization task; Penalized objective $=$ weighted $\overline{C_D}$ + constraint penalty, with lower value corresponding to better performance. Note that within each method, all runs are included in the median (with each trace extended to the full evaluation budget). Failed runs (whose local minimum result in infeasible designs) are present for L-BFGS-B; their start is denoted with the dashed lines. (b) \texttt{NeuralFoil} and \texttt{XFOIL} evaluations of best design for each method, for the 2D airfoil multi-point drag minimization task. }
    \label{fig:neuralfoil-multipoint-results}
\end{figure}

\textbf{Fidelity gap.} The implemented high-fidelity environments are included in ShapeBench to help distinguish real  improvements from surrogate artifacts. Figure \ref{fig:neuralfoil-multipoint-results} shows an example of the fidelity gap analysis for the \texttt{NeuralFoil}. Relative to \texttt{XFOIL} (the airfoil high-fidelity environment), \texttt{NeuralFoil} reports mean $C_D$ (drag coefficient) errors as low as $0.37$\% on attached-flow cases and $2.0$\% on post-stall cases \citep{sharpe2025neuralfoilairfoilaerodynamicsanalysis}. This observation is consistent with our benchmark's findings; for the multi-point task, the error relative to \texttt{XFOIL} validation is at maximum $2.67$\% across the best designs of all methods. For the single point lift-to-drag maximization task, the best design $L/D$ fidelity gap is at $\sim 9$\%, with certain explored cases showing errors of up to $\sim 18$\%; the larger errors here can be rationalized by amplification of prediction errors in the $C_L/C_D$ ratio and by surrogate predictions being least reliable at the edge of the \texttt{NeuralFoil} confidence envelope.

\FloatBarrier

\begin{figure}[h!]
  \centering

  \begin{subfigure}{\linewidth}
    \centering
    \includegraphics[width=\linewidth]{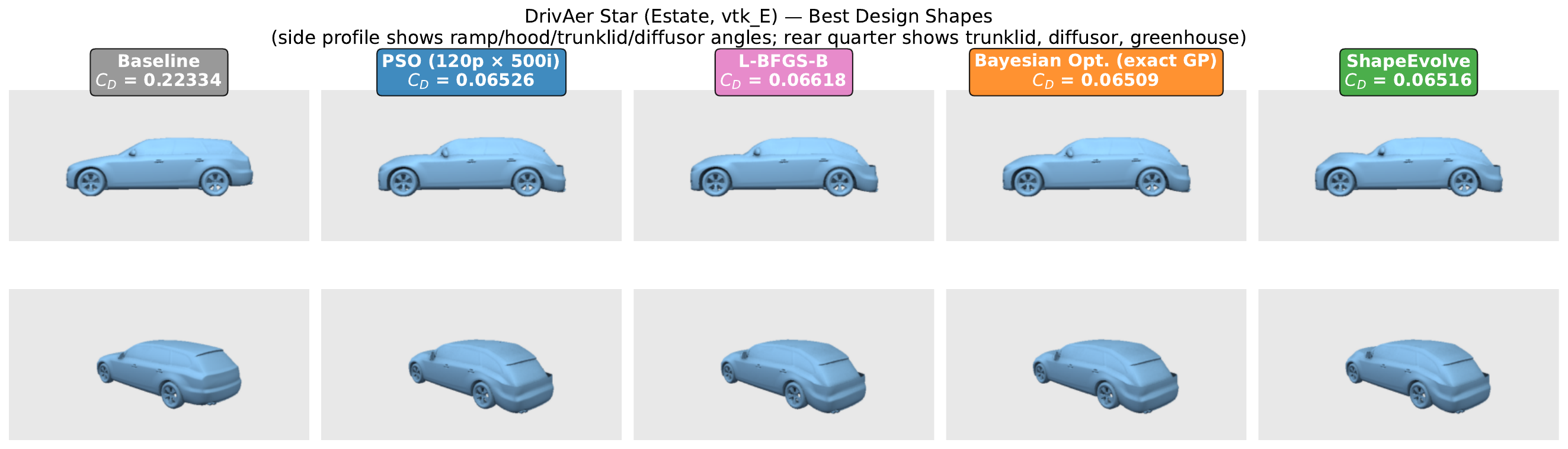}
    \caption{}
    \label{car design failure}
  \end{subfigure}

  \begin{subfigure}{\linewidth}
    \centering
    \includegraphics[width=\linewidth]{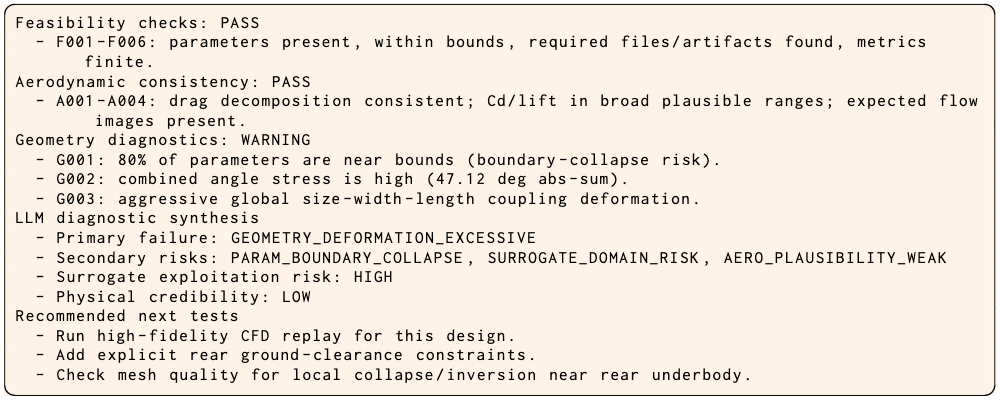}
    \caption{}
    \label{lst:drivaer_diag_highlights}
  \end{subfigure}

  \caption{(a) Best designs for the baseline and for each method (2D side views and 3D isometric views) for minimized $C_D$ task for 3D car design via DrivAerStar. (b) Summary of the diagnostic report for the \texttt{DrivAerStar} 3D passenger car case. }
  \label{fig:stacked-images}
\end{figure}

\textbf{Diagnostic Suite.} Figure \ref{car design failure} shows final optimized designs for the 3D car design problem minimizing the drag  coefficient $C_D$.  All the optimizers report very low $C_D$, but the resulting car geometries are visibly unrealistic (rear droop). Figure \ref{lst:drivaer_diag_highlights} summarizes the diagnostic report. In this case, the objective value is strong, but diagnostic evidence indicates likely surrogate exploitation and low physical credibility. This is exactly the kind of failure mode the Diagnostic Suite is meant to catch. It identifies likely mechanisms (boundary collapse and geometry over-deformation) and suggests concrete mitigation steps --- in particular, adding an explicit rear ground-clearance constraint directly targets the observed rear-droop failure mode and prevents optimizers from exploiting low-clearance geometries that are aerodynamically favorable but physically implausible.

\section{Use Cases, Limitations \& Future Work of ShapeBench}\label{Use Cases, Limitations, Future Work}

ShapeBench helps researchers choose optimizers, validate surrogate-derived gains, diagnose failure modes, and prioritize compute under realistic ASO constraints. Table \ref{tab:shapebench_usecases} summarizes use cases.

\begin{table}[h!]
\centering
\caption{Representative ShapeBench use cases and the benchmark signals each supports.}
\vspace{0.5em}
\label{tab:shapebench_usecases}
\footnotesize
\setlength{\tabcolsep}{4pt}
\renewcommand{\arraystretch}{1.05}
\resizebox{\textwidth}{!}{
\begin{tabular}{>{\raggedright\arraybackslash}p{0.18\textwidth}
                >{\raggedright\arraybackslash}p{0.20\textwidth}
                >{\raggedright\arraybackslash}p{0.28\textwidth}
                >{\raggedright\arraybackslash}p{0.34\textwidth}}
\toprule
\shortstack[l]{\rule{0pt}{2.5ex}\textbf{Use Case}} &
\shortstack[l]{\rule{0pt}{2.5ex}\textbf{Primary User}} &
\shortstack[l]{\rule{0pt}{2.5ex}\textbf{Key Question}} &
\shortstack[l]{\rule{0pt}{2.5ex}\textbf{ShapeBench Value}} \\
\midrule

\textbf{Cross-domain optimizer selection}
&
Method developers; applied ASO teams.
&
Which optimizer families remain competitive across heterogeneous aerodynamic tasks, rather than on one geometry/objective only?
&
Runs the same optimizers across multiple shape categories and formulations under one API, exposing task-dependent ranking shifts and reducing single-task overfitting of conclusions.
\\
\hline

\textbf{Surrogate-to-high-fidelity handoff}
&
Surrogate-driven design workflows.
&
Do surrogate-side gains persist under final high-fidelity verification on comparable metrics?
&
Uses a standardized surrogate-first workflow with paired high-fidelity verification hooks, enabling explicit fidelity-gap reporting.
\\
\hline

\textbf{Validity-aware ranking}
&
Benchmark authors; design reviewers.
&
Are top-scoring designs physically credible, or artifacts of proxy/surrogate exploitation?
&
Adds diagnostic evidence beyond scalar reward (feasibility, geometric risk, aerodynamic plausibility) in schema-valid outputs for interpretable best-design comparisons.
\\
\hline

\textbf{Classical vs.\ LLM optimizer comparison}
&
LLM-for-science and optimization researchers.
&
Are observed gains algorithmic, or caused by mismatched budgets, APIs, and protocols?
&
Enforces a common interface and budget protocol across classical and LLM optimizers, enabling fair head-to-head and reproducible comparison.
\\
\hline

\textbf{Performance--cost trade-offs}
&
Practitioners with fixed compute budgets.
&
Which method gives the best objective quality for a given evaluation/cost budget?
&
Tracks objective trajectories with standardized evaluation counts (and optional cost accounting), supporting deployment-oriented optimizer selection.
\\
\hline

\textbf{Community extension}
&
External contributors
&
How can new tasks be added without breaking comparability across releases?
&
Supports plugin-style task integration with standardized outputs/protocols, lowering integration effort while preserving benchmark semantics.
\\
\bottomrule
\end{tabular}
}
\end{table}

\paragraph{Limitations \& Future Work.} 
ShapeBench improves reproducible ASO evaluation, but several limitations remain. High-fidelity verification is available for many but not all settings. Notably, the 3D passenger car does not have this verification pipeline available. The $12{,}000$ CFD simulations were pre-computed under a restricted institutional STAR-CCM+ license and then distributed as static files. No CFD evaluator is available, and so optimizer-proposed designs must instead be evaluated by the \texttt{Transolver} surrogate. As another important limitation, surrogate-driven optimization is vulnerable to exploitation even when surrogates are well trained and rigorously validated: in our experiments, we observe failure modes such as unphysical reward behavior and boundary saturation, where optimizers push parameters to low-confidence extremes of the design space.

These effects can produce high objective values without corresponding physical credibility. The current benchmark emphasizes conceptual and surrogate-speed regimes; scaling to larger high-fidelity budgets and broader real-world constraints (manufacturability, robustness, and uncertainty-aware objectives) is ongoing work. Future releases will expand full method--task coverage, strengthen standardized surrogate-to-high-fidelity replay, and add stronger uncertainty and validity controls (including richer diagnostics and confidence-calibrated reporting) for more reliable optimizer ranking.

\section{Conclusion}\label{Conclusion}
We introduced \textbf{ShapeBench}, a unified benchmark and diagnostic suite for aerodynamic shape optimization, with 103 tasks spanning eight shape categories, carefully configured established baselines, and a common evaluation interface for both classical and LLM-driven optimizers. Our results show that optimizer rankings are unstable across tasks (mean pairwise Spearman $\rho=0.013$), so single-task conclusions do not generalize reliably. We also show that surrogate-based optimization needs explicit validity checks and, where available, high-fidelity replay, since high objective values can still correspond to physically weak designs. By releasing standardized tasks, surrogates, baselines, and diagnostics, we aim to make ASO evaluation more reproducible, comparable, and useful for both method development and deployment-oriented optimizer selection.

\newpage

\section{Acknowledgments and Disclosure of Funding}
We would like to thank Professor Juan Alonso from Stanford University and Wojciech Zaremba for helpful discussions regarding this work. We gratefully acknowledge support from the National Science Foundation (NSF) Awards IIS-2233762 and 2345740;
the Office of Naval Research (ONR) Awards N000142212825, N000142412306, N000142312203; NASA Award 80NSSC19K1661 under the Commercial Supersonics Technology (CST) program, Supersonic Configurations at Low Speeds (SCALOS), with Sarah Langston as the NASA technical grant monitor;
the Alfred P. Sloan Foundation; and IBM Research as a founding member of Stanford Institute for Human-centered Artificial Intelligence (HAI).

\bibliographystyle{plain}
\bibliography{references}

\newpage

\startcontents[appendices]
\section*{Appendix Table of Contents}
\printcontents[appendices]{}{0}{\normalsize}
\newpage

\appendix

\section{More Information on This Project}

\subsection{Licenses}

Both our codebases are released under the GPL-3.0 license. All used datasets/surrogates/validation tools are released under either the CC BY-NC-SA 4.0 license (\texttt{VortexNet}, \texttt{SuperWing}, \texttt{DrivAerStar}, CERAS), the MIT License (for \texttt{BlendedNet},  \texttt{NeuralFoil}, and the CCA dataset), GNU General Public License (GPL) (for \texttt{XFOIL}), commercial licensing (Flow360), or uses a personally-developed setup by the authors (Supersonic Transport Aircraft).

\subsection{Experimental Setup} 

\paragraph{Computational Resources Details.} \label{Computational Resources Details}
All cases were run using Intel Ice Lake (3rd-generation Xeon, AVX-512), with $128 \times$ computational cores per node with $256$ GB physical RAM. The CCA dataset was generated using GPU nodes that consist of $8 \times$ A100 SXM4 80GB GPUs connected via NVLink.

\subsection{Definitions, Notations \& Metrics}
\label{metrics}
\begin{table}[h!]
\centering
\resizebox{\textwidth}{!}{
\begin{tabular}{llll}
\hline
\textbf{Category} & \textbf{Metric} & \textbf{Formula} & \textbf{Interpretation} \\
\hline
Performance & $C_L$ & $2L/(qU^2c)$ & Lift capacity \\
Performance & $C_D$ & $2D/(qU^2c)$ & Total drag \\
Performance & $C_m$ & $2M/(qU^2c^2)$ & Pitch moment \\
Geometry & $t/c$ & $\max(y_u-y_\ell)/c$ & Thickness ratio \\
Geometry & Camber & $\max\, y_c$ & Camber magnitude \\
Flow & Pressure drag fraction & $C_{D_p}/C_D$ & Separation indicator \\
Performance & $C_x$ & $\dfrac{C_L}{C_D}-\lambda_1\lvert C_m\rvert+\lambda_2\,\mathbf{1}_{0.08\le t/c\le 0.24}$ & Example reward with multiple factors \\
\hline
\end{tabular}
}
\end{table}

\section{Additional Details on ShapeBench}

\subsection{Application Programming Interface} \label{app: api}

\begin{lstlisting}[language=Python, caption={ShapeBench API usage example: 3D BWB multipoint optimisation.}, label={lst:api_bwb}, basicstyle=\ttfamily\fontsize{7}{8}\selectfont]
import os, sys, csv, importlib, numpy as np
from types import SimpleNamespace

sys.path.insert(0, "/scratch/ShapeEvolve")
os.chdir("/scratch/ShapeEvolve")

# 1) Instantiate benchmark task
from environments.BlendedNet.environment import BlendedNetEnvironment
from environments.BlendedNet.rewards.shapebench_5_max_LD import ShapeBench5MaxLDReward

reward      = ShapeBench5MaxLDReward()                    # 5-point mission objective
environment = BlendedNetEnvironment(reward=reward)         # BlendedNet surrogate

# 2) Sample and evaluate one candidate design
lb, ub = environment.get_param_bounds()                   # per-parameter bounds
x0     = lb + np.random.default_rng(42).random(len(lb)) * (ub - lb)

case_dir    = "results/BlendedNet/example/iter_0000"
os.makedirs(case_dir, exist_ok=True)
design_path            = environment.write_design(x0, case_dir, "iter_0000")
reward_val, results    = environment.simulate(design_path, case_dir)
metrics                = results["metrics"]               # L/D, CD, CL, ...

# 3) Bayesian optimisation under standardised budget (3 000 evaluations)
fw_bo   = importlib.import_module("frameworks.BO_torch.run")
args_bo = SimpleNamespace(
    n_calls=3000, n_initial=30, num_restarts=10, raw_samples=256,
    train_cap=0, random_state=42, gradient_infeasible=True,
    warmstart_csv=None, save_fields=False, render_images=False,
)
out_bo = "results/BlendedNet/BO_torch/shapebench_5_max_LD/seed42"
os.makedirs(out_bo, exist_ok=True)
fw_bo.run(environment, args_bo, out_bo)                   # writes results.csv

# 4) LLM-guided optimisation under same budget (600 iter x batch 5 = 3 000 evals)
fw_llm   = importlib.import_module("frameworks.v3_dynamic_optimizer.run")
args_llm = SimpleNamespace(
    iterations=600, batch_size=5, action="gaussian", gaussian_decay=False,
    gaussian_final_scale=0.1, sampler_model="gemini-2.5-flash-preview",
    designer_model=None, sampler_max_retries=3, hybrid=False,
    inspirations=2, initialize_n_sample=0, pw_alpha=3.0,
    num_islands=1, migration_interval=10, migration_rate=0.1,
    baseline=None, debug=False, save_fields=False, render_images=False,
)
out_llm = "results/BlendedNet/v3/shapebench_5_max_LD/seed42"
os.makedirs(out_llm, exist_ok=True)
fw_llm.run(environment, args_llm, out_llm)

# 5) Retrieve best design and score from run history
with open(f"{out_llm}/results.csv") as f:
    history = list(csv.DictReader(f))

best_row   = max(history, key=lambda r: float(r["best_reward"]))
best_x     = environment.read_design(
    f"{out_llm}/{best_row['design']}/{best_row['design']}.json"
)
best_score = float(best_row["best_reward"])
\end{lstlisting}

Listing~\ref{lst:api_bwb} illustrates a representative ShapeBench workflow on a 3D BWB multipoint mission task: instantiate a standardized task object, inspect its design/objective/condition metadata, run optimizers under a matched budget protocol, and perform final CFD verification with diagnostic reporting.

\subsection{Additional Results for Cross-category Studies --- Section \ref{main: cross-category}}

\begin{figure}[h!]
    \centering
    \includegraphics[width=1.0\linewidth]{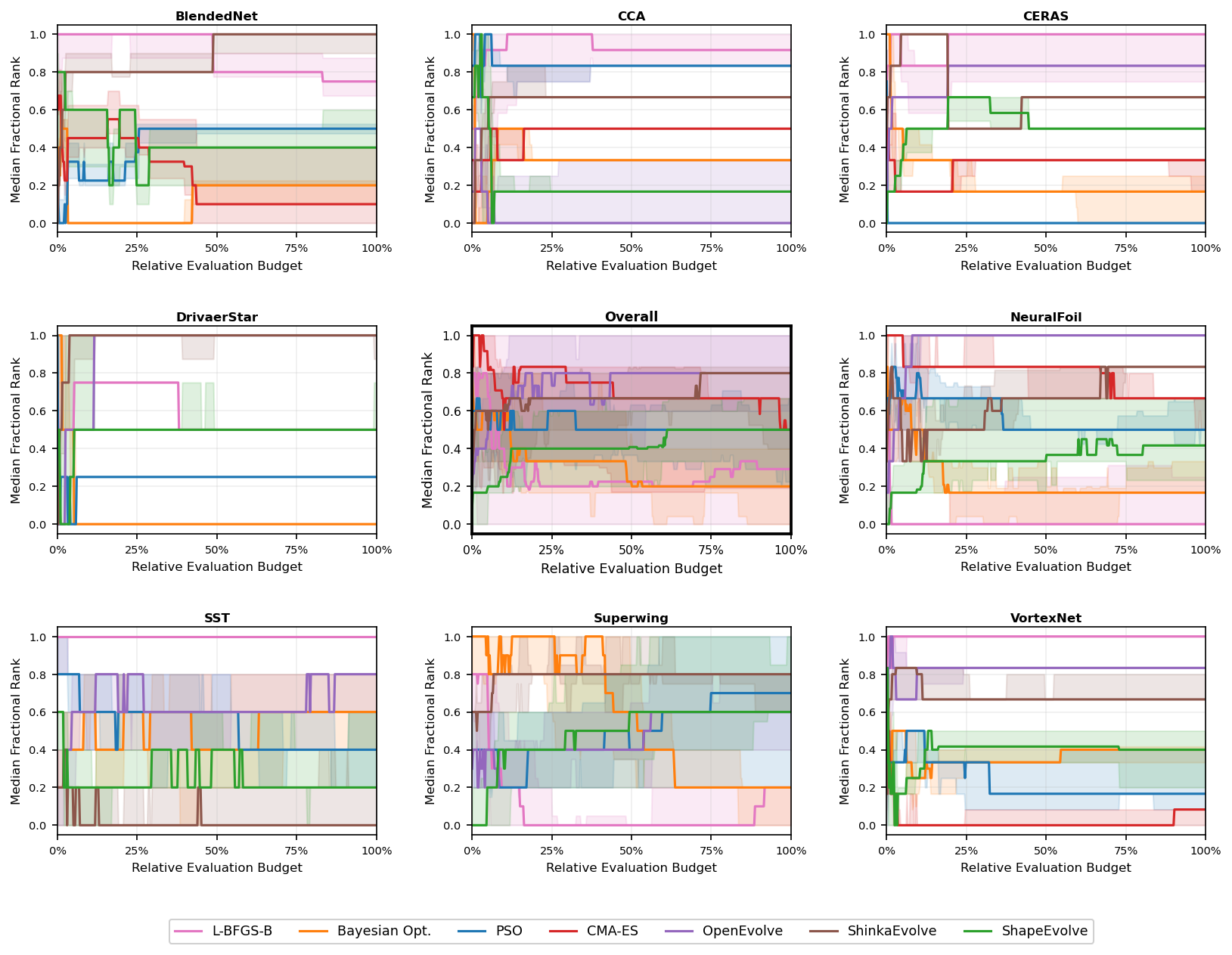}
    \caption{Median normalized rank trajectory of each optimizer over relative evaluation budget across ShapeBench environments. Each subplot shows the median normalized rank across all tasks in that environment, where normalized rank maps the best method to 0 and the worst to 1. Shaded bands show the inter-quartile range across tasks.}
    \label{fig:placeholder}
\end{figure}

\begin{table}[h!]
\centering
\begin{tabular}{lccccc}
\toprule
\textbf{Method} & \textbf{20\%} & \textbf{40\%} & \textbf{60\%} & \textbf{80\%} & \textbf{100\%} \\
\midrule
\textbf{L-BFGS-B} & \cellcolor[RGB]{237,248,252}\textcolor{black}{\makecell{0.333\\$\pm$1.000}} & \cellcolor[RGB]{220,241,249}\textcolor{black}{\makecell{0.225\\$\pm$1.000}} & \cellcolor[RGB]{214,238,248}\textcolor{black}{\makecell{0.200\\$\pm$1.000}} & \cellcolor[RGB]{230,245,251}\textcolor{black}{\makecell{0.292\\$\pm$1.000}} & \cellcolor[RGB]{230,245,251}\textcolor{black}{\makecell{0.292\\$\pm$1.000}} \\
\textbf{Bayesian Opt.} & \cellcolor[RGB]{237,248,252}\textcolor{black}{\makecell{0.333\\$\pm$0.633}} & \cellcolor[RGB]{237,248,252}\textcolor{black}{\makecell{0.333\\$\pm$0.583}} & \cellcolor[RGB]{214,238,248}\textcolor{black}{\makecell{0.200\\$\pm$0.400}} & \cellcolor[RGB]{214,238,248}\textcolor{black}{\makecell{0.200\\$\pm$0.233}} & \cellcolor[RGB]{214,238,248}\textcolor{black}{\makecell{0.200\\$\pm$0.400}} \\
\textbf{PSO} & \cellcolor[RGB]{255,254,254}\textcolor{black}{\makecell{0.500\\$\pm$0.333}} & \cellcolor[RGB]{255,254,254}\textcolor{black}{\makecell{0.500\\$\pm$0.300}} & \cellcolor[RGB]{255,254,254}\textcolor{black}{\makecell{0.500\\$\pm$0.300}} & \cellcolor[RGB]{255,254,254}\textcolor{black}{\makecell{0.500\\$\pm$0.375}} & \cellcolor[RGB]{255,254,254}\textcolor{black}{\makecell{0.500\\$\pm$0.442}} \\
\textbf{CMA-ES} & \cellcolor[RGB]{246,193,158}\textcolor{black}{\makecell{0.833\\$\pm$0.500}} & \cellcolor[RGB]{249,214,191}\textcolor{black}{\makecell{0.750\\$\pm$0.533}} & \cellcolor[RGB]{251,231,218}\textcolor{black}{\makecell{0.667\\$\pm$0.663}} & \cellcolor[RGB]{251,231,218}\textcolor{black}{\makecell{0.667\\$\pm$0.475}} & \cellcolor[RGB]{255,254,254}\textcolor{black}{\makecell{0.500\\$\pm$0.475}} \\
\textbf{OpenEvolve} & \cellcolor[RGB]{249,217,196}\textcolor{black}{\makecell{0.733\\$\pm$0.600}} & \cellcolor[RGB]{251,231,218}\textcolor{black}{\makecell{0.667\\$\pm$0.600}} & \cellcolor[RGB]{247,201,171}\textcolor{black}{\makecell{0.800\\$\pm$0.400}} & \cellcolor[RGB]{247,201,171}\textcolor{black}{\makecell{0.800\\$\pm$0.400}} & \cellcolor[RGB]{247,201,171}\textcolor{black}{\makecell{0.800\\$\pm$0.400}} \\
\textbf{ShinkaEvolve} & \cellcolor[RGB]{251,231,218}\textcolor{black}{\makecell{0.667\\$\pm$0.375}} & \cellcolor[RGB]{251,231,218}\textcolor{black}{\makecell{0.667\\$\pm$0.300}} & \cellcolor[RGB]{251,231,218}\textcolor{black}{\makecell{0.667\\$\pm$0.233}} & \cellcolor[RGB]{247,201,171}\textcolor{black}{\makecell{0.800\\$\pm$0.233}} & \cellcolor[RGB]{247,201,171}\textcolor{black}{\makecell{0.800\\$\pm$0.433}} \\
\textbf{ShapeEvolve} & \cellcolor[RGB]{244,251,253}\textcolor{black}{\makecell{0.400\\$\pm$0.375}} & \cellcolor[RGB]{244,251,253}\textcolor{black}{\makecell{0.400\\$\pm$0.400}} & \cellcolor[RGB]{246,251,254}\textcolor{black}{\makecell{0.417\\$\pm$0.317}} & \cellcolor[RGB]{255,254,254}\textcolor{black}{\makecell{0.500\\$\pm$0.400}} & \cellcolor[RGB]{255,254,254}\textcolor{black}{\makecell{0.500\\$\pm$0.467}} \\
\bottomrule
\end{tabular}
\caption{Median normalized rank trajectory of each optimizer over relative evaluation budget across all ShapeBench tasks. Shaded bands show the interquartile range across tasks.}
\label{tab:rank_table}
\end{table}

\subsection{Optimization Cost for Figure~\ref{variance table}}

\begin{table}[h!]
\centering
\caption{Computational cost 
per 1,000 evaluations (CPU-hours) for experiments of Figure~\ref{variance table}. }
\vspace{0.5em}
\label{cost table}
\renewcommand{\arraystretch}{1.25}

\resizebox{\textwidth}{!}{
\begin{tabular}{>{\raggedright\arraybackslash}p{0.20\textwidth}
                >{\centering\arraybackslash}p{0.2\textwidth}
                >{\centering\arraybackslash}p{0.16\textwidth}
                >{\centering\arraybackslash}p{0.25\textwidth}
                >{\centering\arraybackslash}p{0.16\textwidth}
                >{\centering\arraybackslash}p{0.16\textwidth}}
\toprule
\shortstack[l]{\rule{0pt}{3.1ex}\textbf{Optimizer}\\\textbf{Method}} &
\shortstack[c]{\rule{0pt}{3.1ex}\textbf{2D-AirFoil-SOMP}\\\textbf{Obj.: Weighted }$C_D$} &
\shortstack[c]{\rule{0pt}{3.1ex}\textbf{3D-CAR-SOSP}\\\textbf{Obj.: }$C_D$} &
\shortstack[c]{\rule{0pt}{3.1ex}\textbf{3D-Delta-Wing-MOMP}\\\textbf{Obj.: $-(K_n - w_iC_D)$}} &
\shortstack[c]{\rule{0pt}{3.1ex}\textbf{3D-CCA-SOMP}\\\textbf{Obj.: $- (L / D)$}}&
\shortstack[c]{\rule{0pt}{3.1ex}\textbf{CERAS-SOSP}\\\textbf{Obj.: FuelMass}} \\
\midrule
Adjoint (IPOPT) & $\approx 0.01$ & N/A & N/A & N/A & N/A \\
\hline
L-BFGS-B & $0.05-0.10$ & $1.4$--$3.0$ & $0.93$ & $559$ & $28.4$ \\
\hline
Bayesian Opt. & $ \approx 1{,}100$ -- $3{,}500$ & $13$--$28$ & $8.09$ & $599$ & $43.3$ \\
\hline
PSO & $0.05-0.10$ & $\approx 2.7$ & $0.076$ & $688$ & $78.9$ \\
\hline
CMA-ES & $0.199$ & $1.75$ & $0.071$ & $939$ & $25.6$ \\
\hline
OpenEvolve & $8.20$ & $8.83$ & $3.6$ & $562$ & $18.7$ \\
\hline
ShinkaEvolve & $9.85$ & $2.91$ & $6.47$ & $230$ & $100$ \\
\hline
ShapeEvolve & $0.8-1.0$ & $3.3$--$5.1$ & $3.29$ & $385$ & $71.6$ \\
\bottomrule
\end{tabular}
}
\end{table}

\subsection{Diagnostic Suite}
\label{app: diagnostic suite}

The ShapeBench Diagnostic Suite evaluates whether a candidate design is not only high-scoring but also physically and numerically credible. It is intentionally separated from the optimization objective: objective values measure task performance, while diagnostics measure \emph{validity}. This prevents objective-only comparisons from promoting non-physical false positives.

\paragraph{Overview.}
For each evaluated design $x$ under operating condition $\mu$, the suite computes three diagnostic tiers:
\begin{enumerate}
    \item \textbf{Optimization feasibility} (domain and support checks),
    \item \textbf{Geometric validity} (shape realizability and regularity),
    \item \textbf{Aerodynamic plausibility} (flow/force consistency checks).
\end{enumerate}

\begin{table}[h!]
\centering
\footnotesize
\caption{ShapeBench Diagnostic Suite. Deterministic checks provide primary validity signals; outputs are schema-validated and saved in a JSON.}
\vspace{0.3em}
\label{tab:diagnostic_suite}
\renewcommand{\arraystretch}{1.15}
\resizebox{0.95\textwidth}{!}{
\begin{tabular}{>{\raggedright\arraybackslash}p{0.2\textwidth}
                >{\raggedright\arraybackslash}p{0.36\textwidth}
                >{\raggedright\arraybackslash}p{0.22\textwidth}
                >{\raggedright\arraybackslash}p{0.25\textwidth}}
\toprule
\shortstack[l]{\rule{0pt}{3.0ex}\textbf{Diagnostic Tier}} &
\shortstack[l]{\rule{0pt}{3.0ex}\textbf{Representative Checks}} &
\shortstack[l]{\rule{0pt}{3.0ex}\textbf{Required Inputs}} &
\shortstack[l]{\rule{0pt}{3.0ex}\textbf{Primary Outputs}} \\
\midrule

\textbf{Optimization Feasibility}
&
Required parameter/metric presence; parameter bound compliance \mbox{($x_i \in [l_i,u_i]$)}; required model artifacts (geometry mesh + normalization stats); body-style / normalization compatibility.
&
Design parameters, reported metrics, artifact paths, environment metadata.
&
Per-check status; severity score in \([0,1]\); violated keys; feasibility notes.
\\
\hline

\textbf{Geometric Validity}
&
Near-bound concentration ratio (boundary-collapse risk); combined angle stress \mbox{($\sum_j |\theta_j|$)}; global deformation coupling (scale/width/length consistency heuristic).
&
Design parameters and parameter bounds.
&
Geometry-risk flags; threshold-relative indicators; per-check severity and messages.
\\
\hline

\textbf{Aerodynamic Plausibility}
&
Physics-consistency and regime-sanity checks, including force/moment closure, coefficient plausibility (\(C_L, C_D, C_M\)), multi-point trend consistency, field-based plausibility (\(C_p/C_f\)-derived signals), and task-specific checks (e.g., trim/stability for aircraft, induced-drag consistency for wings).
&
Integrated metrics and coefficients, optional pressure/skin-friction field statistics, operating conditions, and rendered flow diagnostics.
&
Consistency residuals, plausibility/stability flags, confidence-aware risk signals, and tier severity.
\\
\hline

\textbf{Structured Diagnostic Output}
&
Schema validation of deterministic evidence bundle and LLM integrative report.
&
All tier outputs + context/images.
&
JSON containing evidence summary, data-quality notes, failure mechanisms, and recommended mitigations.
\\
\bottomrule
\end{tabular}
}
\end{table}

\begin{lstlisting} [language=Python, caption={Example of a full diagnostic report for the 3D car design problem}, label={lst:api_bwb}, basicstyle=\ttfamily\fontsize{7}{8}\selectfont]
    {
  "version": "0.1.0",
  "environment": "DrivAer_Star",
  "design_id": "iter_517_o2",
  "timestamp_utc": "2026-04-30T04:31:27Z",
  "input_snapshot": {
    "environment": "DrivAer_Star",
    "design_id": "iter_517_o2",
    "design_path": "/scratch/ShapeEvolve/environments/DrivAer_Star/results/run_v3_dynamic_optimizer_cd_only_drivaer_star_attempt_13_flash_2_5_n10000/iter_517_o2.json",
    "case_dir": "/scratch/ShapeEvolve/environments/DrivAer_Star/results/run_v3_dynamic_optimizer_cd_only_drivaer_star_attempt_13_flash_2_5_n10000/iter_517_o2",
    "design_params": {
      "car_size": 0.8,
      "car_width": 0.040242888869530376,
      "car_len": -0.1,
      "ramp_angle": 7.121438172129935,
      "front_bumper_length": 0.1,
      "wind_screen_x": -0.05,
      "wind_screen_z": 0.05,
      "side_mirrors_x": -0.009025765712378033,
      "side_mirrors_z": 0.05,
      "rear_window_x": -0.05,
      "rear_window_z": -0.046316046991307466,
      "trunklid_angle": 8.0,
      "trunklid_x": 0.05,
      "trunklid_z": -0.05,
      "diffusor_angle": -8.0,
      "car_green_house_angle": 8.0,
      "car_front_hood_angle": 8.0,
      "car_air_intake_angle": 8.0,
      "tires_diameter": 0.013,
      "tires_width": -0.00781327375478008,
      "name": "refined_kammback_optimizer"
    },
    "metrics": {
      "drag": 154.4256248474121,
      "Cd": 0.06515849149679837,
      "lift": -153.39218473434448,
      "drag_pressure": 120.70740509033203,
      "drag_shear": 33.71821975708008
    },
    "images": [
      "/scratch/ShapeEvolve/environments/DrivAer_Star/results/run_v3_dynamic_optimizer_cd_only_drivaer_star_attempt_13_flash_2_5_n10000/iter_517_o2/save/sol/Pressure_iso.png",
      "/scratch/ShapeEvolve/environments/DrivAer_Star/results/run_v3_dynamic_optimizer_cd_only_drivaer_star_attempt_13_flash_2_5_n10000/iter_517_o2/save/sol/Pressure_top.png",
      "/scratch/ShapeEvolve/environments/DrivAer_Star/results/run_v3_dynamic_optimizer_cd_only_drivaer_star_attempt_13_flash_2_5_n10000/iter_517_o2/save/sol/Pressure_side.png",
      "/scratch/ShapeEvolve/environments/DrivAer_Star/results/run_v3_dynamic_optimizer_cd_only_drivaer_star_attempt_13_flash_2_5_n10000/iter_517_o2/save/sol/WSSx_iso.png",
      "/scratch/ShapeEvolve/environments/DrivAer_Star/results/run_v3_dynamic_optimizer_cd_only_drivaer_star_attempt_13_flash_2_5_n10000/iter_517_o2/save/sol/WSSx_top.png",
      "/scratch/ShapeEvolve/environments/DrivAer_Star/results/run_v3_dynamic_optimizer_cd_only_drivaer_star_attempt_13_flash_2_5_n10000/iter_517_o2/save/sol/WSSx_side.png"
    ],
    "field_stats": {},
    "model_artifacts": {
      "base_vtk_path": "/scratch/ShapeEvolve/environments/DrivAer_Star/data/vtk_E/00000.vtk",
      "norm_stats_path": "/scratch/ShapeEvolve_diagnostic_suite/environments/DrivAer_Star/model/norm_stats.pt"
    },
    "problem_setup": {
      "objective": {
        "name": "minimize Cd",
        "reward": "-Cd"
      },
      "method": "ShapeEvolve (v3_dynamic_optimizer, Gemini Flash 2.5, n=10000)",
      "best_Cd": 0.06515849,
      "constraints": {
        "hard": [
          "all 20 parameters within bounds"
        ]
      },
      "param_bounds": {
        "car_size": [
          0.8,
          1.2
        ],
        "car_width": [
          -0.1,
          0.1
        ],
        "car_len": [
          -0.1,
          0.1
        ],
        "ramp_angle": [
          -8.0,
          8.0
        ],
        "front_bumper_length": [
          -0.1,
          0.1
        ],
        "wind_screen_x": [
          -0.05,
          0.05
        ],
        "wind_screen_z": [
          -0.05,
          0.05
        ],
        "side_mirrors_x": [
          -0.05,
          0.05
        ],
        "side_mirrors_z": [
          -0.05,
          0.05
        ],
        "rear_window_x": [
          -0.05,
          0.05
        ],
        "rear_window_z": [
          -0.05,
          0.05
        ],
        "trunklid_angle": [
          -8.0,
          8.0
        ],
        "trunklid_x": [
          -0.05,
          0.05
        ],
        "trunklid_z": [
          -0.05,
          0.05
        ],
        "diffusor_angle": [
          -8.0,
          8.0
        ],
        "car_green_house_angle": [
          -8.0,
          8.0
        ],
        "car_front_hood_angle": [
          -8.0,
          8.0
        ],
        "car_air_intake_angle": [
          -8.0,
          8.0
        ],
        "tires_diameter": [
          -0.013,
          0.013
        ],
        "tires_width": [
          -0.015,
          0.015
        ]
      },
      "operating_conditions": {
        "body_style": "vtk_E (Estateback)",
        "rho": 1.25,
        "u": 40.0,
        "area_ref": 2.37
      }
    },
    "run_context": {},
    "raw_feedback": "",
    "timestamp_utc": null
  },
  "evidence_bundle": {
    "environment": "DrivAer_Star",
    "design_id": "iter_517_o2",
    "feasibility": [
      {
        "check_id": "F001_required_params_present",
        "tier": "feasibility",
        "status": "ok",
        "severity": 0.0,
        "message": "All required parameters present.",
        "value": {
          "missing": []
        },
        "threshold": null,
        "evidence_refs": [
          "/scratch/ShapeEvolve/environments/DrivAer_Star/results/run_v3_dynamic_optimizer_cd_only_drivaer_star_attempt_13_flash_2_5_n10000/iter_517_o2.json"
        ],
        "metadata": {}
      },
      {
        "check_id": "F002_param_bounds_respected",
        "tier": "feasibility",
        "status": "ok",
        "severity": 0.0,
        "message": "All parameter values within bounds.",
        "value": {
          "violations": []
        },
        "threshold": "within configured bounds",
        "evidence_refs": [
          "/scratch/ShapeEvolve/environments/DrivAer_Star/results/run_v3_dynamic_optimizer_cd_only_drivaer_star_attempt_13_flash_2_5_n10000/iter_517_o2.json",
          "/scratch/ShapeEvolve/environments/DrivAer_Star/results/run_v3_dynamic_optimizer_cd_only_drivaer_star_attempt_13_flash_2_5_n10000/iter_517_o2"
        ],
        "metadata": {}
      },
      {
        "check_id": "F003_base_vtk_exists",
        "tier": "feasibility",
        "status": "ok",
        "severity": 0.0,
        "message": "Base VTK exists.",
        "value": "/scratch/ShapeEvolve/environments/DrivAer_Star/data/vtk_E/00000.vtk",
        "threshold": null,
        "evidence_refs": [
          "/scratch/ShapeEvolve/environments/DrivAer_Star/data/vtk_E/00000.vtk"
        ],
        "metadata": {}
      },
      {
        "check_id": "F004_norm_stats_exists",
        "tier": "feasibility",
        "status": "ok",
        "severity": 0.0,
        "message": "Norm stats file exists.",
        "value": "/scratch/ShapeEvolve_diagnostic_suite/environments/DrivAer_Star/model/norm_stats.pt",
        "threshold": null,
        "evidence_refs": [
          "/scratch/ShapeEvolve_diagnostic_suite/environments/DrivAer_Star/model/norm_stats.pt"
        ],
        "metadata": {}
      },
      {
        "check_id": "F005_metrics_finite",
        "tier": "feasibility",
        "status": "ok",
        "severity": 0.0,
        "message": "All required metrics are finite.",
        "value": {
          "missing": [],
          "non_finite": []
        },
        "threshold": null,
        "evidence_refs": [
          "/scratch/ShapeEvolve/environments/DrivAer_Star/results/run_v3_dynamic_optimizer_cd_only_drivaer_star_attempt_13_flash_2_5_n10000/iter_517_o2.json",
          "/scratch/ShapeEvolve/environments/DrivAer_Star/results/run_v3_dynamic_optimizer_cd_only_drivaer_star_attempt_13_flash_2_5_n10000/iter_517_o2"
        ],
        "metadata": {}
      },
      {
        "check_id": "F006_body_style_norm_compatibility",
        "tier": "feasibility",
        "status": "ok",
        "severity": 0.0,
        "message": "Norm stats compatible with inferred body style 'E'.",
        "value": {
          "style": "E",
          "norm_stats_path": "/scratch/ShapeEvolve_diagnostic_suite/environments/DrivAer_Star/model/norm_stats.pt"
        },
        "threshold": null,
        "evidence_refs": [
          "/scratch/ShapeEvolve/environments/DrivAer_Star/results/run_v3_dynamic_optimizer_cd_only_drivaer_star_attempt_13_flash_2_5_n10000/iter_517_o2.json",
          "/scratch/ShapeEvolve/environments/DrivAer_Star/results/run_v3_dynamic_optimizer_cd_only_drivaer_star_attempt_13_flash_2_5_n10000/iter_517_o2"
        ],
        "metadata": {}
      }
    ],
    "geometry": [
      {
        "check_id": "G001_param_extremeness_ratio",
        "tier": "geometry",
        "status": "warning",
        "severity": 0.9,
        "message": "High fraction of parameters near bounds (0.80).",
        "value": {
          "near_bound_fraction": 0.8,
          "near_bound_keys": [
            "car_size",
            "car_len",
            "front_bumper_length",
            "wind_screen_x",
            "wind_screen_z",
            "side_mirrors_z",
            "rear_window_x",
            "rear_window_z",
            "trunklid_angle",
            "trunklid_x",
            "trunklid_z",
            "diffusor_angle",
            "car_green_house_angle",
            "car_front_hood_angle",
            "car_air_intake_angle",
            "tires_diameter"
          ]
        },
        "threshold": {
          "warn_fraction": 0.6,
          "margin_ratio": 0.05
        },
        "evidence_refs": [
          "/scratch/ShapeEvolve/environments/DrivAer_Star/results/run_v3_dynamic_optimizer_cd_only_drivaer_star_attempt_13_flash_2_5_n10000/iter_517_o2.json",
          "/scratch/ShapeEvolve/environments/DrivAer_Star/results/run_v3_dynamic_optimizer_cd_only_drivaer_star_attempt_13_flash_2_5_n10000/iter_517_o2"
        ],
        "metadata": {}
      },
      {
        "check_id": "G002_combined_angle_stress",
        "tier": "geometry",
        "status": "warning",
        "severity": 0.9061815033101911,
        "message": "Combined angle stress is high (47.12 deg abs-sum).",
        "value": {
          "combined_abs_angle_sum": 47.121438172129935
        },
        "threshold": {
          "warn_sum": 26.0
        },
        "evidence_refs": [
          "/scratch/ShapeEvolve/environments/DrivAer_Star/results/run_v3_dynamic_optimizer_cd_only_drivaer_star_attempt_13_flash_2_5_n10000/iter_517_o2.json",
          "/scratch/ShapeEvolve/environments/DrivAer_Star/results/run_v3_dynamic_optimizer_cd_only_drivaer_star_attempt_13_flash_2_5_n10000/iter_517_o2"
        ],
        "metadata": {}
      },
      {
        "check_id": "G003_size_width_length_coupling",
        "tier": "geometry",
        "status": "warning",
        "severity": 0.8008096295651012,
        "message": "Global scale + width/length coupling is aggressive; geometry realism risk increased.",
        "value": {
          "car_size": 0.8,
          "abs_car_width": 0.040242888869530376,
          "abs_car_len": 0.1,
          "coupling_score": 2.4024288886953036
        },
        "threshold": {
          "warn_score": 2.4
        },
        "evidence_refs": [
          "/scratch/ShapeEvolve/environments/DrivAer_Star/results/run_v3_dynamic_optimizer_cd_only_drivaer_star_attempt_13_flash_2_5_n10000/iter_517_o2.json",
          "/scratch/ShapeEvolve/environments/DrivAer_Star/results/run_v3_dynamic_optimizer_cd_only_drivaer_star_attempt_13_flash_2_5_n10000/iter_517_o2"
        ],
        "metadata": {}
      }
    ],
    "aero": [
      {
        "check_id": "A001_drag_decomposition_consistency",
        "tier": "aero",
        "status": "ok",
        "severity": 0.0,
        "message": "Drag decomposition consistent (rel_err=0.00000).",
        "value": {
          "drag": 154.4256248474121,
          "drag_pressure_plus_shear": 154.4256248474121,
          "rel_err": 0.0
        },
        "threshold": {
          "warn_rel_err": 0.02
        },
        "evidence_refs": [
          "/scratch/ShapeEvolve/environments/DrivAer_Star/results/run_v3_dynamic_optimizer_cd_only_drivaer_star_attempt_13_flash_2_5_n10000/iter_517_o2.json",
          "/scratch/ShapeEvolve/environments/DrivAer_Star/results/run_v3_dynamic_optimizer_cd_only_drivaer_star_attempt_13_flash_2_5_n10000/iter_517_o2"
        ],
        "metadata": {}
      },
      {
        "check_id": "A002_cd_plausible_range",
        "tier": "aero",
        "status": "ok",
        "severity": 0.0,
        "message": "Cd within plausible warning band.",
        "value": 0.06515849149679837,
        "threshold": {
          "min": 0.0,
          "max": 1.5
        },
        "evidence_refs": [
          "/scratch/ShapeEvolve/environments/DrivAer_Star/results/run_v3_dynamic_optimizer_cd_only_drivaer_star_attempt_13_flash_2_5_n10000/iter_517_o2.json",
          "/scratch/ShapeEvolve/environments/DrivAer_Star/results/run_v3_dynamic_optimizer_cd_only_drivaer_star_attempt_13_flash_2_5_n10000/iter_517_o2"
        ],
        "metadata": {}
      },
      {
        "check_id": "A003_lift_plausible_range",
        "tier": "aero",
        "status": "ok",
        "severity": 0.0,
        "message": "Lift magnitude within plausible warning range.",
        "value": -153.39218473434448,
        "threshold": {
          "warn_abs": 200000.0
        },
        "evidence_refs": [
          "/scratch/ShapeEvolve/environments/DrivAer_Star/results/run_v3_dynamic_optimizer_cd_only_drivaer_star_attempt_13_flash_2_5_n10000/iter_517_o2.json",
          "/scratch/ShapeEvolve/environments/DrivAer_Star/results/run_v3_dynamic_optimizer_cd_only_drivaer_star_attempt_13_flash_2_5_n10000/iter_517_o2"
        ],
        "metadata": {}
      },
      {
        "check_id": "A004_image_availability_signal",
        "tier": "aero",
        "status": "ok",
        "severity": 0.0,
        "message": "All expected flow images are available.",
        "value": {
          "present": 6,
          "total": 6,
          "coverage": 1.0,
          "suffix_map": {
            "Pressure_iso.png": true,
            "Pressure_top.png": true,
            "Pressure_side.png": true,
            "WSSx_iso.png": true,
            "WSSx_top.png": true,
            "WSSx_side.png": true
          }
        },
        "threshold": {
          "expected_total": 6
        },
        "evidence_refs": [
          "/scratch/ShapeEvolve/environments/DrivAer_Star/results/run_v3_dynamic_optimizer_cd_only_drivaer_star_attempt_13_flash_2_5_n10000/iter_517_o2/save/sol/Pressure_iso.png",
          "/scratch/ShapeEvolve/environments/DrivAer_Star/results/run_v3_dynamic_optimizer_cd_only_drivaer_star_attempt_13_flash_2_5_n10000/iter_517_o2/save/sol/Pressure_top.png",
          "/scratch/ShapeEvolve/environments/DrivAer_Star/results/run_v3_dynamic_optimizer_cd_only_drivaer_star_attempt_13_flash_2_5_n10000/iter_517_o2/save/sol/Pressure_side.png",
          "/scratch/ShapeEvolve/environments/DrivAer_Star/results/run_v3_dynamic_optimizer_cd_only_drivaer_star_attempt_13_flash_2_5_n10000/iter_517_o2/save/sol/WSSx_iso.png",
          "/scratch/ShapeEvolve/environments/DrivAer_Star/results/run_v3_dynamic_optimizer_cd_only_drivaer_star_attempt_13_flash_2_5_n10000/iter_517_o2/save/sol/WSSx_top.png",
          "/scratch/ShapeEvolve/environments/DrivAer_Star/results/run_v3_dynamic_optimizer_cd_only_drivaer_star_attempt_13_flash_2_5_n10000/iter_517_o2/save/sol/WSSx_side.png"
        ],
        "metadata": {}
      }
    ],
    "summary": {
      "feasibility": {
        "ok": 6,
        "warning": 0,
        "issue": 0,
        "error": 0,
        "missing": 0
      },
      "geometry": {
        "ok": 0,
        "warning": 3,
        "issue": 0,
        "error": 0,
        "missing": 0
      },
      "aero": {
        "ok": 4,
        "warning": 0,
        "issue": 0,
        "error": 0,
        "missing": 0
      }
    },
    "data_quality_notes": []
  },
  "llm_report": {
    "diagnostic_status": "complete",
    "overall_assessment": "The candidate design 'iter_517_o2' exhibits significant geometric deformation issues, primarily characterized by ground clearance collapse at the rear, driven by multiple parameters pushed to their bounds. While all feasibility and aerodynamic consistency checks pass, the resulting drag coefficient (Cd=0.065) is extremely low and highly suspicious for a DrivAer vehicle, suggesting the surrogate model may be exploiting unrealistic geometries. The visual evidence strongly supports the diagnosis of excessive deformation.",
    "primary_failure_mechanisms": [
      "GEOMETRY_DEFORMATION_EXCESSIVE"
    ],
    "secondary_risks": [
      "PARAM_BOUNDARY_COLLAPSE",
      "AERO_PLAUSIBILITY_WEAK",
      "SURROGATE_DOMAIN_RISK"
    ],
    "surrogate_exploitation_risk": "high",
    "physical_credibility": "low",
    "confidence": 0.9,
    "evidence_weighting_rationale": "Hard data from geometry checks (G001, G002, G003) clearly indicates extreme parameter values and aggressive deformations. This deterministic evidence is strongly corroborated by visual inspection of the pressure and WSSx images, which show a car with severely reduced ground clearance at the rear. The extremely low Cd, while passing a broad plausibility check, further reinforces the suspicion of an unphysical design resulting from surrogate exploitation.",
    "recommended_mitigations": [
      "TIGHTEN_PARAMETER_BOUNDS",
      "ADD_GEOMETRY_REGULARIZATION",
      "ADD_PHYSICS_CONSISTENCY_PENALTY"
    ],
    "recommended_next_tests": [
      "Perform a high-fidelity CFD validation for this design to confirm the Cd value and flow field realism.",
      "Implement a ground clearance constraint in the optimization objective or as a hard constraint.",
      "Visualize the mesh quality of the deformed geometry to check for local collapses or inversions, especially at the rear underbody."
    ],
    "notes_on_missing_evidence": [],
    "citations_to_evidence": [
      "G001_param_extremeness_ratio",
      "G002_combined_angle_stress",
      "G003_size_width_length_coupling",
      "A002_cd_plausible_range",
      "Pressure_side.png",
      "WSSx_side.png"
    ],
    "model_name": "gemini-2.5-flash",
    "prompt_version": "drivaer_mvp_v1",
    "raw_response": "{\n  \"diagnostic_status\": \"complete\",\n  \"overall_assessment\": \"The candidate design 'iter_517_o2' exhibits significant geometric deformation issues, primarily characterized by ground clearance collapse at the rear, driven by multiple parameters pushed to their bounds. While all feasibility and aerodynamic consistency checks pass, the resulting drag coefficient (Cd=0.065) is extremely low and highly suspicious for a DrivAer vehicle, suggesting the surrogate model may be exploiting unrealistic geometries. The visual evidence strongly supports the diagnosis of excessive deformation.\",\n  \"primary_failure_mechanisms\": [\n    \"GEOMETRY_DEFORMATION_EXCESSIVE\"\n  ],\n  \"secondary_risks\": [\n    \"PARAM_BOUNDARY_COLLAPSE\",\n    \"AERO_PLAUSIBILITY_WEAK\",\n    \"SURROGATE_DOMAIN_RISK\"\n  ],\n  \"surrogate_exploitation_risk\": \"high\",\n  \"physical_credibility\": \"low\",\n  \"confidence\": 0.9,\n  \"evidence_weighting_rationale\": \"Hard data from geometry checks (G001, G002, G003) clearly indicates extreme parameter values and aggressive deformations. This deterministic evidence is strongly corroborated by visual inspection of the pressure and WSSx images, which show a car with severely reduced ground clearance at the rear. The extremely low Cd, while passing a broad plausibility check, further reinforces the suspicion of an unphysical design resulting from surrogate exploitation.\",\n  \"recommended_mitigations\": [\n    \"TIGHTEN_PARAMETER_BOUNDS\",\n    \"ADD_GEOMETRY_REGULARIZATION\",\n    \"ADD_PHYSICS_CONSISTENCY_PENALTY\"\n  ],\n  \"recommended_next_tests\": [\n    \"Perform a high-fidelity CFD validation for this design to confirm the Cd value and flow field realism.\",\n    \"Implement a ground clearance constraint in the optimization objective or as a hard constraint.\",\n    \"Visualize the mesh quality of the deformed geometry to check for local collapses or inversions, especially at the rear underbody.\"\n  ],\n  \"notes_on_missing_evidence\": [],\n  \"citations_to_evidence\": [\n    \"G001_param_extremeness_ratio\",\n    \"G002_combined_angle_stress\",\n    \"G003_size_width_length_coupling\",\n    \"A002_cd_plausible_range\",\n    \"Pressure_side.png\",\n    \"WSSx_side.png\"\n  ],\n  \"model_name\": \"refined_kammback_optimizer\",\n  \"prompt_version\": null,\n  \"raw_response\": null,\n  \"parser_warnings\": []\n}",
    "parser_warnings": []
  },
  "trace": {
    "pipeline_version": "phase4_mvp_v1",
    "llm_diagnostic_status": "complete"
  },
  "provenance": {}
}
\end{lstlisting}
\clearpage 

\section{Surrogate Models} \label{app: surrogate models}
In this section we introduce all of the surrogate models included in ShapeBench.

\subsection{\texttt{VortexNet}}
\texttt{VortexNet}\citep{shen2025vortexnet} is a multi-fidelity field surrogate model that augments low fidelity VLM simulation using learned corrections from CFD simulations for wings in steady flow:

\begin{itemize}
  \item \textbf{Quantity of Interest:} The local pressure difference between lower and upper surface of the wing, 
  \[
    \Delta C_p \;=\; C_{p,\mathrm{lower}} - C_{p,\mathrm{upper}},
  \]
  that is used in VLM for computing wing aerodynamic coefficients (e.g. $C_L$, $C_M$).

  \item \textbf{Multi-fidelity Pressure Field:} 
  \begin{itemize}
      \item Low-fidelity input: steady state pressure field from VLM (via SUAVE) producing $\Delta C_{p,\mathrm{LF}}$. 
      \item Multi-fidelity output: $\Delta C_{p,\mathrm{MF}} = \mathcal{F}(\Delta C_{p,\mathrm{LF}})$, where $\mathcal{F}$ is a learnt VortexNet model.
      \item High-fidelity reference: steady RANS CFD simulation using \texttt{SU2} solver. The simulation setup uses Spalart--Allmaras turbulence model with Rotation correction~\citep{Dacles-Mariani1995Numericalexperimental} with no-slip, adiabatic wall on the wing surface. The free-stream conditions are sampled from bounds $Re=[6.5,10]\times 10^7$, angle-of-attack from $0^\circ$ to $20^\circ$, and Mach number from 0.35 to 0.5. For each wing geometry, 40 free-stream samples are generated independently. 
  \end{itemize}
\end{itemize}

\begin{figure}[h!]
    \centering
    \includegraphics[width=0.7\linewidth]{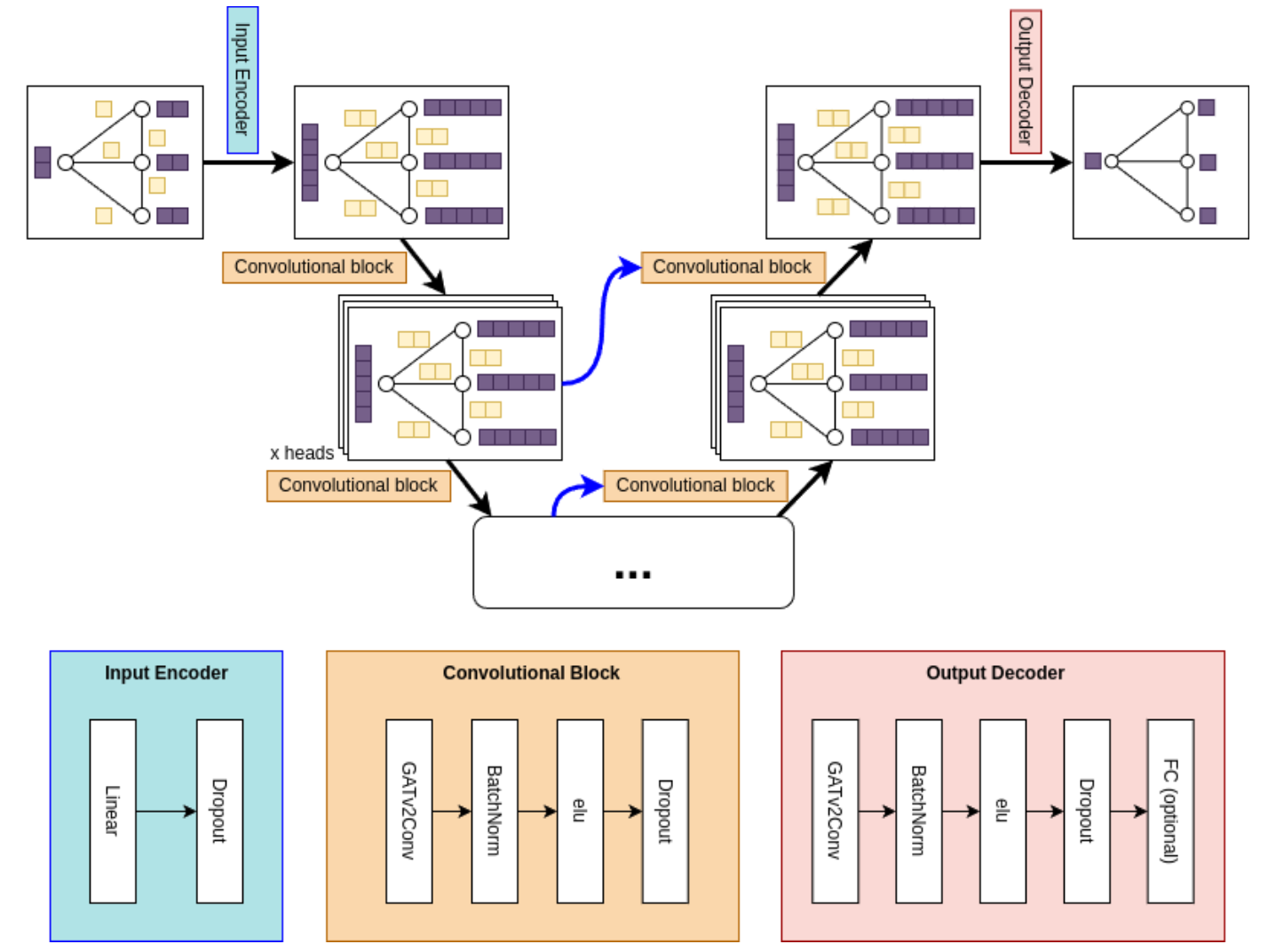}
    \caption{Schematic of the proposed \texttt{VortexNet} graph neural network (GNN) architecture. Colors indicate different network blocks. A black arrow indicates represents direct message passing between blocks, and a blue arrow denotes a skip connection to the receiving block. The figure also presents snapshots of the graph at each computational step, showing nodes, edges, and their associated feature arrays. Figure is from \citep{shen2025vortexnet}.}
    \label{fig: vortexnet}
\end{figure}

\paragraph{Wing geometry variations}

\begin{itemize}
\item \textbf{Wing type}

The entire dataset is \textbf{3-D Delta wings only}.

\item \textbf{Design variables}

They explicitly parameterize the delta wings using {three design variables}:
\begin{enumerate}
  \item \textbf{Leading-edge sweep angle $\Lambda_{LE}$:}
  \begin{itemize}
    \item $55^\circ,\ 65^\circ,\ 75^\circ$
  \end{itemize}

  \item \textbf{Root airfoil shape (NACA 4-digit):}
  \begin{itemize}
    \item NACA0010, NACA0016, NACA0024, NACA2416, NACA4416
  \end{itemize}
  \item \textbf{Root chord length:}  
  \begin{itemize}
      \item Fixed at 0.65 m; changing sweep therefore changes span and reference area.
  \end{itemize}
\end{enumerate}

They generate \textbf{all 15 combinations} ($3$ sweeps $\times\ 5$ airfoils). They also tested interpolation inside this design space with new geometries like $60^\circ/70^\circ$ sweep and NACA0013/NACA3416 to probe generalization, but those are still within the same delta-wing concept and same parameter types. 

\begin{figure}[h!]
    \centering
\includegraphics[width=0.7\linewidth]{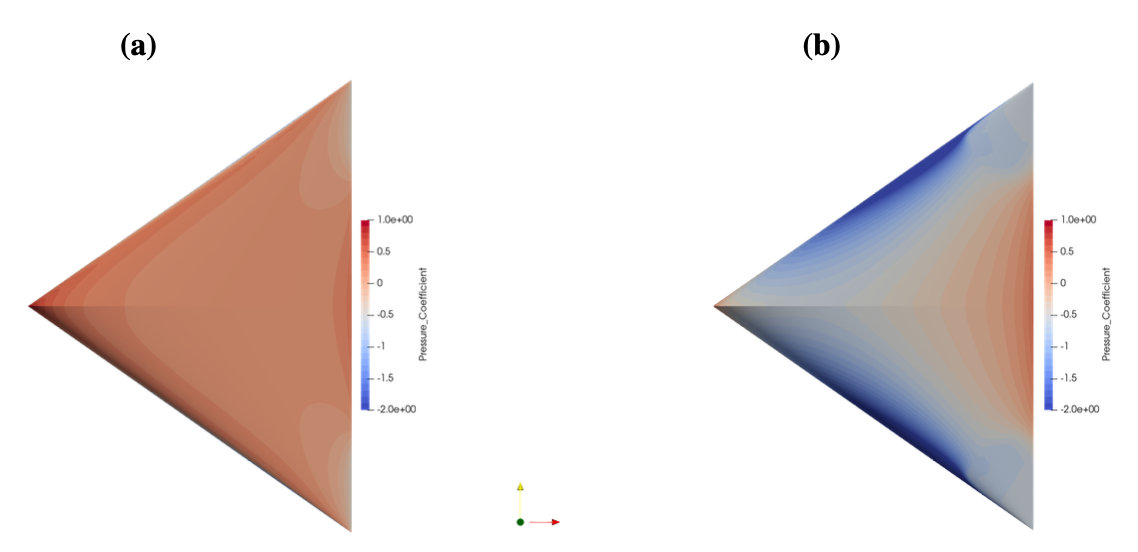}
    \caption{The HF surface pressure coefficient obtained from CFD for a wing with $55^\circ$ leading edge sweep and root airfoil profile of NACA2416 at AOA$=16.4^\circ$, $M_a=0.47$, and $R_e=9.78\times 10^{6}$. Figure is from \citep{shen2025vortexnet}}
    \label{fig:delta wing}
\end{figure}
\end{itemize}

\subsection{\texttt{BlendedNet}} \label{app: blendednet}
BlendedNet \citep{Sung_2025} introduces (i) a large, publicly available, high-fidelity aerodynamic dataset for
\emph{blended wing body (BWB)} aircraft and (ii) an end-to-end surrogate modeling pipeline for \emph{pointwise}
surface aerodynamic coefficient prediction. The dataset contains 999 unique BWB geometries, each simulated across
approximately 9 flight conditions, for a total of 8{,}830 successfully converged CFD cases. \texttt{BlendedNet}'s surrogate
predicts pointwise surface quantities that can be integrated to obtain global lift and drag.

\begin{itemize}
  \item \textbf{Target quantities (what it outputs):} pointwise surface aerodynamic coefficients:
  \[
    \mathbf{u}(\mathbf{x},\mathbf{n})
    =
    \begin{bmatrix}
      C_p(\mathbf{x},\mathbf{n})\\
      C_{f_x}(\mathbf{x},\mathbf{n})\\
      C_{f_z}(\mathbf{x},\mathbf{n})
    \end{bmatrix},
  \]
  where $\mathbf{x}\in\mathbb{R}^3$ is a surface coordinate and $\mathbf{n}\in\mathbb{R}^3$ is the corresponding unit surface
  normal.

  \item \textbf{Physics reference (``high fidelity''):} steady Reynolds-Averaged Navier--Stokes (RANS) CFD using NASA
  \texttt{FUN3D}, with the Spalart--Allmaras turbulence model. 

  \item \textbf{Mesh / resolution (dataset generation):} each CFD case uses a Pointwise-generated unstructured volume mesh
  with roughly 9--14 million volume cells, and surface meshes ranging from 34{,}729 to 85{,}392 points, with boundary-layer
  resolution selected to maintain $y^+<1$. 

  \item \textbf{Dataset outputs stored per case:} in addition to integrated coefficients, the dataset stores surface fields in VTK
  format, including $C_p$ and skin-friction components ($C_{f_x},C_{f_y},C_{f_z}$). 
\end{itemize}

\paragraph{Surrogate model overview (two-stage end-to-end pipeline).}
The \texttt{BlendedNet} surrogate is composed of two neural networks used in sequence:
(1) a permutation-invariant \textbf{PointNet regressor} that infers BWB \emph{geometric design parameters} from a sampled
surface point cloud, and (2) a \textbf{FiLM-conditioned MLP} that predicts pointwise aerodynamic coefficients from surface
coordinates/normals, conditioned on the geometry parameters and the flight conditions. 

\begin{itemize}
  \item \textbf{Stage 1: PointNet geometry-parameter inference.}
  The PointNet model is trained to predict nine geometric design parameters $\hat{\mathbf{p}}\in\mathbb{R}^9$ from sampled
  surface points. The authors sample 5{,}000 points on the surface and then randomly subsample into 15 batches of 2{,}048
  representative points (random sampling for efficiency). 

  Mathematically, for a batch point cloud $X\in\mathbb{R}^{2048\times 3}$, a shared MLP maps points to features, global
  max-pooling aggregates to a latent vector $\mathbf{z}\in\mathbb{R}^{128}$, and a regression head outputs
  $\hat{\mathbf{p}}\in\mathbb{R}^9$. 

  \item \textbf{Stage 2: FiLM network for pointwise aerodynamics.}
  The FiLM model predicts $\mathbf{u}(\mathbf{x},\mathbf{n})=[C_p,\,C_{f_x},\,C_{f_z}]^\top$ using inputs
  $(\mathbf{x},\mathbf{n})$ along with geometric parameters $\mathbf{p}\in\mathbb{R}^9$ and flight conditions $\boldsymbol{\mu}$
  (e.g., Mach number, angle of attack, altitude, Reynolds length, etc.).
 
  FiLM implements layer-wise affine modulation of hidden activations:
  \[
    \eta_\ell(\mathbf{h}_\ell)=\boldsymbol{\gamma}_\ell\odot \mathbf{h}_\ell+\boldsymbol{\beta}_\ell,
  \]
  where $(\boldsymbol{\gamma}_\ell,\boldsymbol{\beta}_\ell)$ depend on $(\mathbf{p},\boldsymbol{\mu})$. A separate
  hypernetwork $h_\psi$ predicts these modulation parameters for each layer:
  \[
    (\boldsymbol{\gamma}_\ell,\boldsymbol{\beta}_\ell)=h_\psi(\mathbf{p},\boldsymbol{\mu}).
  \]

  \item \textbf{Two usage modes (important for ShapeEvolve integration).}
  The authors note the surrogate can operate in two modes: if the geometric parameters are known, the user can input them
  directly into the FiLM network; otherwise, PointNet is used first to predict $\hat{\mathbf{p}}$ from the surface point cloud,
  and FiLM is conditioned on $\hat{\mathbf{p}}$.
\end{itemize}

\begin{figure}[h!]
    \centering
\includegraphics[width=0.7\linewidth]{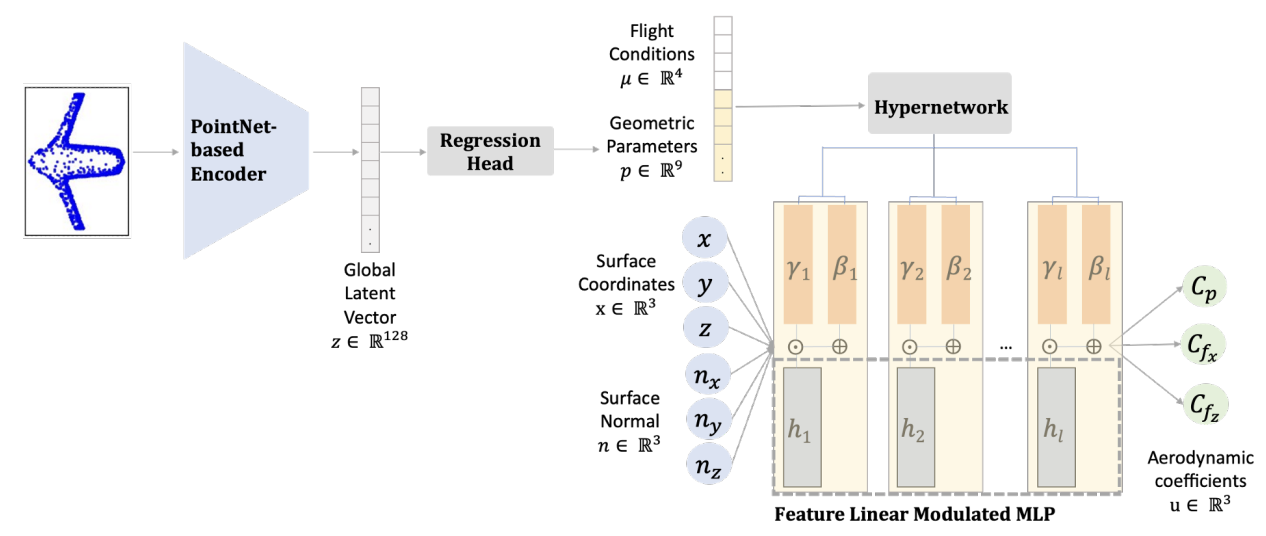}
    \caption{ Overall architecture of the surrogate model. The network consists of two main components: (1) a PointNetbased encoder that maps sampled surface points to global geometric design parameters, and (2) a FiLM-based network
that uses these parameters along with flight conditions to predict pointwise aerodynamic surface properties such as
pressure and friction coefficients. Figure is from \citep{Sung_2025}}
    \label{fig:blendednet}
\end{figure}

\begin{figure}
    \centering
    \includegraphics[width=0.7\linewidth]{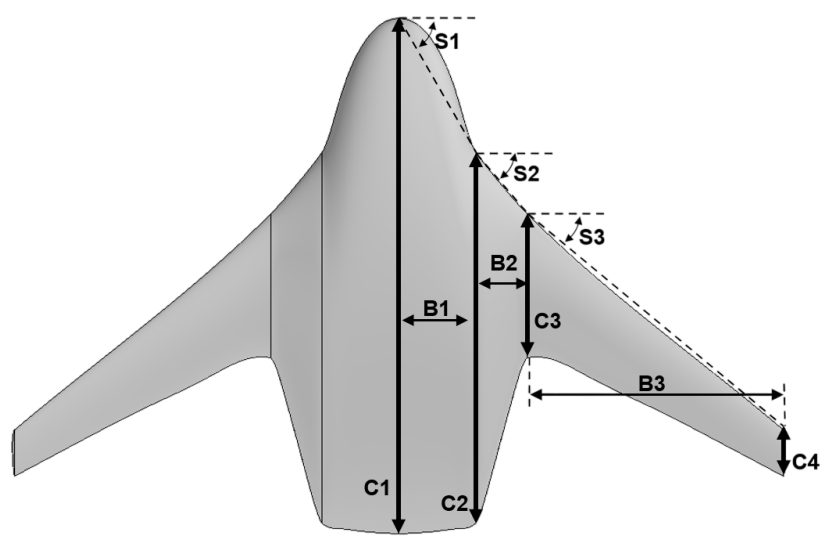}
    \caption{ Illustration of the blended wing body (BWB) planform parameterization, showing key geometric design
parameters used in dataset generation. Figure is from \citep{Sung_2025}}
    \label{fig:bwb}
\end{figure}

\paragraph{BWB geometry parameterization (design variables and ranges).}
BWB geometries are generated in OpenVSP using a planform-focused parameterization. Parameter definitions are
illustrated in Figure \ref{fig:bwb}. The dataset samples nine planform design parameters using Latin
Hypercube Sampling to generate 999 unique geometries. Airfoil cross-sections are parameterized with a degree-4 CST
representation but held constant in this study (to reduce dimensionality). 

The nine design parameters and their ranges (normalized by the centerline length) are: 
\begin{itemize}
  \item Relative chord-length parameters: $C_2/C_1\in[0.55,0.85]$, $C_3/C_1\in[0.18,0.28]$, $C_4/C_1\in[-0.06,0.09]$
  \item Relative spanwise-width parameters: $B_1/C_1\in[0.10,0.20]$, $B_2/C_1\in[0.05,0.20]$, $B_3/C_1\in[0.20,0.70]$
  \item Sweep angles: $S_1\in[40^\circ,60^\circ]$, $S_2\in[40^\circ,60^\circ]$, $S_3\in[24^\circ,40^\circ]$
\end{itemize}

\paragraph{Operating flight conditions (input parameters to the surrogate).}
Flight conditions are sampled using Latin Hypercube Sampling over altitude, Mach number, angle of attack, and Reynolds
length (the Reynolds number is computed during post-processing, rather than directly sampled). 

The flight condition ranges are: 
\begin{itemize}
  \item Altitude $\in [0,40]$ kft
  \item Mach number $\in [0.05,0.5]$
  \item Reynolds length $\in [0.1,10]$ m
  \item Angle of attack $\alpha\in[-10^\circ,20^\circ]$
\end{itemize}
Note that the surrogate is intended to accept any flight condition within these ranges, so we can include more than their $\sim 9$ conditions (per geometry) in our ShapeBench.

\subsection{\texttt{DrivAerStar} with Transolver} \label{app: drivaerstar}

\texttt{DrivAerStar} \citep{qiu2025drivaerstar} is a dataset of $12,000$ Star-CCM+ vehicle geometry simulations for aerodynamic optimization, totalling $20$ TB of data. Three body styles/configurations are included in the database: Estateback (E), Fastback (F), and Notchback (N), which are parameterized via Blender lattice/FFD. $20$ parametric deformation parameters are available, which are varied via Latin Hypercube Sampling, with FFD used to geometrically morph the vehicle shape as a function of the input parameter values. \texttt{DrivAerStar}'s advantages over predecessor datasets such as DrivAerNet++ \citep{elrefaie2025drivaernetlargescalemultimodalcar} and DrivAerML \citep{ashton2024drivaerml} are the inclusion of vehicle features such as engine bays, cooling systems, and internal airflow as well as greater wind tunnel validation accuracy errors of $\sim 1$\% compared to the $>5$\% typical values for the previous studies.

Transolver \citep{wu2024transolver} is a neural operator architecture and one of the models used within \texttt{DrivAerStar}. In validation tests, the Transolver performs best among the tested methods and also shows favorable/consistent scaling behavior when increasing the number of samples. For ShapeEvolve, the aerodynamic drag was evaluated using a pre-trained Transolver surrogate model provided by \citet{qiu2025drivaerstar}, trained on $1,200$ \texttt{DrivAerStar} simulation results. The model was then used in inference mode with frozen weights throughout all optimization runs. \texttt{DrivAerStar} provides a trained Transolver checkpoint at epoch 490 (out of 500), which ShapeEvolve directly uses for surrogate evaluation without retraining. Given the vehicle surface geometry as input, the surrogate model consisting of Transolver trained on \texttt{DrivAerStar} outputs predictions for surface pressure and the $3$ components of the wall shear stress, which then allows for the calculation of derived quantities (notably, drag). For the optimization pipeline in ShapeEvolve, each candidate design generated during optimization follows the full \texttt{DrivAerStar} pipeline steps:
\begin{enumerate}
    \item FFD-based geometric morphing in Blender
    \item VTK mesh extraction
    \item Surrogate evaluation with Transolver
\end{enumerate}
The surrogate achieves a total mean absolute percentage error (MAPE) of $2.422$\%, with per-style MAPEs of $2.633$\% for E, $2.195$\% for F, and $2.437$\% for N.

\subsection{\texttt{NeuralFoil}} \label{app: NeuralFoil}

\texttt{NeuralFoil} \citep{sharpe2025neuralfoilairfoilaerodynamicsanalysis, aerosandbox_phd_thesis}, when combined with the extension \texttt{AeroSandbox}, is a surrogate tool for rapid analysis of airfoils that can provide the aerodynamics for a wide range of airfoils and conditions. It is trained on tens of millions of \texttt{XFOIL} case runs. Beneficial features of \texttt{NeuralFoil} + \texttt{AeroSandbox} include guaranteed convergence, significant computational speedup compared to \texttt{XFOIL}, $C^\infty$ -continuity, fine-grained boundary layer control, and an assessment parameter of confidence/trustworthiness in the solution.

\subsection{\texttt{SuperWing}} \label{app: superwing}
 SuperWing dataset \citep{yang2025superwingcomprehensivetransonicwing} defines a parameterized design space for three-dimensional transonic wings. Each wing is generated from a Class-Shape Transformation (CST)-based~\citep{yang2025superwingcomprehensivetransonicwing} airfoil representation, where thickness and camber distributions are modulated along the span via scalar functions. The global planform follows a kinked (``Yehudi break'') trapezoidal wing configuration, parameterized by sweep angle, aspect ratio, taper ratio, kink location, and root adjustment (See Figure~\ref{fig:placeholder}). To capture realistic three-dimensional effects, spanwise variations of dihedral, twist, thickness, and camber are defined using spline interpolation over a small number of control points. In total, each geometry is described by CST coefficients and approximately 18 additional parameters, resulting in 4,239 unique wing configurations.

\begin{figure}[h!]
    \centering
    \includegraphics[width=0.75\linewidth]{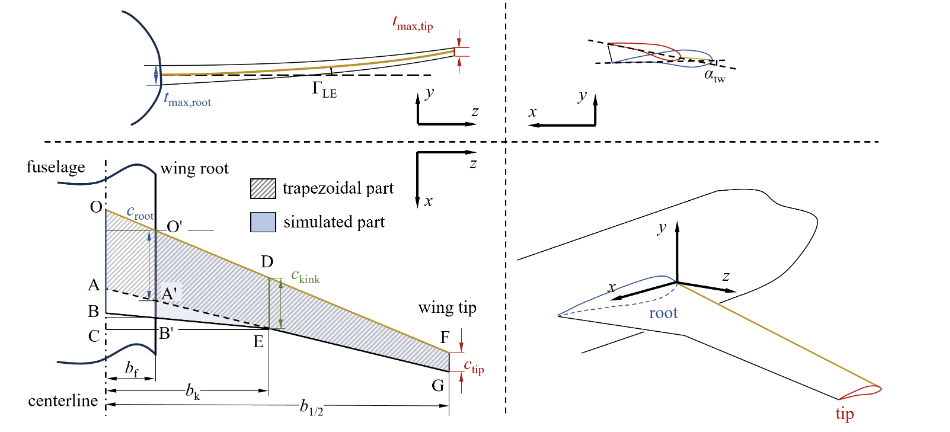}
    \caption{Three-view diagram of a typical kink wing. Figure is from \cite{yang2025superwingcomprehensivetransonicwing} }
    \label{fig:placeholder}
\end{figure}

Operating conditions are sampled independently for each geometry, with Mach number ranging from 0.75 to 0.90 and angle of attack from $2^\circ$ to $12^\circ$ while fixing the Reynolds number to be $2\times10^7$, yielding 28,856 RANS flow-field solutions.

The corresponding surrogate model maps geometry and operating conditions to surface flow fields. Geometry is represented on a structured reference mesh of size $256\times128$, where each point contains spatial coordinates, enabling a consistent grid-based representation across all wing shapes. The model predicts surface pressure and skin-friction coefficients, from which aerodynamic coefficients ($C_L$, $C_D$) are computed via surface integration.
The model training is performed using supervised regression with mean squared error (MSE) loss. Three model architectures are provided, including U-Net, Vision Transformers (ViT)~\citep{dosovitskiy2020imagetransolver}, and Transolver~\citep{wu2024transolver}. The dataset and the provided benchmark provide complex aerodynamic features within a rich design space for downstream ASO tasks.

\subsection{\texttt{CERASNet}} \label{app: CERAS Surrogate}
We train CERAS surrogate on a dataset of 2,981 FAST-OAD evaluations of parametric designs for the Central Reference Aircraft System $\cite{Saves_2022}$. The dataset is formed of 3,000 Latin Hypercube Sampled designs; 2,981 successful FAST-OAD MDA runs. The surruagte consists of a four-layer neural network with 10 encoded design variables mapping to all six simulation outputs $\hat{\mathbf{y}} = [\text{fuel mass},\ \text{static margin},\ \text{MTOW},\ \text{OWE},\ \text{specific fuel},\ \text{cruise fuel}]^\top$.

\begin{table}[ht]
\centering
\caption{CERAS surrogate model accuracy on validation set (20\% of dataset).}
\label{tab:surrogate_accuracy}
\begin{tabular}{lrrrrc}
\toprule
Output & RMSE & MAE & nRMSE (\%) & $r$ & Unit \\
\midrule
Fuel mass & 137.6 & 100.4 & 1.00 & 0.9993 & kg \\
Static margin & 0.0092 & 0.0069 & 0.63 & 0.9998 & -- \\
MTOW & 220.9 & 170.1 & 0.70 & 0.9997 & kg \\
OWE & 130.3 & 101.4 & 0.58 & 0.9997 & kg \\
Specific fuel & 3.25e-06 & 2.36e-06 & 1.00 & 0.9993 & kg/N/s \\
Cruise fuel & 174.1 & 121.8 & 1.59 & 0.9965 & kg \\
\bottomrule
\end{tabular}
\end{table}

\subsection{\texttt{STANet}} \label{app: STA Surrogate}
Supersonic Transport Aircraft (STA) Surrogate is trained on our database of 2,932 valid pairs of flight conditions and designs sampled through LHS. The dataset is created with the Vortex Lattice Method (VLM) solver. The surrogate is based on a Transolver architecture $\cite{wu2024transolver}$ with inputs of design variable, flight conditions, and local surface normals and outputs of aerodynamic coefficients $\mathbf{y} = [C_L,\ C_{Di}]$.

\begin{figure}[h!]
    \centering
    \includegraphics[width=0.75\linewidth]{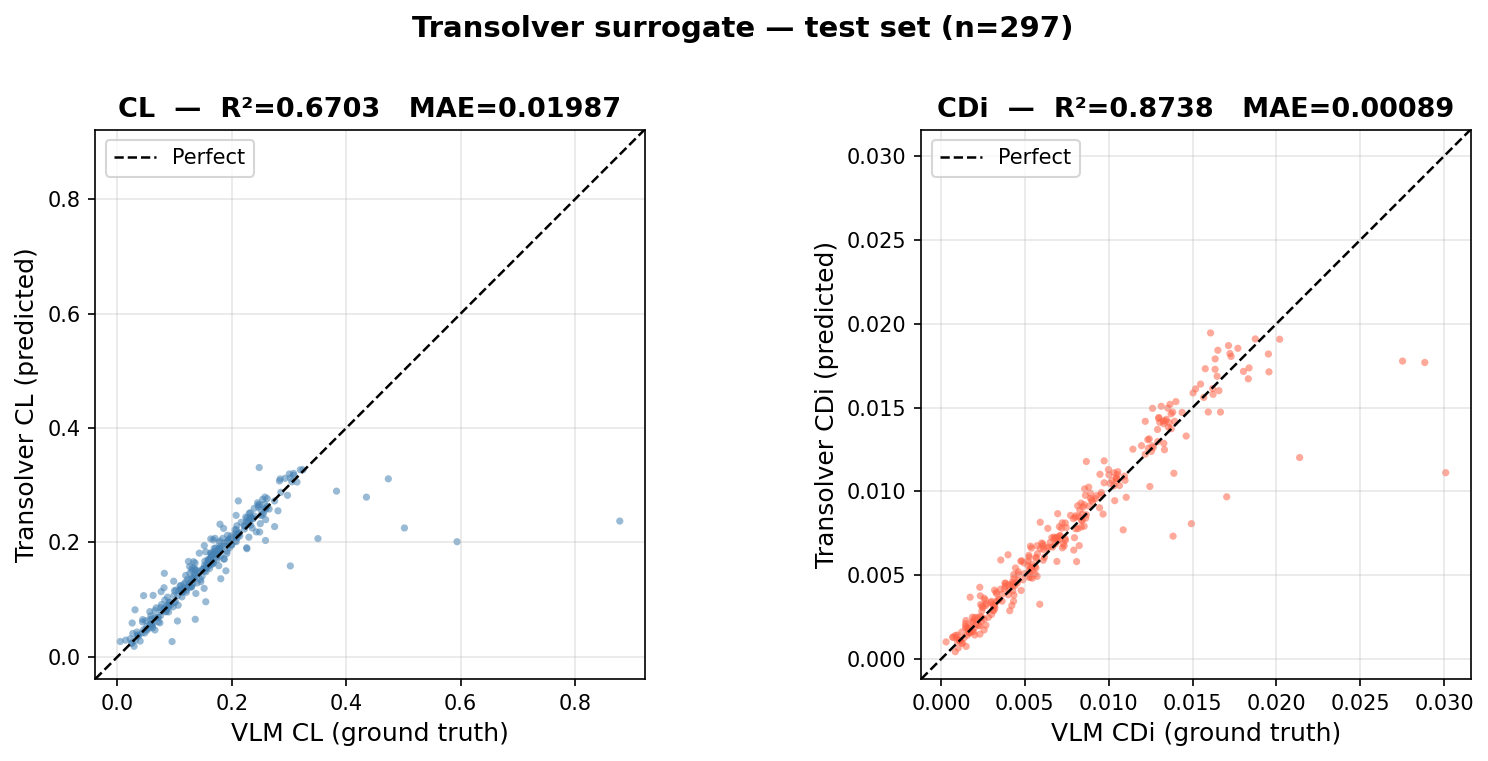}
    \caption{correlation of predicted aerodynamic coefficient CL and CD against VLM data for supersonic transport aircraft}
    \label{fig:placeholder}
\end{figure}

\subsection{\texttt{COCOANet} (CCA Surrogate)} \label{app: cca surrogate}
\texttt{COCOANet} (COllaborative COmbat Aircraft NETwork Surrogate) is our in-house surrogate model to accelerate design and analysis of Collaborative Combat Aircraft (CCA) for mission readiness. It is trained on a dataset of 3,570 high-fidelity CFD simulations. The data set contains 401 unique CCA geometries with up to 10 flight conditions corresponding to each design. The dataset is publicly available as well as a pretrained Transolver model \cite{wu2024transolver} using the \texttt{NeuralSolver} library \cite{thuml2025neuralsolver} following a similar adaption to blendednet++ \cite{sung2025blendednetlargescaleblendedwing}.

\begin{figure}[h!]
    \centering
    \includegraphics[width=0.5\linewidth]{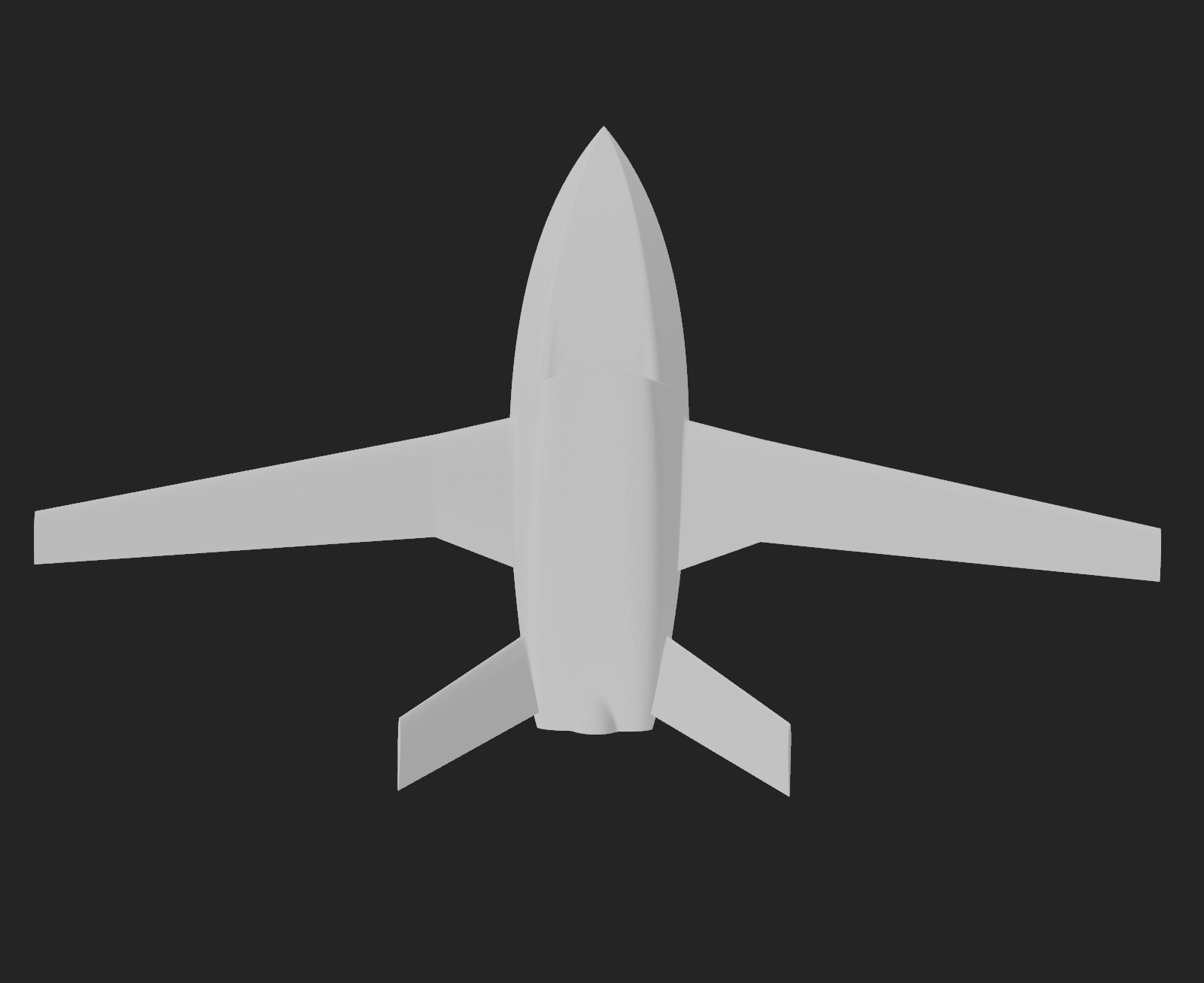}
    \caption{CCA: Collaborative Combat Aircraft}
    \label{fig:placeholder}
\end{figure}

\begin{table}[h!]
\centering
\begin{tabular}{|l|c|c|}
\hline
\textbf{Parameter} & \textbf{Lower Bound} & \textbf{Upper Bound} \\ \hline

Dihedral angle (deg) & 0.25 & 15 \\ \hline
Max Wing Blend (mm) & 25 & 1000 \\ \hline
Inlet Angle 1 (deg) & 0 & 45 \\ \hline
Inlet Angle 2 (deg) & 0 & 10 \\ \hline
Wing Position & 0.22 & 0.51 \\ \hline
Rear Point (mm) & (4500, 0, 0) & (7500, 0, 0) \\ \hline

$\vdots$ & $\vdots$ & $\vdots$ \\ \hline

Inlet Location & 0.2 & 0.6 \\ \hline
NACA 4-digit code & \multicolumn{2}{c|}{\{1412, 0012, 2408, 4412\}} \\ \hline
Fore Top Angle (deg) & 0 & 10 \\ \hline
Aft Top Angle (deg) & 12 & 32.5 \\ \hline
Top Height Aft (mm) & 36 & 220 \\ \hline
Bottom Height Aft (mm) & 38 & 208 \\ \hline
Wing span (mm) & 6500 & 20000 \\ \hline
Rear tail offset (mm) & 992 & 1770 \\ \hline
Root Chord (mm) & 1431 & 2700 \\ \hline
Tail Root chord (mm) & 800 & 1200 \\ \hline

\end{tabular}
\caption{CCA Drone Parameter Bounds}
\end{table}

Surface meshes are generated in ntop \cite{ntop2025}, and mesh healing is applied through meshlib \cite{meshlib2025}. Volume meshes are generated through flow360 with boundary layer growth rate of 1.2 and first boundary layer thickness of 17.87 $\mu m$. 

Simulation data is generated through \texttt{Flow360}, a GPU-accelerated solver for CFD. We apply farfield boundary conditions and run steady state RANS with spalart allmaras turbulence model. Flight conditions for mach, angle of attack, and altitude are applied corresponding to the case and each case is run for 3500 steps. 

\begin{figure}[h!]
    \centering
    \includegraphics[width=1\linewidth]{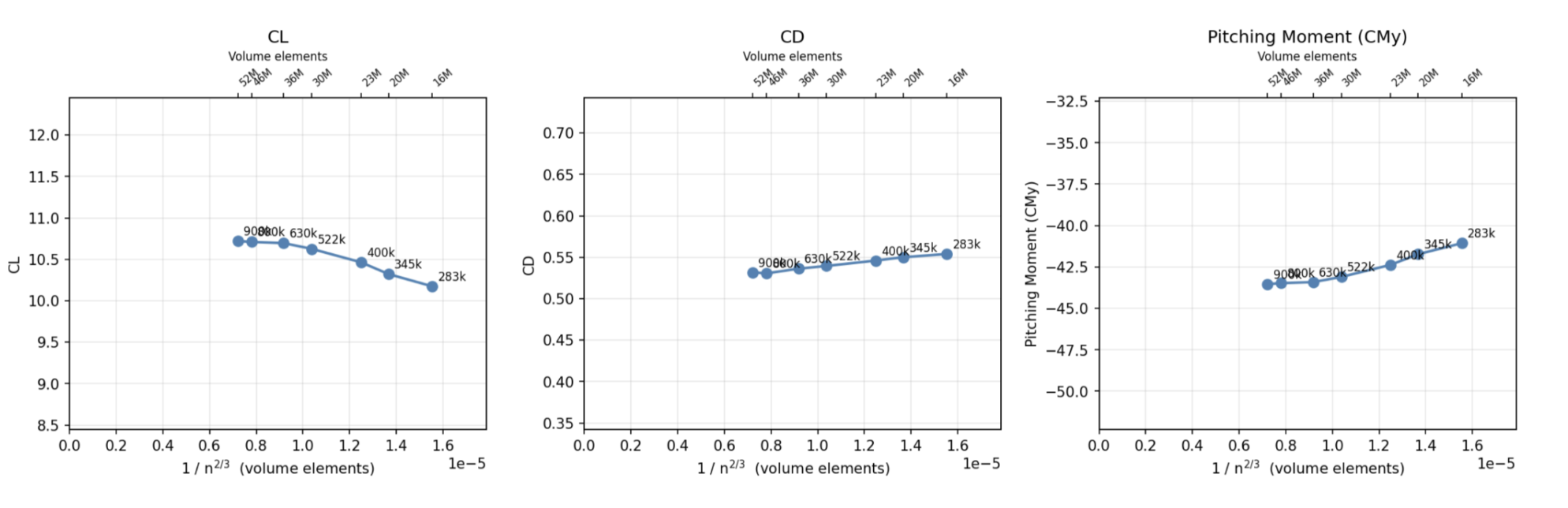}
    \caption{Grid convergence}
    \label{fig:placeholder}
\end{figure}

\begin{figure}[h!]
    \centering
    \includegraphics[width=1\linewidth]{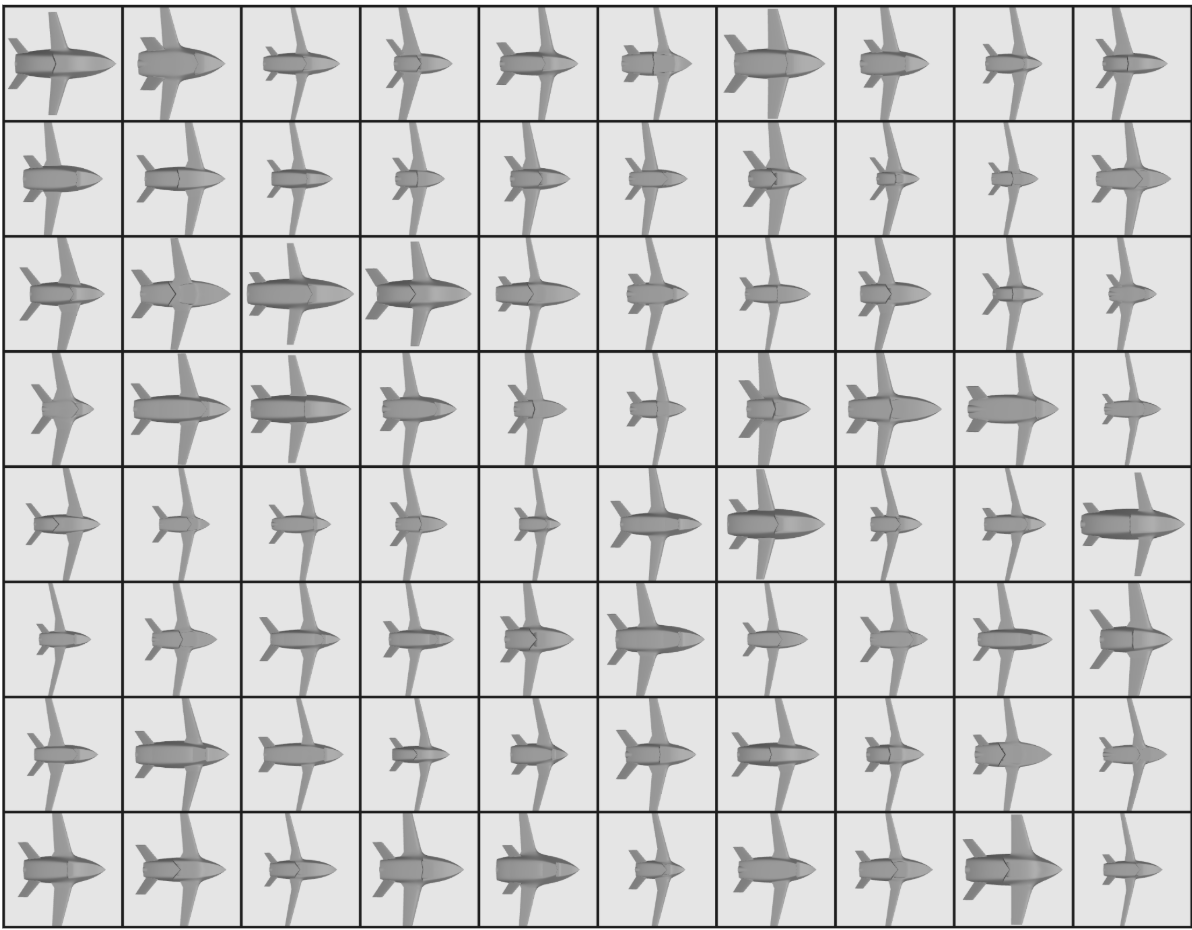}
    \caption{Sampled CCA designs representative of geometries in \texttt{COCOANet} CCA surrogate dataset. purple{Scale varies.}}
    \label{fig:placeholder}
\end{figure}

Dataset creation involves Latin hypercube sampling of both geometric parameters as well as flight conditions for a total of 3570 CFD configurations. CCA designs are generated parametrically using nTop \cite{ntop2025}. We define flight conditions with the following three variables: Mach (between $0.05$ and $0.5$), altitude (between $0$m and $10000$m), and angle of attack (AoA) (between $8^\circ$ and $10^\circ$). An $80$/$20$ training/testing split for the data was used.

\begin{figure}[h!]
    \centering
    \includegraphics[width=1\linewidth]{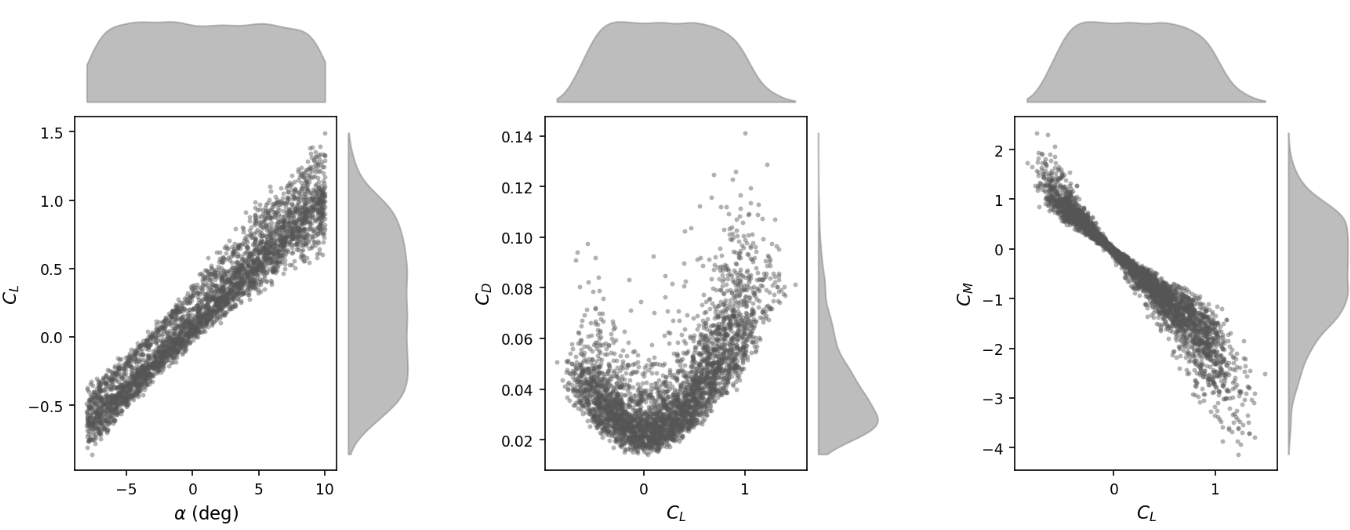}
    \caption{\texttt{COCOANet} CCA surrogate data characteristics}
    \label{fig: cca_dataset_characteristics}
\end{figure}

\begin{table}[h!]
\centering
\caption{Flight conditions for CFD cases.}
\begin{tabular}{lcc}
\hline
\textbf{Parameter} & \textbf{Lower Bound} & \textbf{Upper Bound} \\
\hline
Mach [-]        & 0.05     & 0.5  \\
Altitude [m]              & 0  & 10000 \\
Angle of Attack [deg] & -8   & 10  \\
\hline
\end{tabular}
\end{table}

\begin{figure}[h!]
    \centering
    \includegraphics[width=1\linewidth]{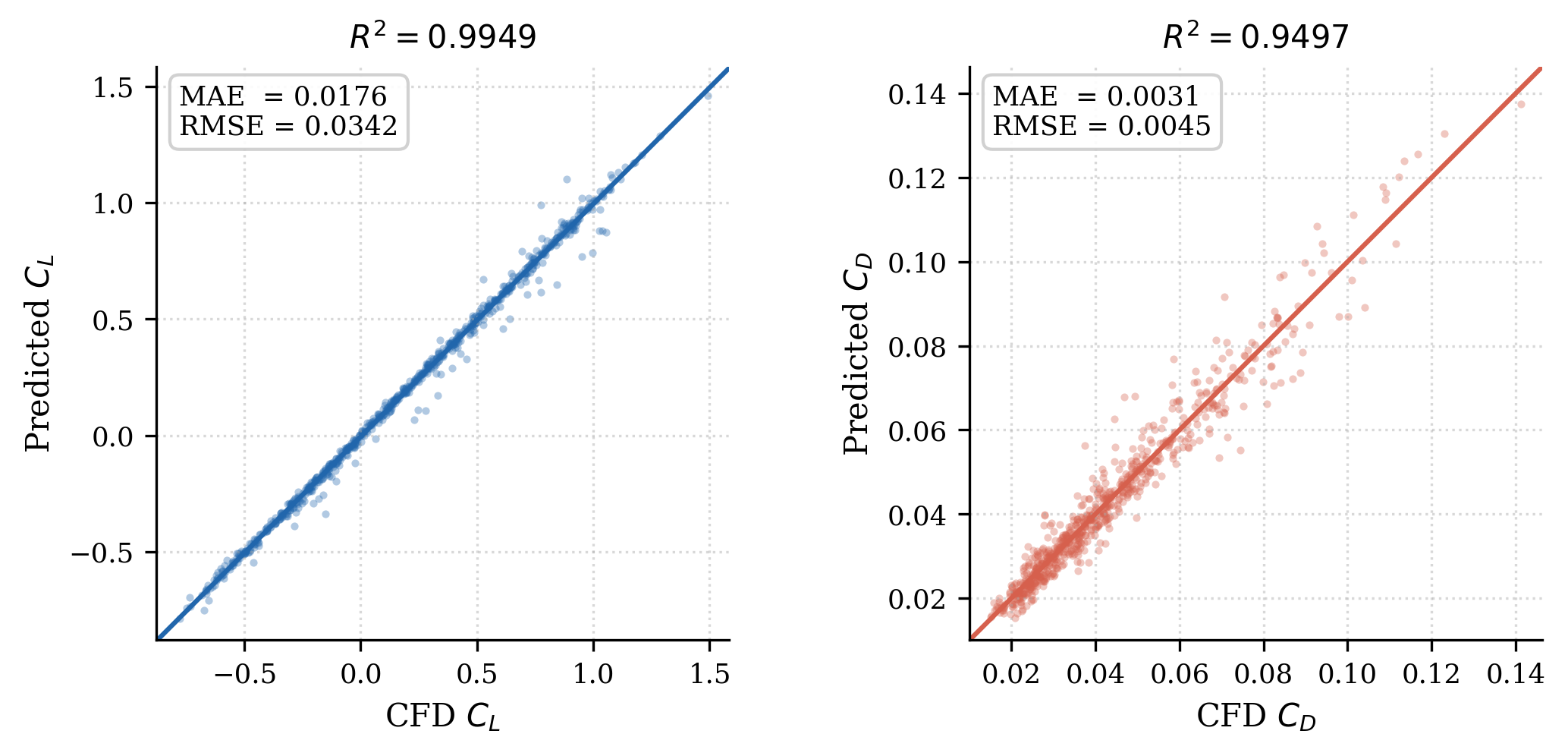}
    \caption{correlation of predicted aerodynamic coefficient CL and CD against CFD data}
    \label{fig:placeholder}
\end{figure}

\begin{table}[h!]
\centering
\caption{\texttt{COCOANet} CCA Surrogate Aerodynamic Coefficient Prediction Errors}
\begin{tabular}{lcccc}
\hline
\textbf{Metric} & \textbf{CfVecX} & \textbf{CfVecY} & \textbf{CfVecZ} & \textbf{Cp} \\
\hline
MSE        & 3.46e-07 & 1.45e-07 & 2.76e-07 & 8.12e-03 \\
MAE        & 2.42e-04 & 1.45e-04 & 1.72e-04 & 3.68e-02 \\
Rel L1 (\%) & 9.31     & 28.64    & 16.67    & 14.08    \\
Rel L2 (\%) & 17.26    & 30.95    & 16.61    & 16.27    \\
\hline
\end{tabular}
\end{table}

\clearpage

\section{ShapeBench Task Setups \& Additional Plots} \label{ShapeBench Task Setups and Additional Plots}

The following diagram breaks down the main components of ShapeBench. \\ 

\begin{forest}
for tree={
    font=\ttfamily,
    grow'=0,
    parent anchor=east,
    child anchor=west,
    anchor=west,
    base=middle,
    l sep=12pt,
    s sep=8pt,
    inner sep=1pt,
    edge={thin},
    edge path={
        \noexpand\path[\forestoption{edge}]
        (!u.parent anchor) -- +(8pt,0) |- (.child anchor)\forestoption{edge label};
    },
}
[ShapeBench/
    [Continuous/
        [3D Delta Wing Design/ {$\rightarrow$ VortexNet}
        ]
        [3D Blended Wing Body (BWB) Design/ {$\rightarrow$  BlendedNet}
        ]
        [2D Airfoil Design/ {$\rightarrow$  \texttt{NeuralFoil}}
        ]
        [3D Transonic Swept-wing Design/ {$\rightarrow$  SuperWing}
        ]
        [3D Car Design/ {$\rightarrow$  DrivAerStar}
        ]
        [3D Collaborative Combat Aircraft (CCA)/ {$\rightarrow$  COCOANet}
        ]
    ]
    [Mixed-variable/
        [Supersonic Transport Aircraft (STA)/ {$\rightarrow$  FAST-OAD}] 
        [Central Reference Aircraft System (CERAS)/ {$\rightarrow$  FAST-OAD}] 
    ]
]
\end{forest}

In this section, we discuss the full set of problem environments provided in ShapeBench, providing an example of each one with the corresponding numerical experimental results.

\subsection{3D Delta Wing Design} \label{app: 3D Delta Wing}

\paragraph{Design variables}

We use the 3D delta wing parametrization from \cite{shen2025vortexnet}. Let

\[
\mathbf{x} = (\Lambda_{\mathrm{LE}}, \psi_{\mathrm{root}}),
\]
where $\Lambda_{\mathrm{LE}}$ is the leading-edge sweep angle (deg), and
$\psi_{\mathrm{root}}$ is the root airfoil choice. The root chord is fixed at
\[
c_{\mathrm{root}} = 0.65~\mathrm{m}.
\]

\subsubsection{Single-objective Task}

\paragraph{Operating point}
The aerodynamic coefficients are evaluated at a fixed operating point
\[
(\alpha_0, M_0, \mathrm{Re}_0),
\]
chosen a priori within
\[
\alpha_0 \in [0^\circ,20^\circ],
\qquad
M_0 \in [0.35,0.5],
\qquad
\mathrm{Re}_0 \in [6.5\times 10^6,\,10^7].
\]

\paragraph{Objective}
Maximize the lift-to-drag ratio
\[
\max_{\mathbf{x}} \frac{C_L(\Lambda_{\mathrm{LE}},\psi_{\mathrm{root}};\alpha_0,M_0,\mathrm{Re}_0)}
{C_D(\Lambda_{\mathrm{LE}},\psi_{\mathrm{root}};\alpha_0,M_0,\mathrm{Re}_0)},
\]
where $C_L(\cdot)$ and $C_D(\cdot)$ denote the lift and drag coefficients
for the delta-wing geometry at the fixed operating point above.

\paragraph{Constraints}
The design is subject to
\[
55^\circ \le \Lambda_{\mathrm{LE}} \le 75^\circ,
\]
\[
c_{\mathrm{root}} = 0.65~\mathrm{m},
\]
\[
\psi_{\mathrm{root}} \in \mathcal{A},
\qquad
\mathcal{A} =
\{\mathrm{NACA0010},\mathrm{NACA0016},\mathrm{NACA0024},
\mathrm{NACA2416},\mathrm{NACA4416}\}.
\]

\begin{figure}[h!]
    \centering
    \includegraphics[width=0.75\linewidth]{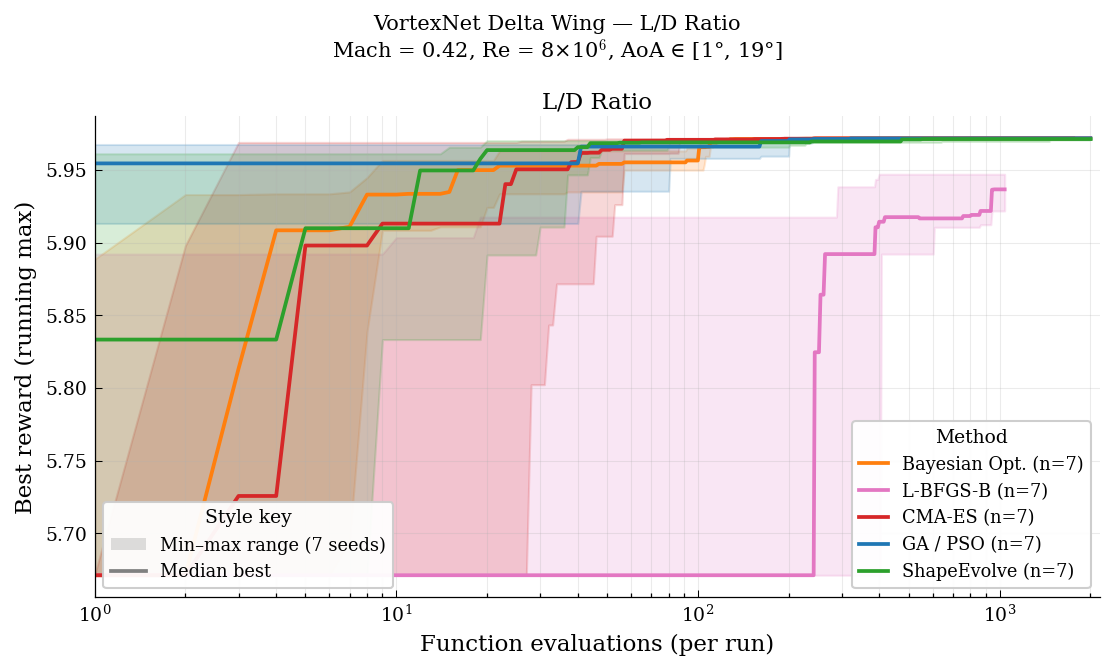}
    \caption{Single point lift-to-drag optimization results}
    \label{fig:placeholder}
\end{figure}

\begin{figure}[h!]
    \centering
    \includegraphics[width=1\linewidth]{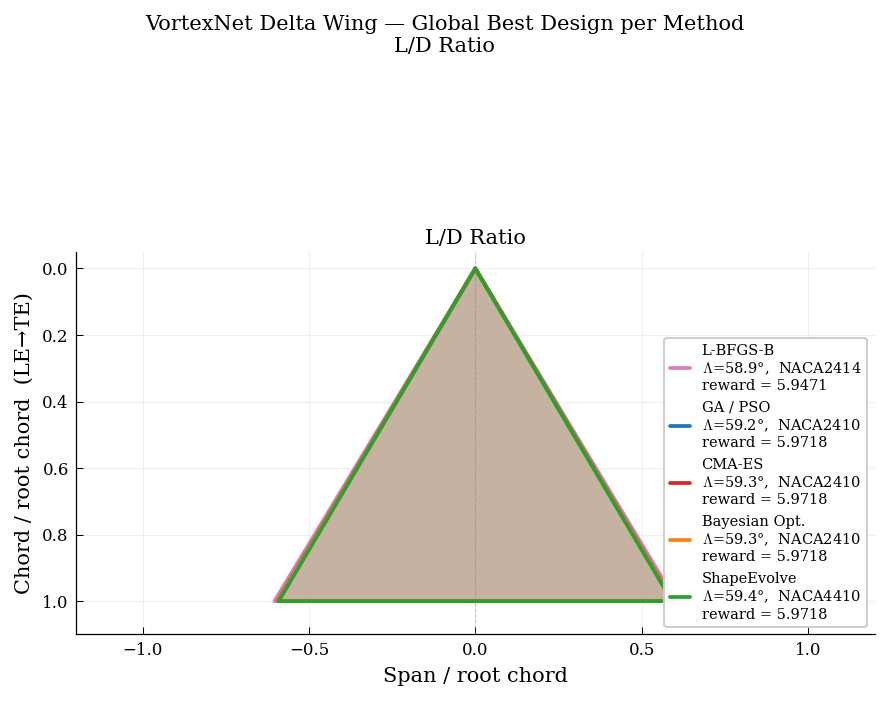}
    \caption{Best shape per method for lift-to-drag delta wing task}
    \label{fig:placeholder}
\end{figure}

\subsubsection{Multi-point Task}

\paragraph{\textcolor{blue}{Option A --- Mission objective (weighted-sum multi-point)}}

\paragraph{Operating points}
The aerodynamic coefficients are evaluated at $K$ prescribed operating points
\[
(\alpha_k, M_k, \mathrm{Re}_k), \qquad k=1,\ldots,K,
\]
chosen within
\[
\alpha_k \in [0^\circ,20^\circ],
\qquad
M_k \in [0.35,0.5],
\qquad
\mathrm{Re}_k \in [6.5\times 10^6,\,10^7].
\]

\begin{figure}[h!]
    \centering
    \includegraphics[width=0.75\linewidth]{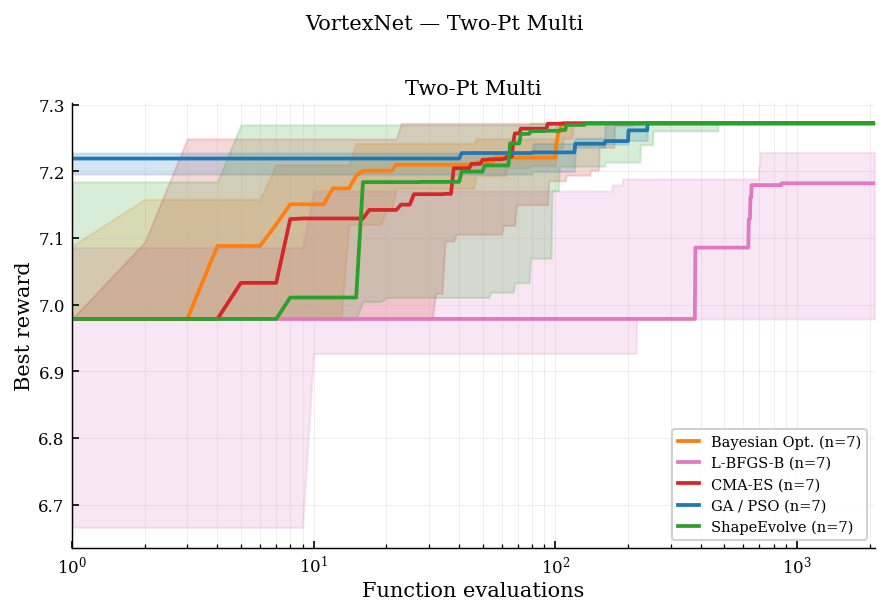}
    \caption{Convergence trajectories for optimizers for two point Vortnet task}
    \label{fig:placeholder}
\end{figure}

\begin{figure} [h!]
    \centering
    \includegraphics[width=0.75\linewidth]{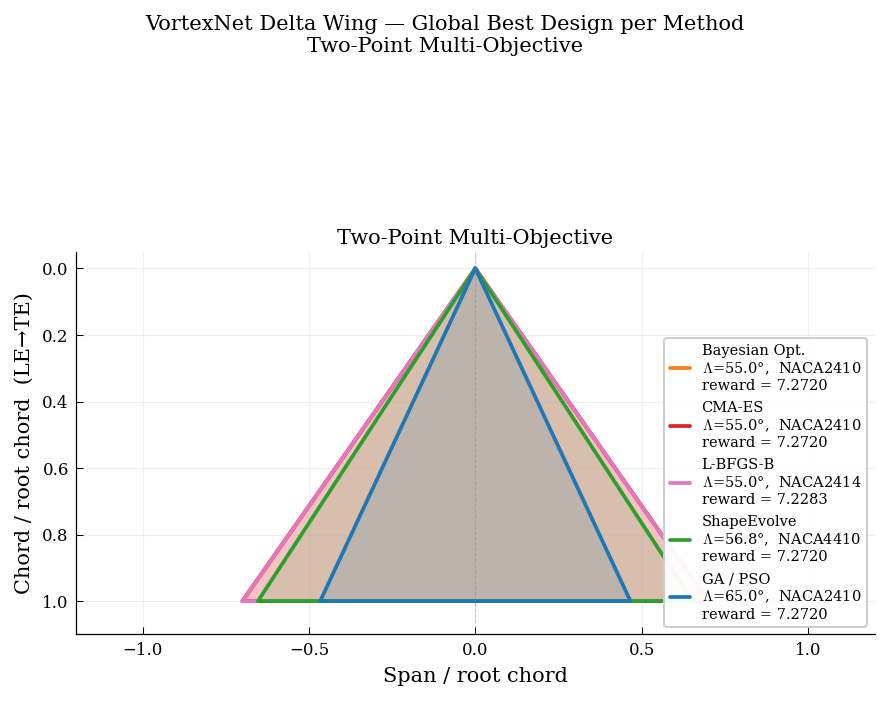}
    \caption{Best Design Overlay of Delta Wingdesign for two point objective}
    \label{fig:placeholder}
\end{figure}

\paragraph{Objective}
Maximize the weighted mission performance
\[
\max_{\mathbf{x}} J_{\mathrm{mission}}(\mathbf{x})
=
\max_{\mathbf{x}}
\sum_{k=1}^{K} w_k \,
\frac{C_{L,k}(\mathbf{x})}{C_{D,k}(\mathbf{x})},
\]
where
\[
C_{L,k}(\mathbf{x}) := C_L(\mathbf{x};\alpha_k,M_k,\mathrm{Re}_k),
\qquad
C_{D,k}(\mathbf{x}) := C_D(\mathbf{x};\alpha_k,M_k,\mathrm{Re}_k),
\]
and the weights satisfy
\[
w_k \ge 0,
\qquad
\sum_{k=1}^{K} w_k = 1.
\]

\paragraph{Constraints}
\[
\Lambda_{\mathrm{LE}} \in \{55^\circ,\,65^\circ,\,75^\circ\},
\]
\[
\psi_{\mathrm{root}} \in \mathcal{A},
\qquad
\mathcal{A} =
\{\mathrm{NACA0010},\mathrm{NACA0016},\mathrm{NACA0024},
\mathrm{NACA2416},\mathrm{NACA4416}\}.
\]

\paragraph{\textcolor{blue}{Option B --- Robust objective (maximize worst-case multi-point)}}

\paragraph{Operating points}
The aerodynamic coefficients are evaluated at $K$ prescribed operating points
\[
(\alpha_k, M_k, \mathrm{Re}_k), \qquad k=1,\ldots,K,
\]
chosen within
\[
\alpha_k \in [0^\circ,20^\circ],
\qquad
M_k \in [0.35,0.5],
\qquad
\mathrm{Re}_k \in [6.5\times 10^6,\,10^7].
\]

\paragraph{Objective}
Maximize the worst-case lift-to-drag ratio across the prescribed operating
points:
\[
\max_{\mathbf{x}} J_{\mathrm{robust}}(\mathbf{x})
=
\max_{\mathbf{x}}
\min_{k=1,\ldots,K}
\frac{C_{L,k}(\mathbf{x})}{C_{D,k}(\mathbf{x})},
\]
where
\[
C_{L,k}(\mathbf{x}) := C_L(\mathbf{x};\alpha_k,M_k,\mathrm{Re}_k),
\qquad
C_{D,k}(\mathbf{x}) := C_D(\mathbf{x};\alpha_k,M_k,\mathrm{Re}_k).
\]

\paragraph{Constraints}
\[
\Lambda_{\mathrm{LE}} \in \{55^\circ,\,65^\circ,\,75^\circ\},
\]
\[
\psi_{\mathrm{root}} \in \mathcal{A},
\qquad
\mathcal{A} =
\{\mathrm{NACA0010},\mathrm{NACA0016},\mathrm{NACA0024},
\mathrm{NACA2416},\mathrm{NACA4416}\}.
\]
\subsubsection{Multi-objective Task}

\paragraph{Operating point (held fixed)}
The aerodynamic coefficients are evaluated at a fixed operating point

\[
(\alpha_0, M_0, \mathrm{Re}_0),
\]
chosen a priori within
\[
\alpha_0 \in [0^\circ,20^\circ],
\qquad
M_0 \in [0.3, 0.42,0.5],
\qquad
\mathrm{Re}_0 \in [6.5\times 10^6, 8.9 \times 10^6 \, 10^7].
\]

\paragraph{Objectives} Maximize longitudinal static margin while minimizing drag. 

\[
\max_{\mathbf{x}} \mathbf{f}(\mathbf{x})
=
\max_{\mathbf{x}}
\begin{pmatrix}
f_1(\mathbf{x})\\[4pt]
f_2(\mathbf{x})
\end{pmatrix}
=
\max_{\mathbf{x}}
\begin{pmatrix}
{K_n(\mathbf{x};\alpha_0,M_0,\mathrm{Re}_0)}
       \\[10pt]
-w_{c_D} \cdot C_D(x;a_o, M_0, Re_0
\end{pmatrix},
\]

where \[
K_n(\mathbf{x})
=
-\,\frac{C_M(\mathbf{x};\,\alpha_0+\delta\alpha)-C_M(\mathbf{x};\,\alpha_0)}
        {C_L(\mathbf{x};\,\alpha_0+\delta\alpha)-C_L(\mathbf{x};\,\alpha_0)}
\]

\paragraph{\textcolor{blue}{Option A --- 2 objectives, single-point}}

\paragraph{Operating point (held fixed)}
The aerodynamic coefficients are evaluated at a fixed operating point
\[
(\alpha_0, M_0, \mathrm{Re}_0),
\]
chosen a priori within
\[
\alpha_0 \in [0^\circ,20^\circ],
\qquad
M_0 \in [0.35,0.5],
\qquad
\mathrm{Re}_0 \in [6.5\times 10^6,\,10^7].
\]

\paragraph{Objectives}
Minimize simultaneously the negative lift-to-drag ratio and the trim penalty:
\[
\min_{\mathbf{x}} \mathbf{f}(\mathbf{x})
=
\min_{\mathbf{x}}
\begin{pmatrix}
f_1(\mathbf{x})\\[4pt]
f_2(\mathbf{x})
\end{pmatrix}
=
\min_{\mathbf{x}}
\begin{pmatrix}
-\dfrac{C_L(\mathbf{x};\alpha_0,M_0,\mathrm{Re}_0)}
       {C_D(\mathbf{x};\alpha_0,M_0,\mathrm{Re}_0)}\\[10pt]
\left|C_M(\mathbf{x};\alpha_0,M_0,\mathrm{Re}_0)\right|
\end{pmatrix},
\]
interpreted in the Pareto sense with respect to the usual componentwise
order on $\mathbb{R}^2$, where $C_L(\cdot)$, $C_D(\cdot)$, and $C_M(\cdot)$
denote the lift, drag, and pitching-moment coefficients at the fixed
operating point above.

\paragraph{Constraints}
\[
55^\circ \le \Lambda_{\mathrm{LE}} \le 75^\circ,
\]
\[
\psi_{\mathrm{root}} \in \mathcal{A},
\qquad
\mathcal{A} =
\{\mathrm{NACA0010},\mathrm{NACA0016},\mathrm{NACA0024},
\mathrm{NACA2416},\mathrm{NACA4416}\}.
\]

\paragraph{\textcolor{blue}{Option B --- 3 objectives, multi-point ``mission” version}}

\paragraph{Operating points}
The aerodynamic coefficients are evaluated at $K$ prescribed operating points
\[
(\alpha_k, M_k, \mathrm{Re}_k), \qquad k=1,\ldots,K,
\]
chosen within
\[
\alpha_k \in [0^\circ,20^\circ],
\qquad
M_k \in [0.35,0.5],
\qquad
\mathrm{Re}_k \in [6.5\times 10^6,\,10^7].
\]
The mission weights satisfy
\[
w_k \ge 0,
\qquad
\sum_{k=1}^{K} w_k = 1.
\]

\paragraph{Objectives}
Minimize simultaneously the negative weighted lift-to-drag ratio, the
weighted drag coefficient, and the weighted trim penalty:
\[
\min_{\mathbf{x}} \mathbf{f}(\mathbf{x})
=
\min_{\mathbf{x}}
\begin{pmatrix}
f_1(\mathbf{x})\\[4pt]
f_2(\mathbf{x})\\[4pt]
f_3(\mathbf{x})
\end{pmatrix}
=
\min_{\mathbf{x}}
\begin{pmatrix}
-\displaystyle\sum_{k=1}^{K} w_k \,
\frac{C_{L,k}(\mathbf{x})}{C_{D,k}(\mathbf{x})}\\[12pt]
\displaystyle\sum_{k=1}^{K} w_k \, C_{D,k}(\mathbf{x})\\[10pt]
\displaystyle\sum_{k=1}^{K} w_k \,
\left|C_{M,k}(\mathbf{x})\right|
\end{pmatrix},
\]
interpreted in the Pareto sense with respect to the usual component-wise
order on $\mathbb{R}^3$, where
\[
C_{L,k}(\mathbf{x}) := C_L(\mathbf{x};\alpha_k,M_k,\mathrm{Re}_k),
\qquad
C_{D,k}(\mathbf{x}) := C_D(\mathbf{x};\alpha_k,M_k,\mathrm{Re}_k),
\]
\[
C_{M,k}(\mathbf{x}) := C_M(\mathbf{x};\alpha_k,M_k,\mathrm{Re}_k).
\]

\paragraph{Constraints}
\[
55^\circ \le \Lambda_{\mathrm{LE}} \le 75^\circ,
\]
\[
\psi_{\mathrm{root}} \in \mathcal{A},
\qquad
\mathcal{A} =
\{\mathrm{NACA0010},\mathrm{NACA0016},\mathrm{NACA0024},
\mathrm{NACA2416},\mathrm{NACA4416}\}.
\]

\subsection{3D Blended Wing Body (BWB) Design}

\paragraph{Design variables} We use the 3D BWB parametrization from \cite{Sung_2025}. Let
\[
\mathbf{x}=\mathbf{p}
=
\left(
\frac{C_2}{C_1},
\frac{C_3}{C_1},
\frac{C_4}{C_1},
\frac{B_1}{C_1},
\frac{B_2}{C_1},
\frac{B_3}{C_1},
S_1,S_2,S_3
\right)^{\top}
\in \mathbb{R}^{9},
\]
where \(C_1\) is the centerline length used for normalization, \(C_i/C_1\) are
normalized chord-length parameters, \(B_i/C_1\) are normalized spanwise-width
parameters, and \(S_1,S_2,S_3\) are sweep angles (in degrees). Note that the angle of attack is not an independent design variable; for each operating point, it is determined by bisection to satisfy the lift target, as will be discussed shortly.

\subsubsection{Multi-point Task} \label{app: bwb-mp}

Two benchmark objectives are evaluated using BlendedNet; both tasks share the same geometric parameterization and operating conditions but differ in aerodynamic objective.

\paragraph{Operating points}
The aerodynamic coefficients are evaluated at \(K = 5\) prescribed operating points (lift-coefficient targets)
\[
\bigl(C_{L,1}^{\star},\ldots,C_{L,5}^{\star}\bigr)
    = \bigl(0.185,\;0.206,\;0.206,\;0.206,\;0.227\bigr)           
\]
at fixed free-stream conditions
\[
M_{\infty} = 0.3, \qquad \mathrm{Re} = 10^{7}.
\] 
For each operating point \(k\), the angle of attack \(\alpha_{k}^{\star}\) is found by bisection over \([-5^{\circ},\,12^{\circ}]\) (8 iterations), in order to satisfy
\[
C_{L}\!\left(\mathbf{p};\,M_{\infty},\mathrm{Re},\alpha_{k}^{\star}\right)
    = C_{L,k}^{\star},
\]
where \(C_{L}(\cdot)\) is the integrated lift coefficient output by BlendedNet.

\paragraph{Surrogate drag proxy}
BlendedNet predicts the pointwise streamwise skin friction coefficient $C_{fx}$ as its drag-related output field. The scalar drag metric used in the objectives here is the mean skin-friction coefficient over the surface,
\begin{equation}
    C_{fx,k}(\mathbf{x}) = \frac{1}{N}\sum_{i=1}^{N} C_{fx,i} \left(\mathbf{p};\,M_{\infty},\mathrm{Re},\alpha_{k}^{\star}(\mathbf{p})\right),
\end{equation}
with the summation being over the $N$ sampled surface points. This quantity serves as a surrogate proxy for drag; note that it is not equivalent to the aerodynamic drag coefficient ($C_D$), which also includes the area-weighted pressure-drag contribution.

\paragraph{Task 1: Friction-drag minimization}
Across the $K = 5$ operating points, the skin-friction proxy is minimized, i.e.
\[
 \min_{\mathbf{x}}\;\overline{C}_{fx}(\mathbf{x}),
 \qquad
\overline{C}_{fx} (\mathbf{x})
\;=\;
\frac{1}{K}\sum_{k=1}^{K} C_{fx,k}(\mathbf{x}),
\]
where \(C_{fx,k}(\mathbf{x}) := C_{fx}\!\left(\mathbf{p};\,M_{\infty}, \mathrm{Re},\alpha_{k}^{\star}(\mathbf{p})\right)\) is the skin friction coefficient at the \(k\)-th operating point as evaluated at the bisection-solved \(\alpha_{k}^{\star}(\mathbf{p})\).

\paragraph{Task 2: Lift-to-friction-proxy maximization}
Across the $5$ operating points, the mean lift-to-skin-friction-proxy ratio is maximized
\[
\max_{\mathbf{x}}\;
\overline{L/D}_{\mathrm{proxy}} (\mathbf{x}),
\qquad
\overline{L/D}_{\mathrm{proxy}} (\mathbf{x})
\;=\;
\frac{1}{K}\sum_{k=1}^{K}
\frac{C_{L,k}^{\star}}{C_{fx,k}(\mathbf{x})}.
\]
Since $C_{L,k}^{\star}$ is fixed by construction, this objective can be thought of as a $C_L$-weighted mean of the inverse friction-proxy. This choice of objective weights operating points with larger $C_{L,k}^{\star}$ to have proportionally more influence. It also causes nonlinearity in $\overline{C}_{fx}$ with marginal sensitivity $\partial \overline{L/D}_{\mathrm{proxy}}/\partial \overline{C}_{fx} = -C_{L,k}^\star / \overline{C}_{fx}^2$. The denominator captures only the friction component of drag, and so while this ratio does not equal the true aerodynamic lift-to-drag, it still serves as a well-defined and physically-motivated surrogate objective.

\paragraph{Constraints}
The design variables are bounded by 
\[
0.55 \le \frac{C_2}{C_1} \le 0.85,
\qquad
0.18 \le \frac{C_3}{C_1} \le 0.28,
\qquad
0.06 \le \frac{C_4}{C_1} \le 0.09,
\]
\[
0.10 \le \frac{B_1}{C_1} \le 0.20,
\qquad
0.05 \le \frac{B_2}{C_1} \le 0.20,
\qquad
0.20 \le \frac{B_3}{C_1} \le 0.70,
\]
\[
40^\circ \le S_1 \le 60^\circ,
\qquad
40^\circ \le S_2 \le 60^\circ,
\qquad
24^\circ \le S_3 \le 40^\circ,
\qquad
-3^\circ \le \alpha \le 3^\circ.
\]
Note that the centerline chord \(C_1 = 1000\,\mathrm{mm}\) is fixed as a normalization anchor.

\paragraph{Surrogate drag proxy --- additional investigation}

As a reminder, these tasks introduced above use $\overline{C}_{fx}$ as a surrogate proxy for the drag as opposed to the true (total) integrated drag coefficient. To assess whether this proxy affects optimization outcomes and also for physical completeness, an additional study with the integrated drag metric
\begin{equation}
    C_D^{\mathrm{int}}(\mathbf{p}) \;=\; \frac{1}{S_{\mathrm{ref}}} \sum_{\text{cells}} \Bigl( C_{p,i}\,A_i\,n_{x,i} \;+\; C_{fx,i}\,A_i \Bigr), \label{eq:cd_int}
\end{equation}
was performed. Here, $A_i$ is the cell area, $n_{x,i}$ is the streamwise component of the outward face-normal vector, and $S_{\mathrm{ref}}$ is the planform-projected reference area. $C_{p}$ is the pressure coefficient at cell $i$. This additional study uses minimizing $\overline{C}_D^{\mathrm{int}} = (1/K) \sum_k C_{D,k}^{\mathrm{int}}$ as the objective in the multi-point task. A constrained variant with an aspect-ratio floor $AR \geq 2.5$ was also evaluated. Findings from this investigation are discussed at the end of the Results section.

\paragraph{Results}

\textit{Computational cost per $1{,}000$ evaluations per run} ($n$ = independent runs); $1$ core unless otherwise noted: 

Task 1 (min-$\overline{C}_{fx}$):
\begin{itemize}
    \item L-BFGS-B ($n = 10$): $10$ -- $24$ CPU-hours
    \item Bayesian Opt ($n = 10$): $15$ -- $16$ CPU-hours
    \item PSO ($n = 10$): $14$ -- $15$ CPU-hours
    \item CMA-ES ($n = 10$): $8$ -- $14$ CPU-hours
    \item ShapeEvolve ($n = 10$): $1.4$ -- $12$ CPU-hours
\end{itemize}
Task 2 (max-$\overline{L/D}_{\mathrm{proxy}}$):
\begin{itemize}
    \item Bayesian Opt ($n = 10$): $10$ -- $20$ CPU-hours
    \item PSO ($n = 10$): $13$ -- $24$ CPU-hours
    \item CMA-ES ($n = 10$): $\approx 14$ CPU-hours
    \item ShapeEvolve ($n = 10$): $12$ -- $13$ CPU-hours
\end{itemize}

\begin{figure}
    \centering
    \includegraphics[width=0.95\linewidth]{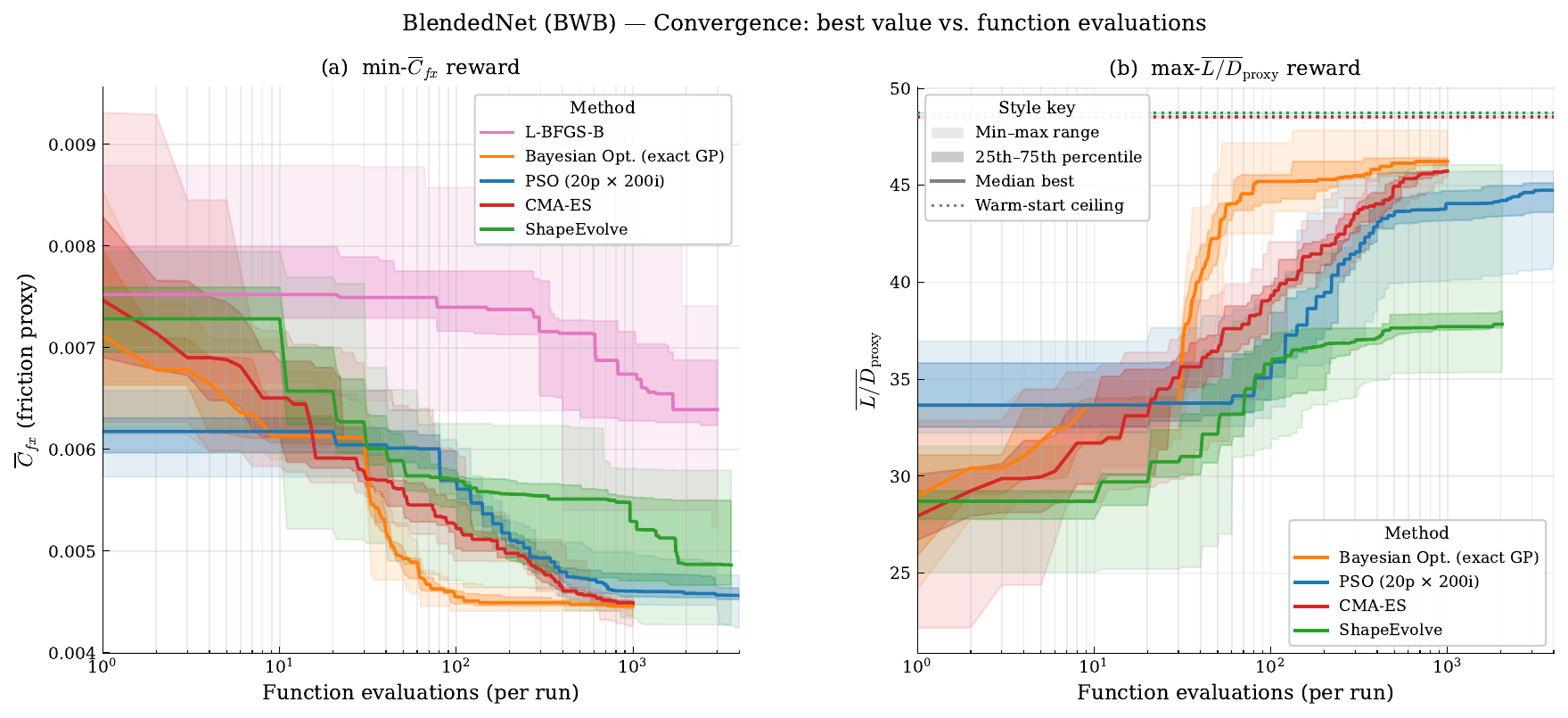}
    \caption{Objectives vs. evaluations plot for 3D BWB Design multipoint tasks. As given in the objective/task definitions, $\overline{C}_{fx} = \frac{1}{5}\sum_{i=1}^{5} C_{fx,i}^\star$ and $\overline{L/D}_{\mathrm{proxy}} = \frac{1}{5}\sum_{i=1}^{5} (C_{L,i}^\star / C_{fx,i}^\star)$ denote mean drag and mean lift-to-drag ratio over the operating points $(C_{L,1}^\star,\ldots,C_{L,5}^\star)$.}
    \label{fig:BlendedNet_convergence_combined}
\end{figure}

Figure \ref{fig:BlendedNet_convergence_combined} provides the convergence plots for both objectives (min-$\overline{C}_{fx}$ and max-$\overline{L/D}_{\mathrm{proxy}}$. All methods show plateauing behavior before the budget is exhausted, suggesting that all curves have settled on their converged values.

For the min-$\overline{C}_{fx}$ cases in subfigure a, the best overall convergence in median value in terms of fastest per evaluation count is provided by Bayesian optimization, followed by CMA-ES, then PSO, then ShapeEvolve. All four of these methods, however, are able to reach similar best designs with $\overline{C}_{fx} \simeq 0.0043$, with a spread of only $4$\%. The L-BFGS-B struggles, taking many evaluations and also converging to a suboptimal median and best solution, suggesting that gradient methods become easily trapped in the non-convex landscape presented here. Nevertheless, the fact that all methods except L-BFGS-B are able to converge to the same best design suggests that the reward landscape is relatively well-behaved.

For the max-$\overline{L/D}_{\mathrm{proxy}}$ cases in subfigure b, L-BFGS-B was not considered for this objective; its poor performance in the min-$\overline{C}_{fx}$ objective suggests that it is expected to perform relatively poorly for this max-$\overline{L/D}_{\mathrm{proxy}}$ setup as well. For the other methods, the performance ranking of Bayesian Optimization > CMA-ES > PSO > ShapeEvolve holds, which is the same ranking as found in the min-$\overline{C}_{fx}$ objective. One key difference that is revealed for the max-$\overline{L/D}_{\mathrm{proxy}}$ is that the best performing designs of each method are separated into two distinct groups. The Bayesian Optimization case achieves a best design with $\overline{L/D}_{\mathrm{proxy}} \simeq 48$, while the other three methods are only able to achieve best designs of $\overline{L/D}_{\mathrm{proxy}} \simeq 46 - 48$. Both of these groups are unable to achieve the performance of the warm-started cases (which are shown as dotted lines in subfigure b). These two distinct groups and the warm-started cases provide evidence of a local optimum trap; further details including how the warm-started cases are motivated and constructed will be detailed further in the discussion related to later figures.

\begin{figure}
    \centering
    \includegraphics[width=0.7\linewidth]{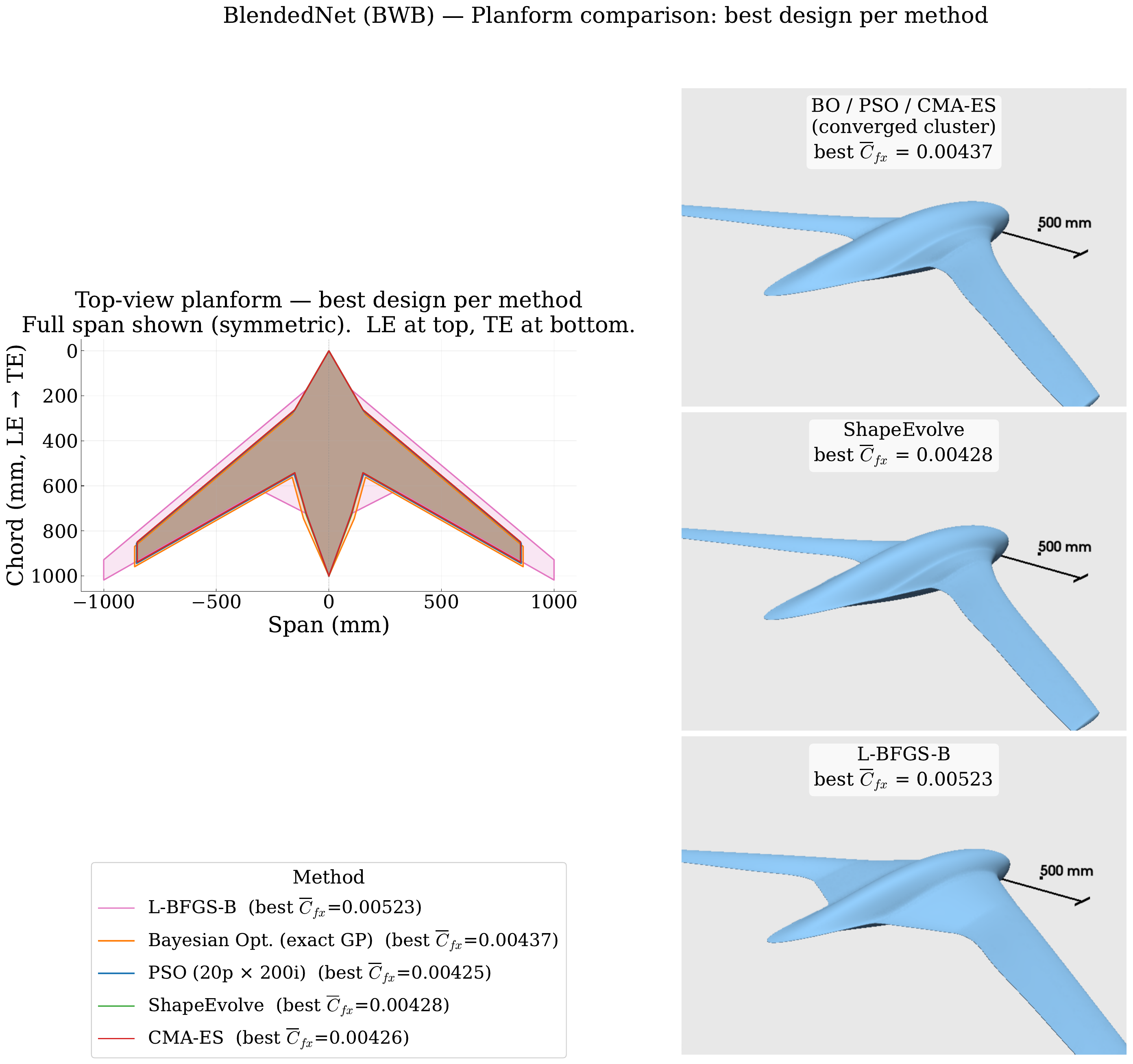}
    \caption{ Best designs, 2D planform and 3D isometric views, for the min-$C_D$ objective. }
    \label{fig:BlendedNet_planform_best_designs_best_cd}
\end{figure}

Figure \ref{fig:BlendedNet_planform_best_designs_best_cd} shows the 2D planform and 3D isometric visualizations of the aircraft body for the best designs as found for each method for the min-$C_D$ objective. Confirming the behavior seen in figure \ref{fig:BlendedNet_convergence_combined}, Bayesian optimization, PSO, and CMA-ES all converge to a single geometry that exhibits notable features of high wing sweep and short tip chord. The L-BFGS-B best design is stuck in a suboptimal basin that results in higher $\overline{C_D}$, as is also consistent from the convergence plot behavior.

\begin{figure}
    \centering
    \includegraphics[width=0.6\linewidth]{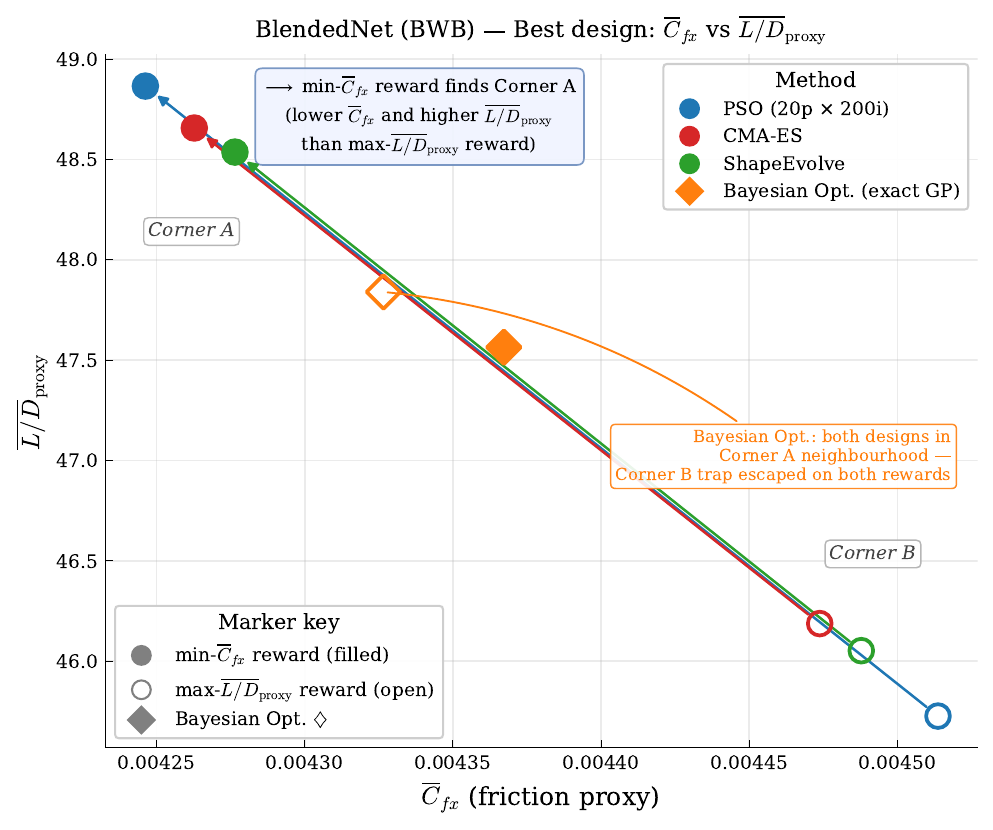}
    \caption{Cross-objective scatter plot for the min-$\overline{C}_{fx}$ and max-$\overline{L/D}_{\mathrm{proxy}}$ objectives. This anomaly, where minimizing for $\overline{C_D}$ is a consequence of the local optimum trap.}
    \label{fig:BlendedNet_cd_ld_scatter}
\end{figure}

Before introducing and discussing the aircraft designs for the max-$\overline{L/D}_{\mathrm{proxy}}$ reward cases, it is informative to compare the performances of the best designs of each method in both min-$\overline{C}_{fx}$ and max-$\overline{L/D}_{\mathrm{proxy}}$ objectives. Figure \ref{fig:BlendedNet_cd_ld_scatter} shows this through the cross-objective scatter plot, with the best designs from min-$\overline{C}_{fx}$ shown with filled circles and the best designs from max-$\overline{L/D}_{\mathrm{proxy}}$ shown with open circles. A few interesting features are visible:
\begin{itemize}
    \item Two distinct regions of performance occur: an optimal corner A with $\overline{L/D}_{\mathrm{proxy}} \simeq 48, \overline{C}_{fx} \simeq 0.00430$, and a sub-optimal corner B with $\overline{L/D}_{\mathrm{proxy}} \simeq 46, \overline{C}_{fx} \simeq 0.00450$.
    \item The min-$\overline{C}_{fx}$ cases for PSO, CMA-ES, and ShapeEvolve converge to corner A, while the max-$\overline{L/D}_{\mathrm{proxy}}$ for these same methods converge to corner B (which results in higher $\overline{C_D}$ and lower $\overline{L/D}_{\mathrm{proxy}}$). To reiterate, directly optimizing for maximum $\overline{L/D}_{\mathrm{proxy}}$ produces worse lift-over-drag than from directly optimizing for minimum $\overline{C}_{fx}$. 
\end{itemize}
These observations indicate the existence of a local optimum trap due to the different landscapes that result from each objective: the max-$\overline{L/D}_{\mathrm{proxy}}$ objective creates a more complex optimization landscape, which importantly features a less dominant attractor for corner A. 

For Bayesian optimization, the best results for both objectives appear to escape the corner B trap and find the corner A basin. However, the best designs slightly underperform those of other methods. This is likely due to the specifics of the method: Bayesian optimization tends to have less-efficient budget utilization that spreads evaluations across both basins, and it also will tend to find the right basin quickly but then not perform well with tight localization once the basin is found. This less-precise convergence is geometrically observable through the slightly wider aircraft design seen in figure \ref{fig:BlendedNet_planform_best_designs_best_cd}.

\begin{figure}
    \centering
    \includegraphics[width=0.99\linewidth]{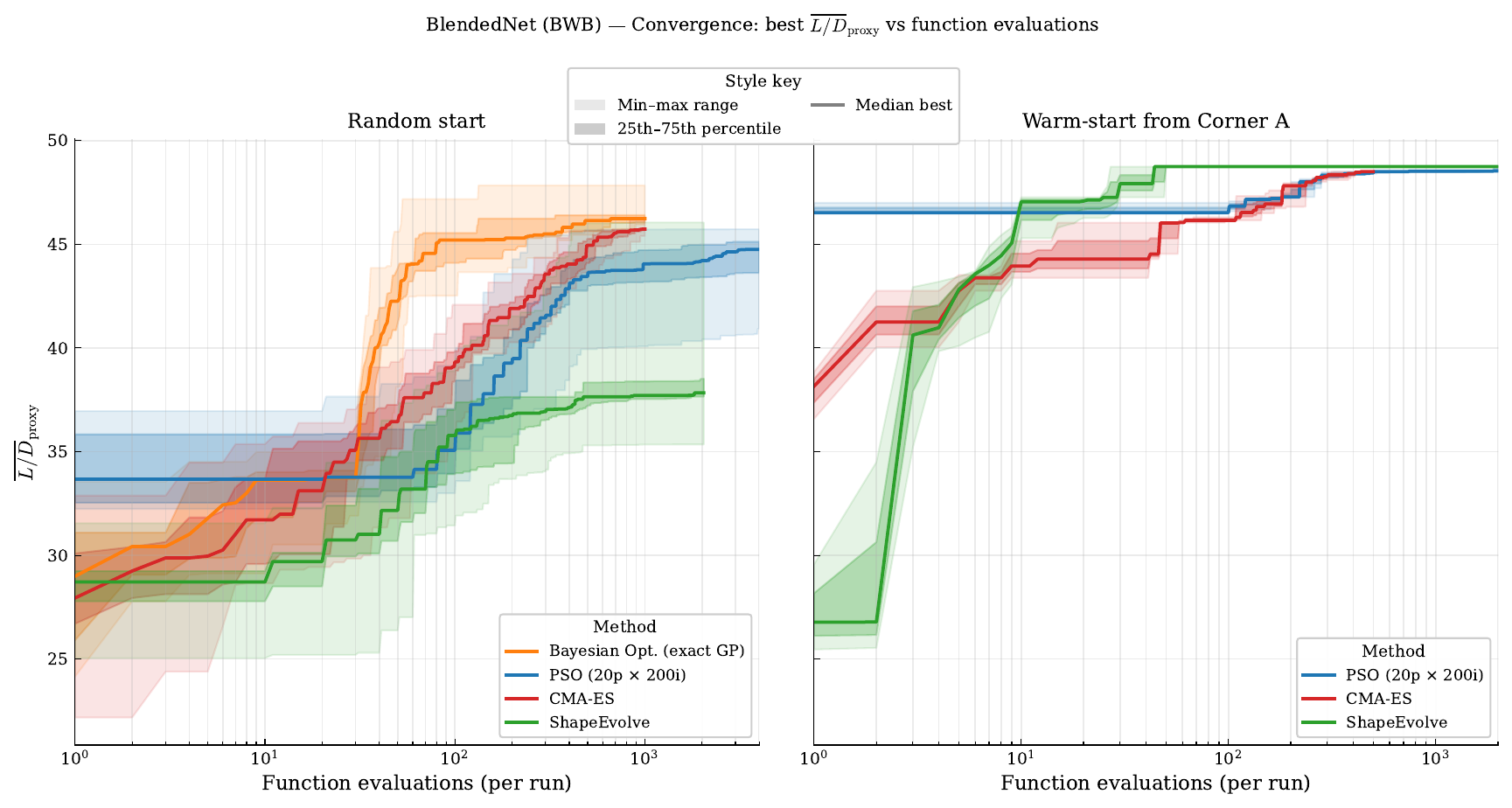}
    \caption{Objectives vs. evaluations plot for max-$\overline{L/D}_{\mathrm{proxy}}$ objective, with warm-start. Warm-started runs begin (at evaluation 1) below corner A (with $\overline{L/D}_{\mathrm{proxy}} \simeq 48.5$) because PSO ($\pm 10$\% Gaussian noise around the seed), CMA-ES (with initial step size $\sigma_0 = 0.3$ in normalized space), and ShapeEvolve (LLM-generated first batch) each perturb/scatter the initial evaluations around the warm-start point.}
    \label{fig:BlendedNet_convergence_LD_vs_evals_warmstart_comparison}
\end{figure}

\begin{figure}
    \centering
    \includegraphics[width=0.7\linewidth]{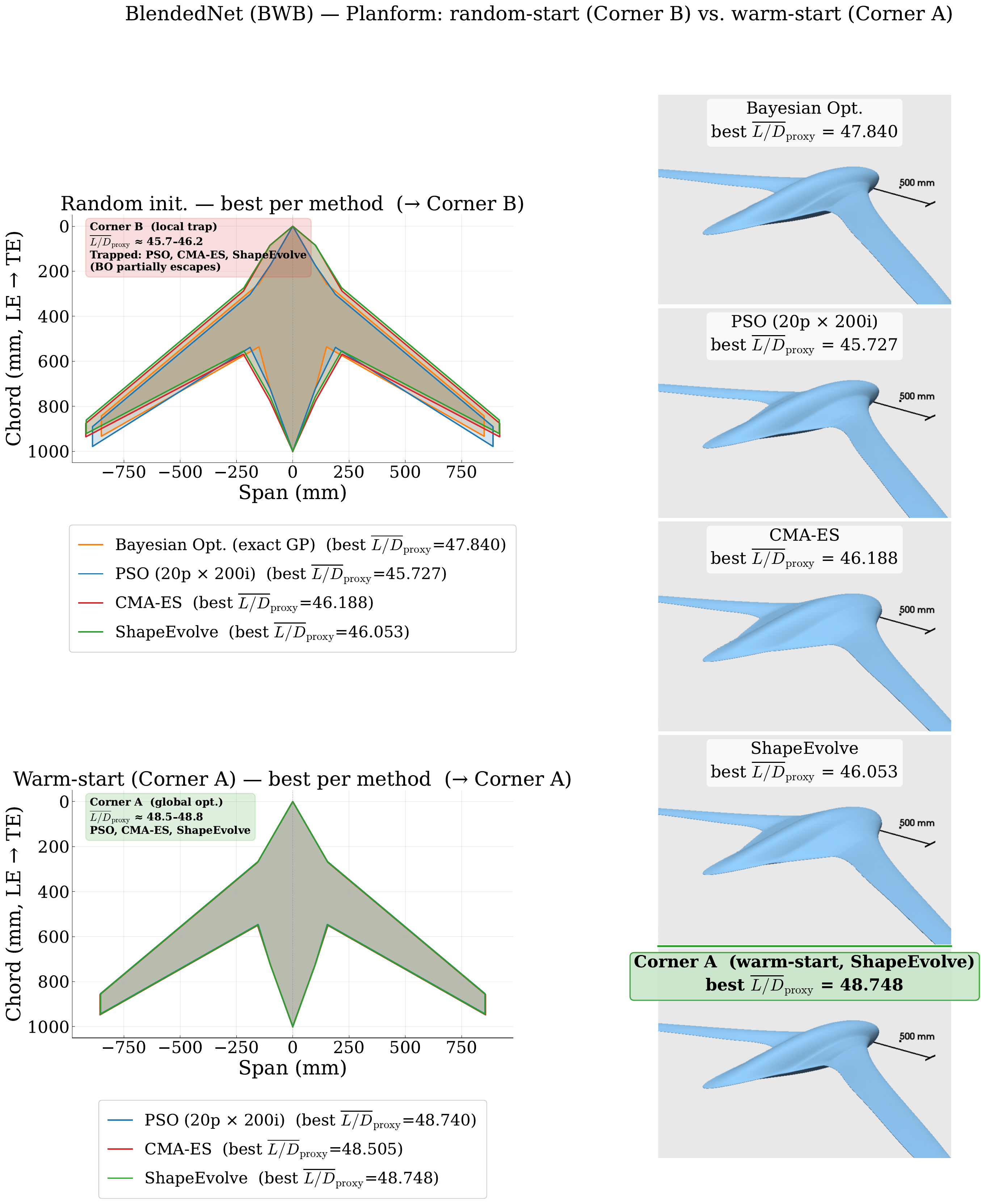}
    \caption{Best designs, 2D planform and 3D isometric views, max-$\overline{L/D}_{\mathrm{proxy}}$ objective}
    \label{fig:BlendedNet_planform_best_designs_max_LD}
\end{figure}

Figures  \ref{fig:BlendedNet_convergence_LD_vs_evals_warmstart_comparison} and \ref{fig:BlendedNet_planform_best_designs_max_LD} show the aircraft design geometries and the convergence plots for the max-$\overline{L/D}_{\mathrm{proxy}}$ objective. The two different regions (corner A and corner B) that were introduced previously and that the best designs converge to can now be seen geometrically. Physically, corner A designs feature high sweep and a narrower tip that also coincidentally achieves better $\overline{L/D}_{\mathrm{proxy}}$.

Both figures  \ref{fig:BlendedNet_convergence_LD_vs_evals_warmstart_comparison} and \ref{fig:BlendedNet_planform_best_designs_max_LD} show the warm-started results for the max-$\overline{L/D}_{\mathrm{proxy}}$. The motivation for warm-starting is as follows: all methods (except for Bayesian optimization) in the max-$\overline{L/D}_{\mathrm{proxy}}$ cases plateau at corner B, which is $\sim 5$\% less in $\overline{L/D}_{\mathrm{proxy}}$ than corner A. Importantly, none of the seeds for these methods venture into the corner A design solution, and all profiles are plateauing towards converged values; this indicates that the failure to find corner A is due to a local optimum trap and not because of insufficient evaluation budget.

For the warm-start setup, $3$ seeds each are run for the max-$\overline{L/D}_{\mathrm{proxy}}$ objective for PSO, CMA-ES, and ShapeEvolve, as initialized at corner A using the best design from each method's corresponding min-$\overline{C}_{fx}$ case. These results are shown in the right-hand panel of figure \ref{fig:BlendedNet_convergence_LD_vs_evals_warmstart_comparison}. For all three methods, the warm-started cases converge to corner A within budget; thus, warm-starting is sufficient to overcome the basin-of-attraction bias found in this max-$\overline{L/D}_{\mathrm{proxy}}$ objective, which caused the initial- (random)-start cases of the max-$\overline{L/D}_{\mathrm{proxy}}$ objective to settle in corner B rather than in the global optimum corner A. This confirms that the failure to find corner A from random initialization is an exploration problem (rather than a convergence issue), since all three methods are capable of converging to corner A once the optimal basin is reached. It is emphasized that the warm-starting exercise here is performed as a diagnostic tool and is not recommended as an approach for general optimization design.   

\begin{figure}
    \centering
    \includegraphics[width=0.85\linewidth]{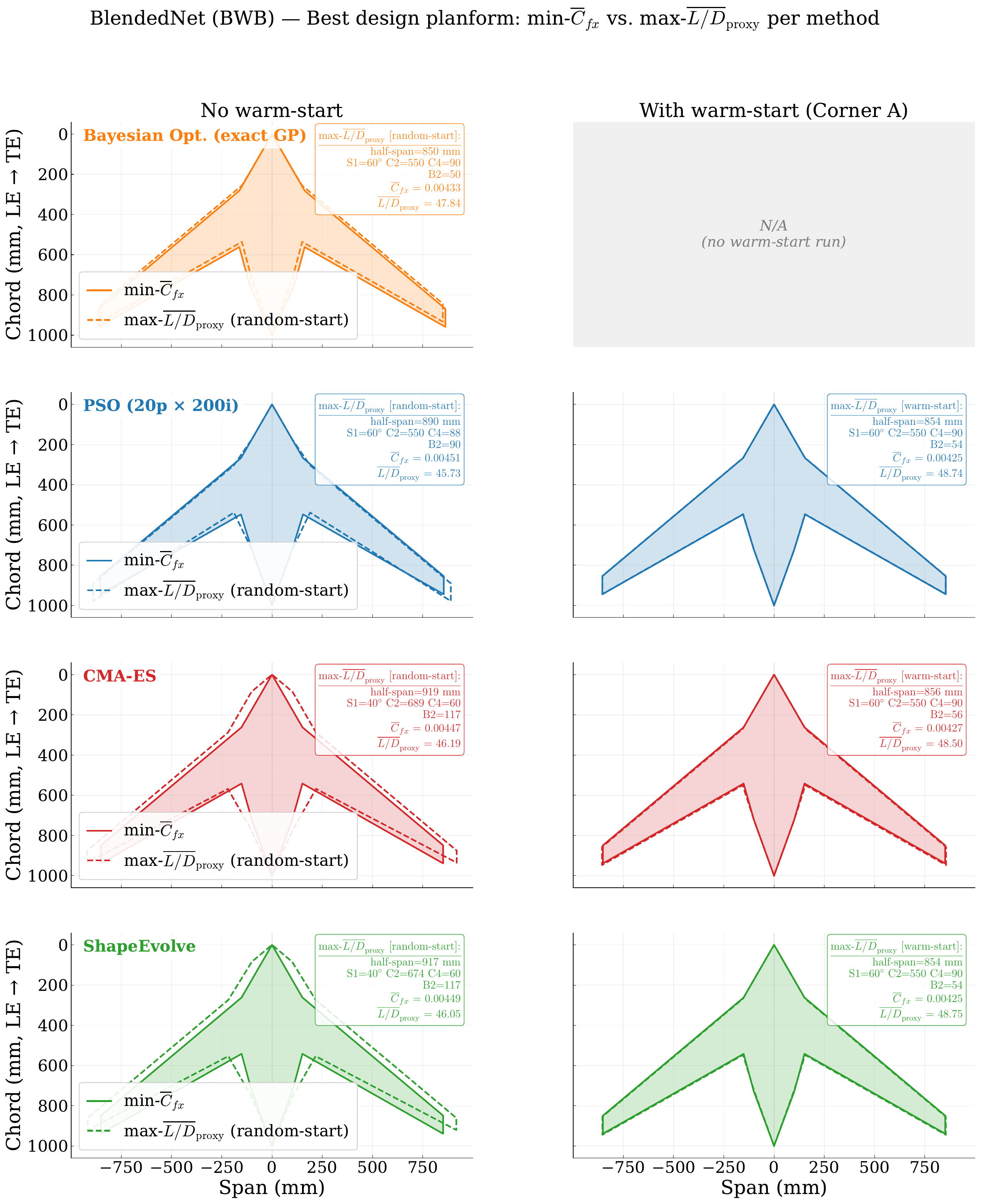}
    \caption{Best designs, 2D planform overlay comparisons across both objectives. Left-hand side plots show random-start max-$\overline{L/D}_{\mathrm{proxy}}$ cases overlaid on min-$\overline{C}_{fx}$ cases, and right-hand side plots show warm-start max-$\overline{L/D}_{\mathrm{proxy}}$ cases overlaid on min-$\overline{C}_{fx}$ cases.}
    \label{fig:BlendedNet_planform_reward_comparison}
\end{figure}

The planform comparison plot in figure \ref{fig:BlendedNet_planform_reward_comparison} shows the geometric features that arise from the cross-reward anomaly. Without warm-starting, the max-$\overline{L/D}_{\mathrm{proxy}}$ cases for PSO, CMA-ES, and ShapeEvolve converge to designs that are wider with less wing sweep, leading to suboptimal high drag and low lift-to-drag ratio. With warm-starting, the converged designs of the same cases are guided to essentially the same optimal designs that the min-$\overline{C}_{fx}$ finds. Thus, once the optimizer for each method is guided to the correct basin, the same global optimum is eventually found. 

As detailed further in the discussion for figure \ref{fig:BlendedNet_cd_ld_scatter}, the Bayesian optimization cases are not warm-started and are slightly under-performing compared to the rest of corner A due to limitations of the Bayesian optimization method.

\begin{figure}[h!]
    \centering

    \begin{subfigure}{0.48\textwidth}
        \centering
        \includegraphics[width=\linewidth]{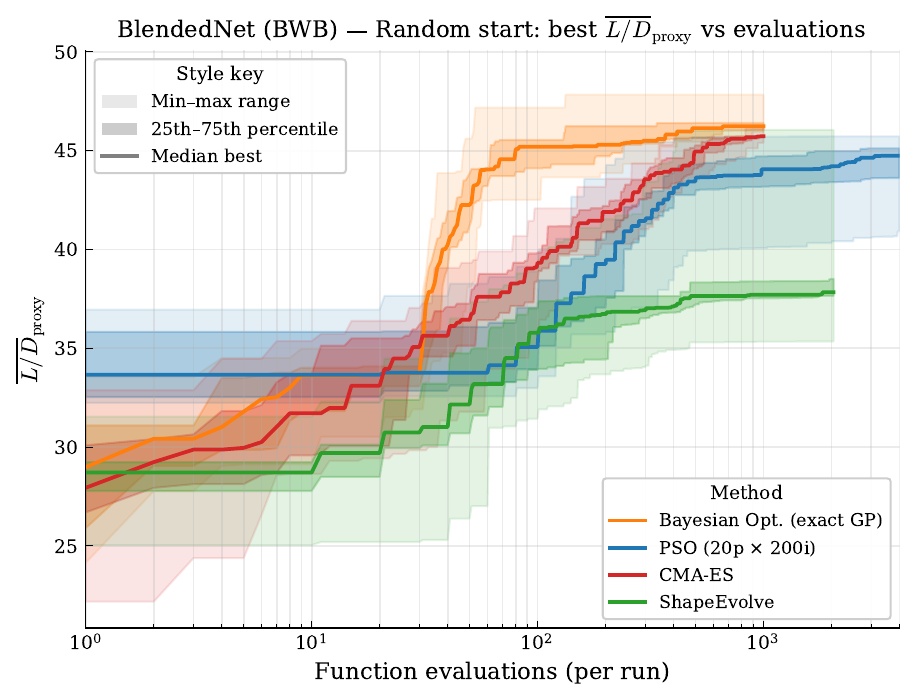}
        \caption{}
        \label{fig:bwb-random-init-plot}
    \end{subfigure}
    \hfill
    \begin{subfigure}{0.48\textwidth}
        \centering
        \includegraphics[width=\linewidth]{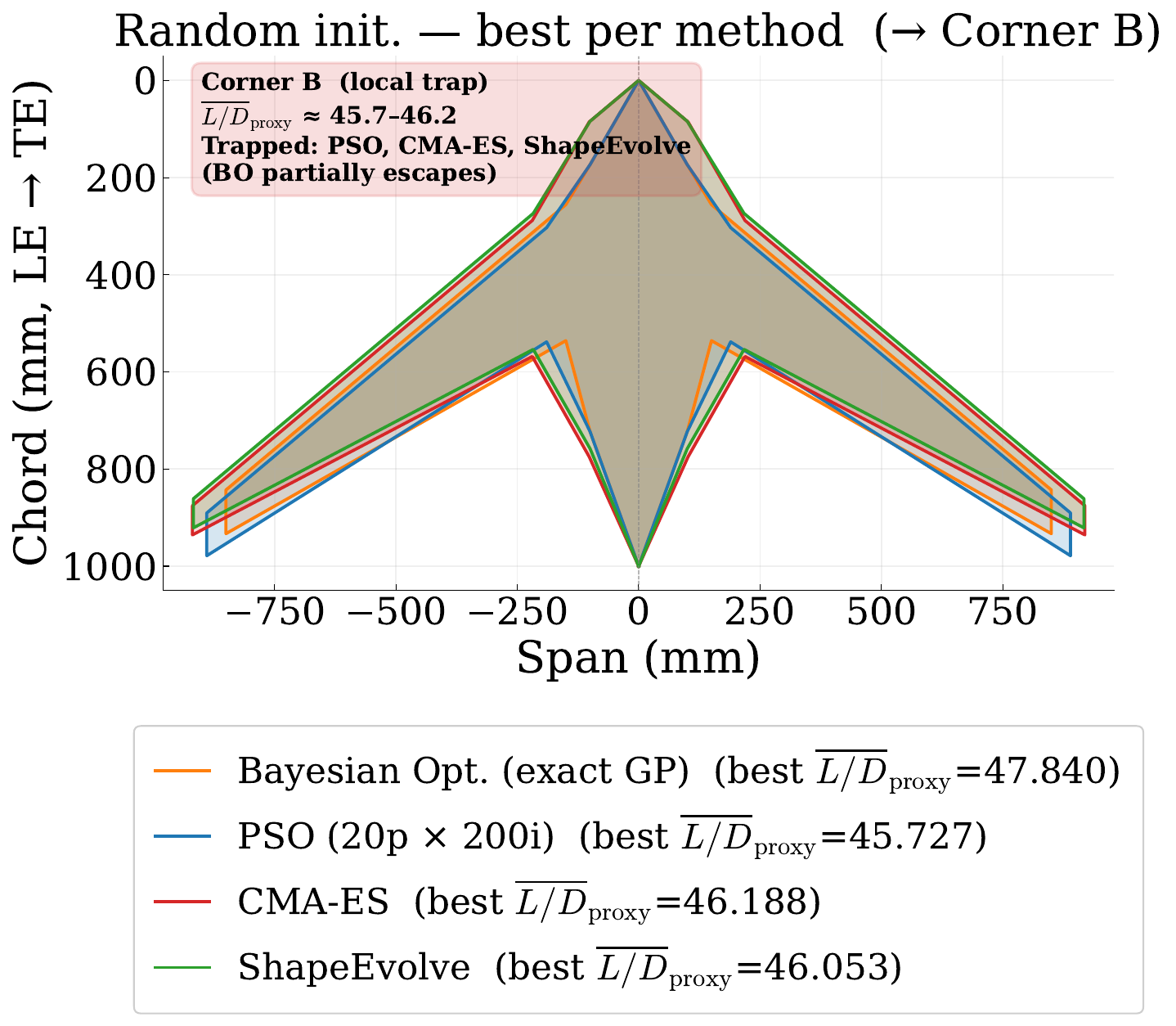}
        \caption{}
        \label{fig:bwb-random-init-shape}
    \end{subfigure}

    \vspace{0.5em}

    \begin{subfigure}{0.48\textwidth}
        \centering
        \includegraphics[width=\linewidth]{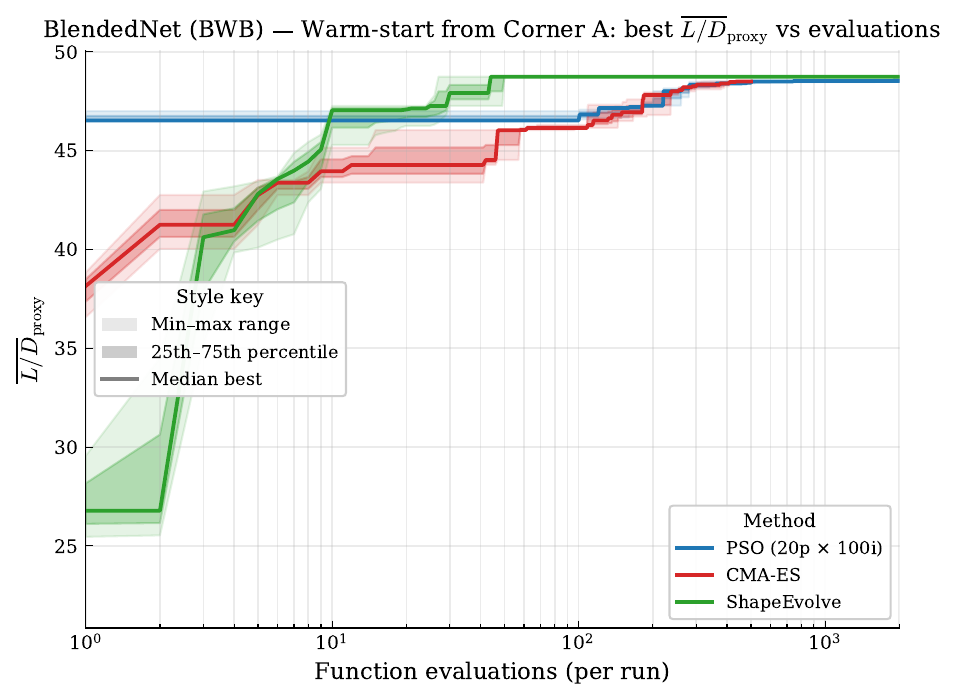}
        \caption{}
        \label{fig:bwb-warm-start-plot}
    \end{subfigure}
    \hfill
    \begin{subfigure}{0.48\textwidth}
        \centering
        \includegraphics[width=\linewidth]{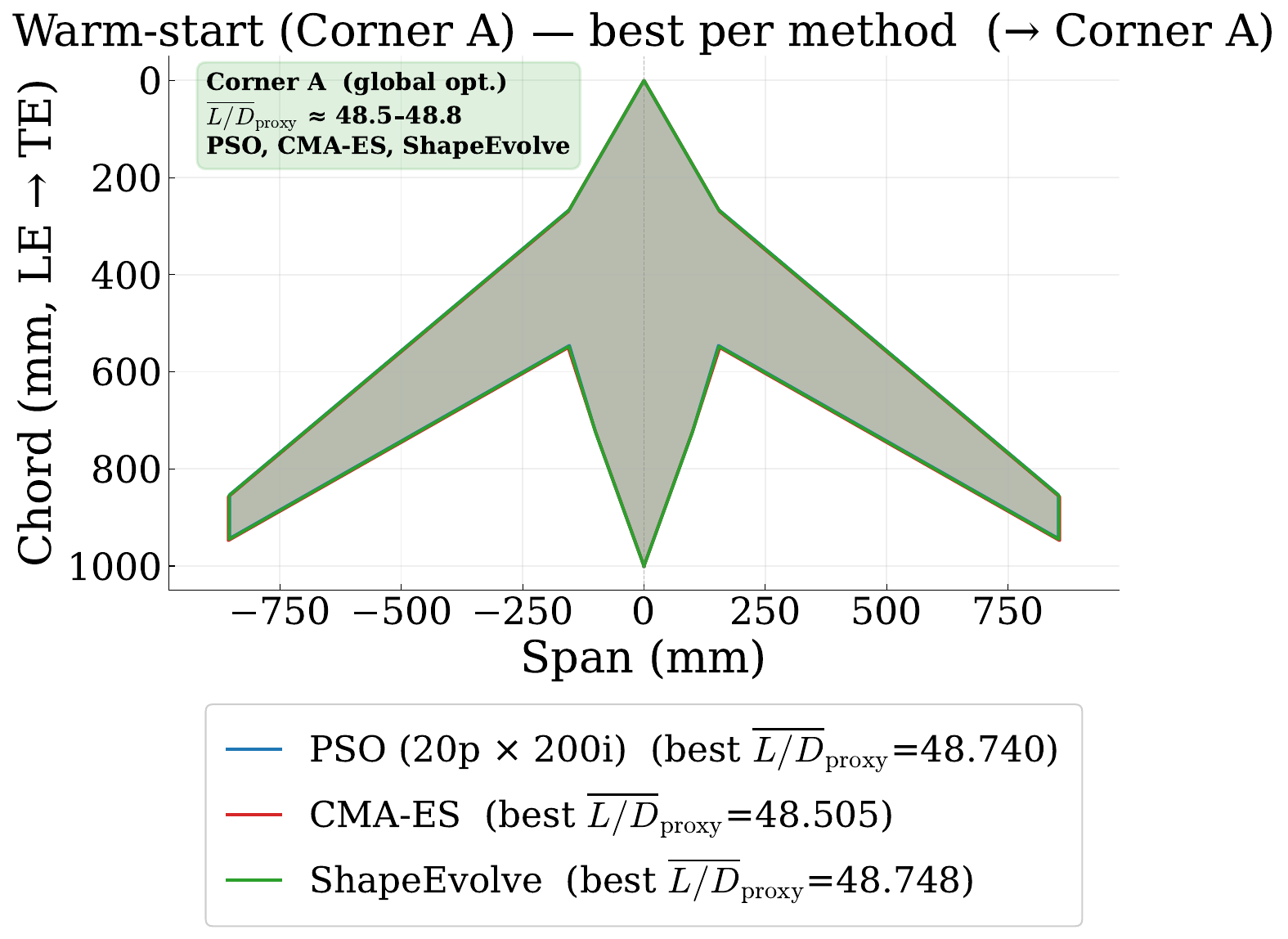}
        \caption{}
        \label{fig:bwb-warm-start-shape}
    \end{subfigure}

    \caption{3D BWB experiment with different initializations; (a) reward vs. evaluations \& (b) BWB shape with random initialization, (c) reward vs. evaluations \& (d) BWB shape with a warm start.}
    \label{fig:bwb-results}
\end{figure}

\paragraph{Discussion of surrogate exploitation with the corrected (total) drag metric}

When using $\overline{C}_D^{\mathrm{int}}$ in place of $\overline{C}_{fx}$ as the objective, the different methods all converged to a design that displays significant boundary saturation of parameters, but to a very different design from the corner A solution found when using $\overline{C}_{fx}$. Instead, a very-low-aspect-ratio geometry ($AR \approx 1.02$) is produced, which has the following unusual features:
\begin{itemize}
    \item the lift target is satisfied but at a suspiciously small angle of attack ($4.928^\circ$ vs. the reference design's $8.979^\circ$)
    \item $\overline{C}_D^{\mathrm{int}} \approx 0.003$--$0.004$, which is physically implausible for a lifting configuration at the chosen conditions and also about an order of magnitude less than the reference design's $\overline{C}_D^{\mathrm{int}} \approx 0.027$
    \item pressure drag $C_{D,\mathrm{pressure}} \approx 0.001$, which is $<5$\% that of the reference design's and physically inconsistent with classical aerodynamic theory for a $AR \approx 1.0$ geometry (whose induced drag alone should be larger than this value)
    \item bounds saturation of a significant number (at least $5$ of the $10$ for each method) of the parameters (shared across all methods: \texttt{B1}$\to$upper, \texttt{B3}$\to$lower, \texttt{C3}$\to$lower)
\end{itemize}
There is thus strong evidence that the surrogate exploitation is present, which is the mechanism causing the optimizer to find implausible designs.

As also discussed in the setup, cases that constrained $AR \geq 2.5$ were then investigated to remove this low-$AR$ exploitation option. The converged designs found from this constrained $AR$ setup have $\overline{C}_D^{\mathrm{int}}$ that are now roughly double compared to the unconstrained design's value, at $\overline{C}_D^{\mathrm{int}} \approx 0.006$ -- $0.007$. The unconstrained $AR$ result is thus confirmed to be due to surrogate exploitation rather than a robust optimum. However, the constrained $AR$ converged designs suffer from:
\begin{itemize}
    \item near-zero profile drag (an unphysical result for any real aircraft body)
    \item bounds saturation of at least $4$ of the $10$ parameters for each method, resulting in convergence to a different boundary corner from the unconstrained case (parameters pushed to their bounds include: \texttt{B1}$\to$upper, \texttt{B2}$\to$lower, \texttt{C3}$\to$lower, \texttt{S1}$\to$upper)
\end{itemize}
Just as is seen in other surrogates such as DrivAerStar for the 3D car design in Section \ref{sec:3D_Car_Design}, there is strong evidence that the boundary-collapse failure mode is a property of the BlendedNet surrogate. Adjusting the objective simply shifts which corner is exploited rather than eliminating surrogate exploitation.

\subsection{2D Airfoil Design}
\paragraph{Design variables} We use the 2D airfoil parametrization from \cite{sharpe2025neuralfoilairfoilaerodynamicsanalysis}. Let
\[
\mathbf{x} = \mathbf{p}
=
\left(
u_1,\ldots,u_8,\,
\ell_1,\ldots,\ell_8,\,
p_{\mathrm{LE}},\,
t_{\mathrm{TE}}
\right)^{\top}
\in \mathbb{R}^{18},
\]
where $u_i$ are upper-surface CST weights, $\ell_i$ are lower-surface CST
weights, $p_{\mathrm{LE}}$ is the Kulfan leading-edge modification weight,
and $t_{\mathrm{TE}}$ is the trailing-edge thickness (fraction of chord).

As introduced previously, the cases are simulated with the airfoil surrogate \texttt{NeuralFoil}.

\subsubsection{Single-point Task} \label{sec:2D_Airfoil_Design_Single_point_Task}

\paragraph{Operating point}
The aerodynamics model is evaluated at the single operating point
\[
\alpha^\star = 5^\circ,
\qquad
\mathrm{Re}_c^\star = 10^7,
\qquad
M_{\infty} = 0.2
\qquad
N_{\mathrm{crit}}^\star = 9,
\]
with natural transition.

\paragraph{Objective}
Maximize the lift-to-drag ratio
\[
\max_{\mathbf{x}} \frac{C_L(\mathbf{p})}{C_D(\mathbf{p})} -  \lambda \sum_k v_k(\mathbf{p}), \qquad \lambda = 500,
\]
where $C_L(\mathbf{p})$ and $C_D(\mathbf{p})$ are the lift and drag
coefficients predicted by \texttt{NeuralFoil} at
$\left(\alpha^\star,\mathrm{Re}_c^\star,N_{\mathrm{crit}}^\star\right)$
for the shape $\mathbf{p}$. $\lambda$ is the penalty weight and $v_k$ is the fractional (normalized) violation of the $k$-th constraint. The justification and explanation for this formulation as a penalty-weighted objective are provided later in this section.

\paragraph{Design space bounds}
\[
-0.30 \le u_i \le 0.60
\qquad \text{for } i=1,\ldots,8,
\]
\[
-0.30 \le \ell_i \le 0.30
\qquad \text{for } i=1,\ldots,8,
\]
\[
-0.50 \le p_{\mathrm{LE}} \le 0.50,
\]
\[
0.000 \le t_{\mathrm{TE}} \le 0.010.
\]

\paragraph{Constraints}
\[ 
\text{thickness } t(x/c) > 0 \quad \forall\, x/c \in [0,1], \text{ where $c$ is the chord length,}
\]            
\[
t(0.33) \ge 0.128,
\]              
\[ 
t(0.90) \ge 0.014,
\]       
\[ 
\text{trailing-edge wedge angle } \theta_{\mathrm{TE}} \ge 6.03^\circ,
\]                  
\[ 
\text{leading-edge angle } \theta_{\mathrm{LE}} = 180^\circ,
\]
\[
\text{wiggliness functional } \mathcal{W}(\mathbf{u}, \boldsymbol{\ell}) \le 2\,\mathcal{W}_{\mathrm{NACA\,0012}} \text{ with } \mathcal{W} = \sum_s \sum_i (\Delta^2 w_{s,i})^2,
\]
\[
\text{pitching moment } C_M \ge -0.133,
\]
\[ 
\text{analysis confidence (in-distribution) } \sigma_{\mathrm{conf}} > 0.90.
\]

\paragraph{Justification for objective}
Because \texttt{NeuralFoil} is a surrogate, directly maximizing maximizing $C_L / C_D$ without enforcing the constraints would lead to surrogate exploitation, in which the optimizer discovers that the highest predicted lift-over-drag values occur in the regions of the design space in which \texttt{NeuralFoil}'s analysis confidence $\sigma_{\mathrm{conf}} \rightarrow 0$ and the constraints (especially $C_M$) are violated. This leads the optimizer to converge to aerodynamically meaningless geometries with $C_L/C_D$ values in the range of $400$--$700$. Problematically, for this problem setup, thin airfoils with the most camber simultaneously and monotonically maximize $C_L/C_D$ and minimize $\sigma_{\mathrm{conf}}$.

Constraints thus need to be enforced numerically. In assessing some standard used approaches, hard constraint filtering (in which infeasible evaluations are rejected outright) provides no gradient signal outside of the feasible region, log-barrier methods (in which optimizers are initialized strictly within the feasible region) are incompatible with the random start and update steps of genetic algorithms and Bayesian optimization methods, and automated Lagrangian with standard optimization methods would either require heavy non-standard modifications or are incompatible by formulation.

Thus, a normalized penalty formulation is adopted. The fractional constraint violation $v_k$ is constructed such that a complete failure of a constraint yields $v_k = 1$. The value of the weight $\lambda = 500$ was chosen to have the following features.
\begin{itemize}
    \item A complete violation of one constraint ($v_k = 1$) costs more reward than the largest feasible $C_L/C_D$ gain observed in preliminary numerical experiments; thus, infeasible designs are decisively unprofitable for the optimizer. 
    \item Additionally, mild near-boundary violations ($v_k \ll 1$) cause a proportionally small penalty, which preserves the gradient signal that certain optimization methods require in order for them to function. 
\end{itemize}
This method was found to be robust across all tested methods.

\paragraph{Results}

\textit{Computational cost per $1{,}000$ evaluations per run} ($n$ = independent runs); $1$ core unless otherwise noted: 
\begin{itemize}
    \item L-BFGS-B ($n = 40$): $\approx 0.02$ CPU-hours
    \item Bayesian Opt ($n = 9$): $1{,}000$ - $7{,}500$ CPU-hours ($128$-core exclusive node)
    \item PSO ($n = 25$): $\approx 0.02$ CPU-hours
    \item ShapeEvolve ($n = 20$): $0.6$ -- $1.8$ CPU-hours
\end{itemize}

In order to evaluate the end-to-end quality of the optimization, an evaluation protocol is constructed which separates the optimizer, the surrogate, and the ground-truth solver into three separate stages:
\begin{itemize}
    \item Stage 1: each optimization method runs using the \texttt{NeuralFoil} surrogate to find a candidate design
    \item Stage 2: a local IPOPT-based gradient method refines upon the best design using \texttt{NeuralFoil} gradients and evaluation
    \item Stage 3: the refined design is evaluated using \texttt{XFOIL}, serving as the ground truth for the airfoil performance.
\end{itemize}

\begin{figure}
    \centering
    \includegraphics[width=0.7\linewidth]{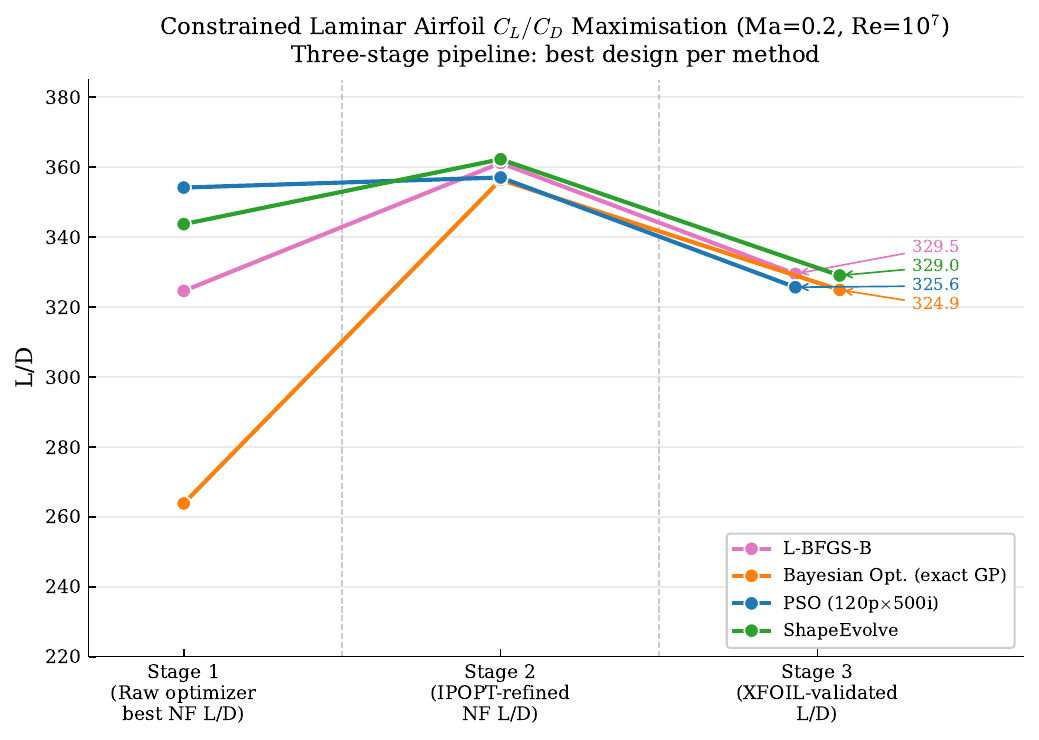}
    \caption{Three-stage optimization pipeline for the 2D airfoil single-point maximum lift-to-drag task. Each line traces a method's best design through: (1) raw \texttt{NeuralFoil} optimization and evaluation, (2) local IPOPT refinement, still evaluated with \texttt{NeuralFoil}, and (3) final validation of the stage 2 design via \texttt{XFOIL} evaluation.}
    \label{fig:NeuralFoil_LD_pipeline_stages}
\end{figure}

Figure \ref{fig:NeuralFoil_LD_pipeline_stages} shows the trajectory in the three-stage protocol for each method. The stage 1 raw outputs via \texttt{NeuralFoil} span a wide range of outputs (with $L/D$ values from $255$ to $364$). The IPOPT refinement in stage 2, however, then collapses all four methods to $360$ with a spread of only $\simeq 1.4$\%. Thus, stage 1 was primarily used to locate the correct basin of attraction for each method, rather than finding the best $L/D$ design quality. Finally, stage 3 is used as a surrogate validation step; all four methods show a uniform and systematic $\sim\!9\%$ drop in value of $L/D \simeq 360$ in stage 2 via \texttt{NeuralFoil} to $L/D \simeq 327$ when evaluated using \text{XFOIL} as the ground truth.

This $\sim\!9\%$ gap is consistent with known biases of airfoil surrogates such as \texttt{NeuralFoil} at high camber and near the confidence boundary. As a technical note, the stage 2 IPOPT refinement step applied a constraint back-off of $C_M \geq -0.125$ in place of the original $-0.133$ value. This compensates for a systematic \texttt{NeuralFoil} underprediction of $C_M$ (of values of $0.003$ to $0.005$ at the constraint boundary) and helps to ensure that the final designs are feasible when validated using \texttt{XFOIL}.

\begin{figure}
    \centering
    \includegraphics[width=0.95\linewidth]{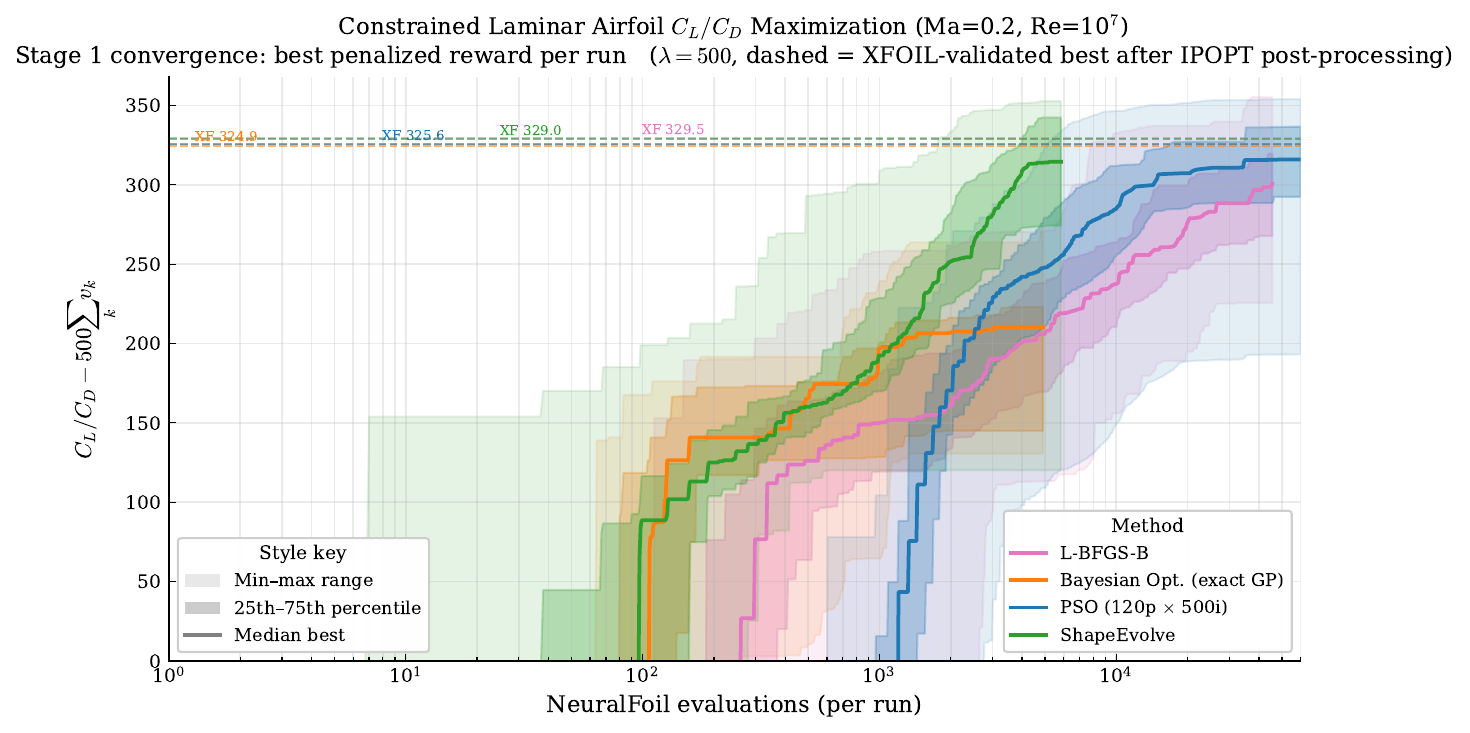}
    \caption{Stage 1 convergence plot (objective vs. evaluations) of the four optimization methods for the 2D airfoil single-point maximum lift-to-drag task. The results of stage 3 (\texttt{XFOIL}-validated best designs) are also shown as dashed lines for each method. Note that within each method, all runs are included in the median (with each trace extended to the full evaluation budget). }
    \label{fig:NeuralFoil_LD_convergence_best_reward}
\end{figure}

Figure \ref{fig:NeuralFoil_LD_convergence_best_reward} shows the convergence bands for all four methods. ShapeEvolve finds nontrivial designs at the earliest stage, and also reaches a competitive "stage 1 starting point" earlier than the other methods, at $O(1000)$s of evaluations. The Bayesian optimization runs initially behaves similarly to the ShapeEvolve for the median profile but then stagnates at a suboptimal value. PSO and and L-BFGS-B each require about an order of magnitude more evaluations than ShapeEvolve to reach their stage 1 starting point values. It is reiterated and emphasized, given the discussion related to \ref{fig:NeuralFoil_LD_pipeline_stages}, that stage 1 is only important for finding the general (optimal) basin; thus, the final $L/D$ values reached by each method are of only marginal relevance.

\begin{figure}
    \centering
    \includegraphics[width=0.7\linewidth]{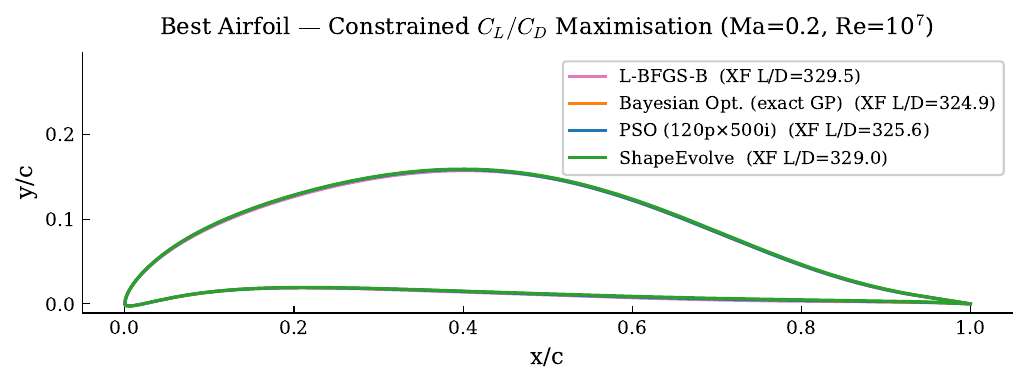}
    \caption{Overlay of best IPOPT-refined airfoil profiles (with $C_M \geq -0.125$) for stage 2 for all four optimization methods for the 2D airfoil single-point maximum lift-to-drag task, with the \texttt{XFOIL}-validated $L/D$ value reported in the legend. }
    \label{fig:NeuralFoil_LD_airfoil_comparison}
\end{figure}

Figure \ref{fig:NeuralFoil_LD_airfoil_comparison} shows the best IPOPT-refined airfoil profiles from stage 2; all four profiles are essentially indistinguishable, consistent with the \texttt{XFOIL}-validated values in stage 3 being converged to the same global optimum. Physically, the airfoil geometry is characterized by high camber on the upper surface (peaking at $x/c \simeq 0.40, y/c \simeq 0.165$, a flat lower surface, and a thin trailing edge.

\subsubsection{Multi-point Task}\label{sec:2D_Airfoil_Design_Multi_point_task}
We use the Drela’s multipoint Daedalus airfoil problem (minimum mean drag at six lift targets with structural and moment constraints) \citep{drela1998pros, sharpe2025neuralfoilairfoilaerodynamicsanalysis}, for which the DAE-11 is the canonical expert design.

\paragraph{Reynolds schedule (fixed-lift polar)}
For each target lift coefficient $C_{L,i} = C_L(\mathbf{p}, \alpha_i)$ in the set below ($\alpha_i$ angles of attack at the operating points),
\[
\mathrm{Re}_{c,i} = 500{,}000 \left(\frac{C_{L,i}}{1.25}\right)^{-1/2},
\qquad
M_{\infty} = 0.03.
\]

\paragraph{Objective}
Minimize the weighted mean drag
\[
\min_{\mathbf{p}} \overline{C_D}
=
\min_{\mathbf{p}}
\frac{\sum_{i=1}^{6} w_i \, C_D\!\left(\mathbf{p}, \alpha_i\right)}
{\sum_{i=1}^{6} w_i},
\]
with operating points
\[
C_{L,i} \in \{0.8,\,1.0,\,1.2,\,1.4,\,1.5,\,1.6\}
\]
and weights
\[
w_i \in \{5,\,6,\,7,\,8,\,9,\,10\},
\]
where $C_D(\cdot)$ is evaluated by the aerodynamics model at
$(\mathrm{Re}_{c,i}, M_{\infty})$ above.

\paragraph{Constraints}
Same as in single-objective task, with the following modifications:
\[
C_{M,i} \geq -0.133
\qquad \text{for all } i=1,\ldots,6,
\]
\[
\sigma_{\mathrm{conf}} > 0.90 \qquad \text{for all } i=1,\ldots,6 ,
\] 
\[
\alpha \text{ monotonicity, i.e. } C_L \text{ increases with } \alpha,
\]
\[
C_L \text{ target reachability (all multi-point targets solvable) }.
\]

\begin{figure}
    \centering
    \includegraphics[width=0.95\linewidth]{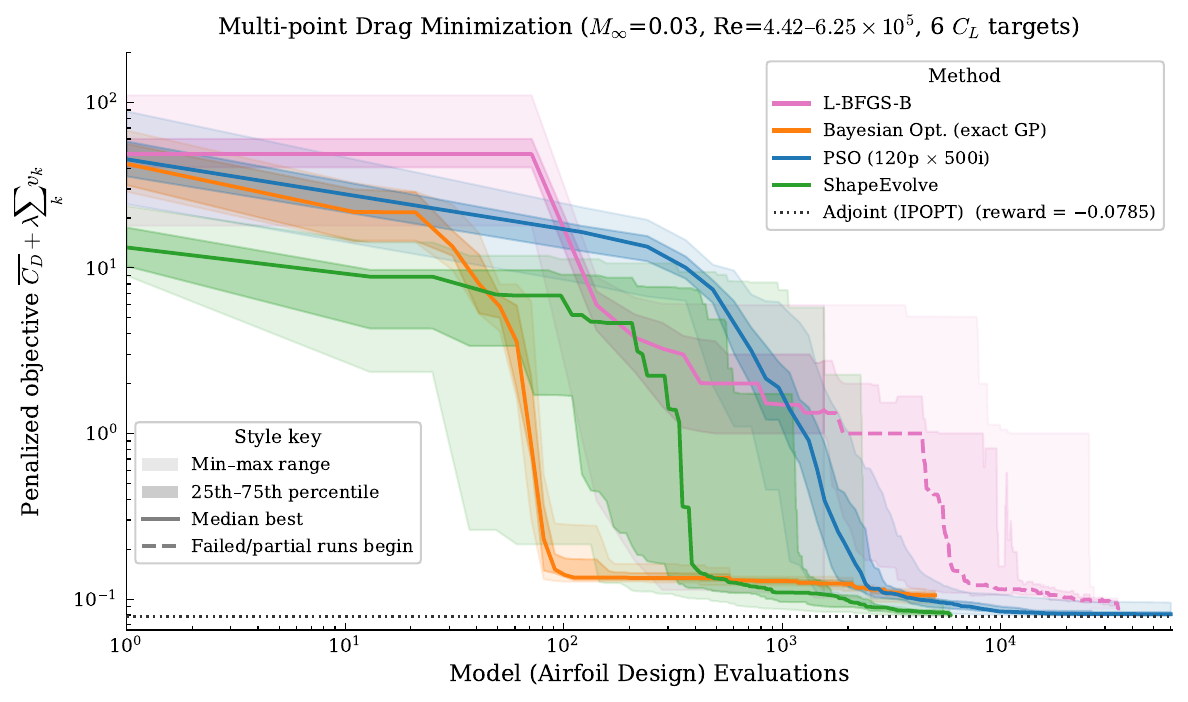}
    \caption{Convergence plot (objective vs. evaluations) plot for the 2D airfoil multi-point drag minimization task; Penalized objective $= $ weighted $\overline{C_D}$ + constraint penalty, with lower value corresponding to better performance. Note that within each method, all runs are included in the median (with each trace extended to the full evaluation budget). Failed runs (whose local minimum result in infeasible designs) are present for L-BFGS-B; their start is denoted with the dashed lines; this is discussed further in the text. }
    \label{fig:NeuralFoil_objective_vs_iterations}
\end{figure}

\paragraph{Results}

\textit{Computational cost per $1{,}000$ evaluations per run} ($n$ = independent runs); $1$ core unless otherwise noted: 
\begin{itemize}
    \item Adjoint (IPOPT) ($n = 1$): $\approx 0.01$ CPU-hours
    \item L-BFGS-B ($n = 40$): $0.05$ -- $0.10$ CPU-hours
    \item Bayesian Opt ($n = 4$): $\approx 1{,}100$ -- $3{,}500$ CPU-hours ($128$-core exclusive node)
    \item PSO ($n = 25$): $0.05$ -- $0.10$ CPU-hours
    \item ShapeEvolve ($n = 21$): $0.8$ -- $1.0$ CPU-hours
\end{itemize}

Figure \ref{fig:NeuralFoil_objective_vs_iterations} shows the convergence of each method. The adjoint solution (via IPOPT \texttt{CasADi} autodifferentiation through \texttt{NeuralFoil}) is warm-started from the NACA-0012 airfoil; it represents a gradient-based local optimum reference value, which is the best achievable from a single starting point using exact surrogate gradients. ShapeEvolve converges the earliest in evaluation count, reaching near-adjoint performance at around $2,000$-$3,000$ evaluations with a relatively tight min/max band at the end. PSO and L-BFGS-B both reach competitive objective values at \textit{O}$(10,000)$ evaluations, which is significantly more than for ShapeEvolve. PSO converges steadily across all of its individual attempts. L-BFGS-B displays a wide min/max band; $3$ of the $40$ total attempts initialize in basins whose local minimum is infeasible, and so no gradient signal that points towards feasible designs. This $7.5$\% failure rate among the attempts serves as a limitation of gradient-based methods in this specific objective + environment setup. Bayesian optimization does eventually reach its converged values at \textit{O}$(5,000)$ evaluations, but settles in an unoptimal basin. It is worth emphasizing that Bayesian optimization is by far the most expensive in computational cost out of these methods.

Note that OpenEvolve and ShinkaEvolve were also tested on this task but were unable to find a feasible solution, as indicated in Figure~\ref{variance table} in Section \ref{main: cross-category}.

\begin{figure}
    \centering
    \includegraphics[width=0.7\linewidth]{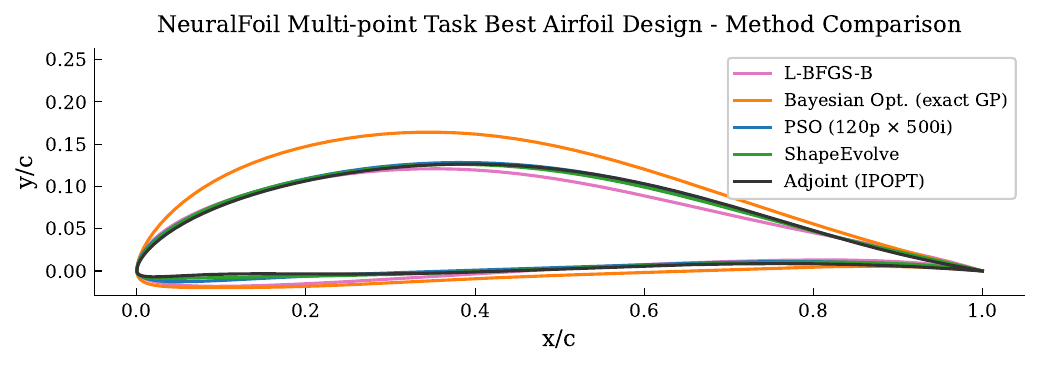}
    \caption{Airfoil design profiles ($y/c$ vs. $x/c$) for best-performing designs from all five methods, for the 2D airfoil multi-point drag minimization task.}
    \label{fig:Neuralfoil_yc_xc}
\end{figure}

Figure \ref{fig:Neuralfoil_yc_xc} shows the airfoil profiles for the best designs of all methods tested. The best designs of each except Bayesian optimization converge to the same adjoint-confirmed optimum solution, with the following geometric features:
\begin{itemize}
    \item thin airfoil height, relative to that of the 2D airfoil single-point maximum lift-to-drag task seen in figure \ref{fig:NeuralFoil_LD_airfoil_comparison}
    \item moderate camber (maximum camber value of $y/c \simeq 0.12$ near $x/c \simeq 0.35$)
\end{itemize}
The Bayesian optimization design is over-cambered and thicker overall, with a correspondingly larger $\overline{C_D}$ value. Unlike for the single-point maximum lift-to-drag task, the multi-point objective here yields two distinct reward basins; the Bayesian optimization found the suboptimal one.

\paragraph{Surrogate fidelity gap}

\begin{figure}
    \centering
    \includegraphics[width=0.95\linewidth]{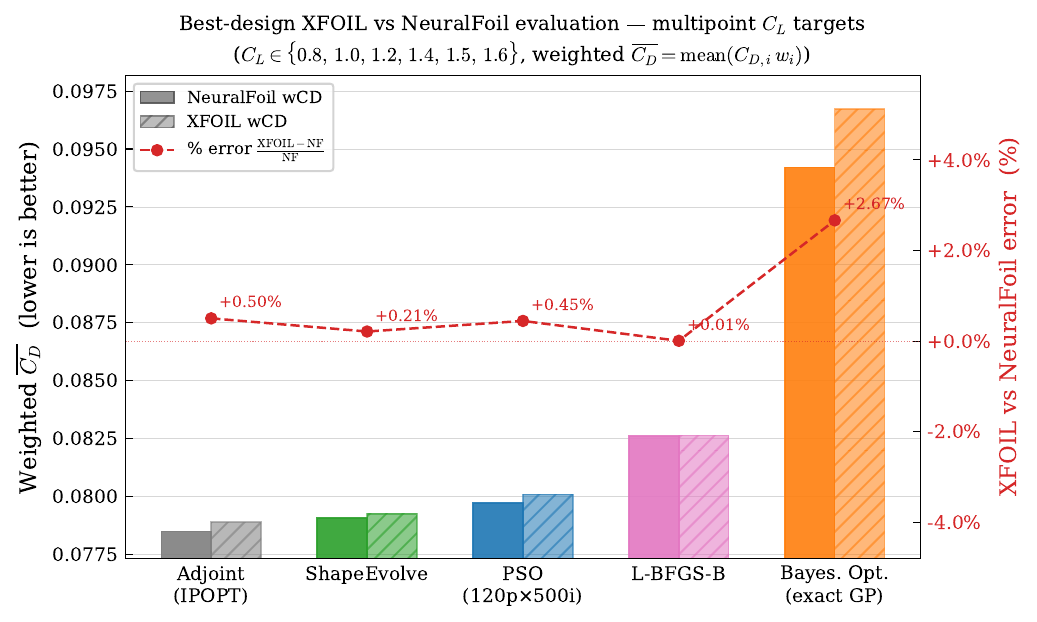}
    \caption{ \texttt{NeuralFoil} and \texttt{XFOIL} evaluations of best design for each method, for the 2D airfoil multi-point drag minimization task.}
\label{fig:NeuralFoil_multipoint_xfoil_vs_nf_fidelity}
\end{figure}

Figure \ref{fig:NeuralFoil_multipoint_xfoil_vs_nf_fidelity} serves to assess the \texttt{NeuralFoil} surrogate fidelity at the optimal designs found for each method. In other words, it answers whether these designs are truly high-performing (as validated by \texttt{XFOIL}), or are they exploiting the surrogate confidence just as was discussed in the single-point task. From the plot, it is observed that the \texttt{NeuralFoil} to \texttt{XFOIL} bias is negligible for the top four performing methods, with errors all within $0.50$\%; it is also observed that the \texttt{XFOIL} results preserve the rankings of the best $\overline{C_D}$ values among the designs, with the following ranking in drag minimization: 1) adjoint (IPOPT), 2) ShapeEvolve, 3) PSO, 4) L-BFGS-B, and finally, by far the worst-performing method here, 5) Bayesian optimization. Overall, for this multi-point drag minimization task, \texttt{NeuralFoil} is shown to be a highly accurate proxy to \texttt{XFOIL}.

\subsection{3D Transonic Swept-wing Design}

We use the 3D Transonic wing parametrization from \cite{yang2025superwingcomprehensivetransonicwing}. SupeWing is defined by a high parameter space, and as a result medians in shapebench are representative of time bound best designs rather than converged optima. Let 

\[
\mathbf{x}
=
\begin{pmatrix}
SA \\ AR \\ TR \\ \eta_k \\ \kappa_r \\
\Gamma_k \\ \Gamma_t \\
t_r \\ r_{t2} \\ r_{t3} \\ r_{t4} \\
r_{d1} \\ r_{d2} \\ r_{d4} \\
\theta_1 \\ \theta_2 \\ \theta_3 \\ \theta_4 \\
\mathbf{c}_u \\ \mathbf{c}_l
\end{pmatrix}
\in \mathbb{R}^{38}
\]

where

\[
25 \leq SA \leq 40\ \text{deg}, \qquad 8 \leq AR \leq 11
\]
\[
0.15 \leq TR \leq 0.40, \qquad 0.36 \leq \eta_k \leq 0.42
\]
\[
0.10 \leq \kappa_r \leq 1.10, \qquad 0.5 \leq \Gamma_k \leq 6.0\ \text{deg}
\]
\[
4.0 \leq \Gamma_t \leq 6.0\ \text{deg}, \qquad 0.14 \leq t_r \leq 0.17
\]
\[
0.60 \leq r_{t2} \leq 0.70, \qquad 0.90 \leq r_{t3} \leq 0.98
\]
\[
0.92 \leq r_{t4} \leq 1.00, \qquad 0.30 \leq r_{d1} \leq 0.80
\]
\[
0.50 \leq r_{d2} \leq 1.00, \qquad 0.00 \leq r_{d4} \leq 0.80
\]
\[
-4 \leq \theta_1,\,\theta_2 \leq -2\ \text{deg}, \qquad
-3 \leq \theta_3,\,\theta_4 \leq -1\ \text{deg}
\]
\[
-0.3 \leq c_{u,i} \leq 0.6,\ i=0,\dots,9 \qquad
-0.3 \leq c_{l,i} \leq 0.3,\ i=0,\dots,9
\]

\paragraph{Single-objective Task}

\paragraph{Operating point}
\[
\alpha_0 \in \{2.0^{\circ},\; 3.0^{\circ},\; 4.0^{\circ}\},
\qquad
M_0 \in \{0.70,\; 0.82,\; 0.85\}
\]

\paragraph{Objective}
Minimize aerodynamic drag subject to a lift coefficient floor:
\[
\min_{\mathbf{x}} \mathbf{f}(\mathbf{x})
=
\min_{\mathbf{x}}
\begin{pmatrix}
C_D(\mathbf{x};\,\alpha_0, M_0, Re_0) \\[6pt]
\lambda_{C_L}\,\max\!\bigl(0,\,C_L^* - C_L(\mathbf{x};\,\alpha_0, M_0, Re_0)\bigr)^2
\end{pmatrix}^{\!\top},
\]
where $C_L(\cdot)$ and $C_D(\cdot)$ denote the lift and drag coefficients,
$C_L^* = 0.45$ is the minimum required lift coefficient, and
$\lambda_{C_L} = 10.0$ is a quadratic penalty weight,
evaluated at $\alpha_0 = 3.0^{\circ}$, $M_0 = 0.82$.

\begin{figure}[h!]
    \centering
    \includegraphics[width=0.75\linewidth]{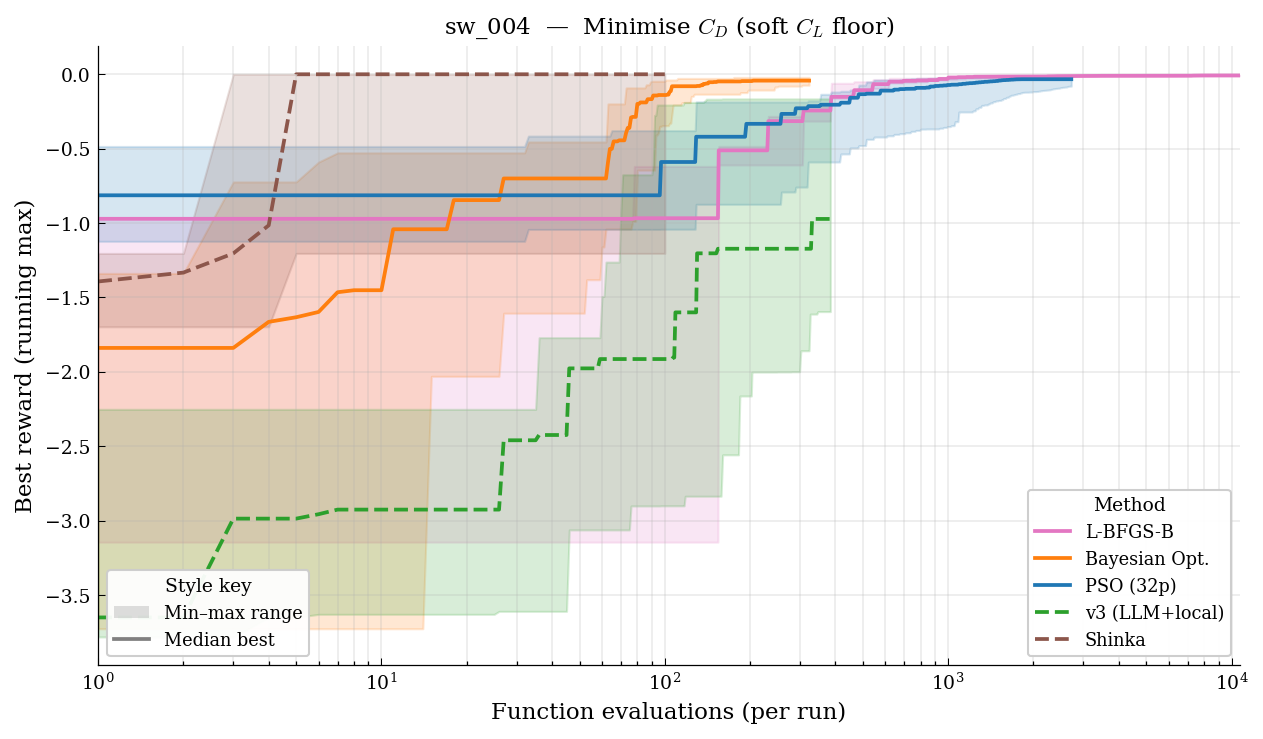}
    \caption{Single Objective \texttt{SuperWing} case: minimize $C_D$ with $C_L$ constraint}
    \label{fig:placeholder}
\end{figure}

\begin{figure}[h!]
    \centering
    \includegraphics[width=1\linewidth]{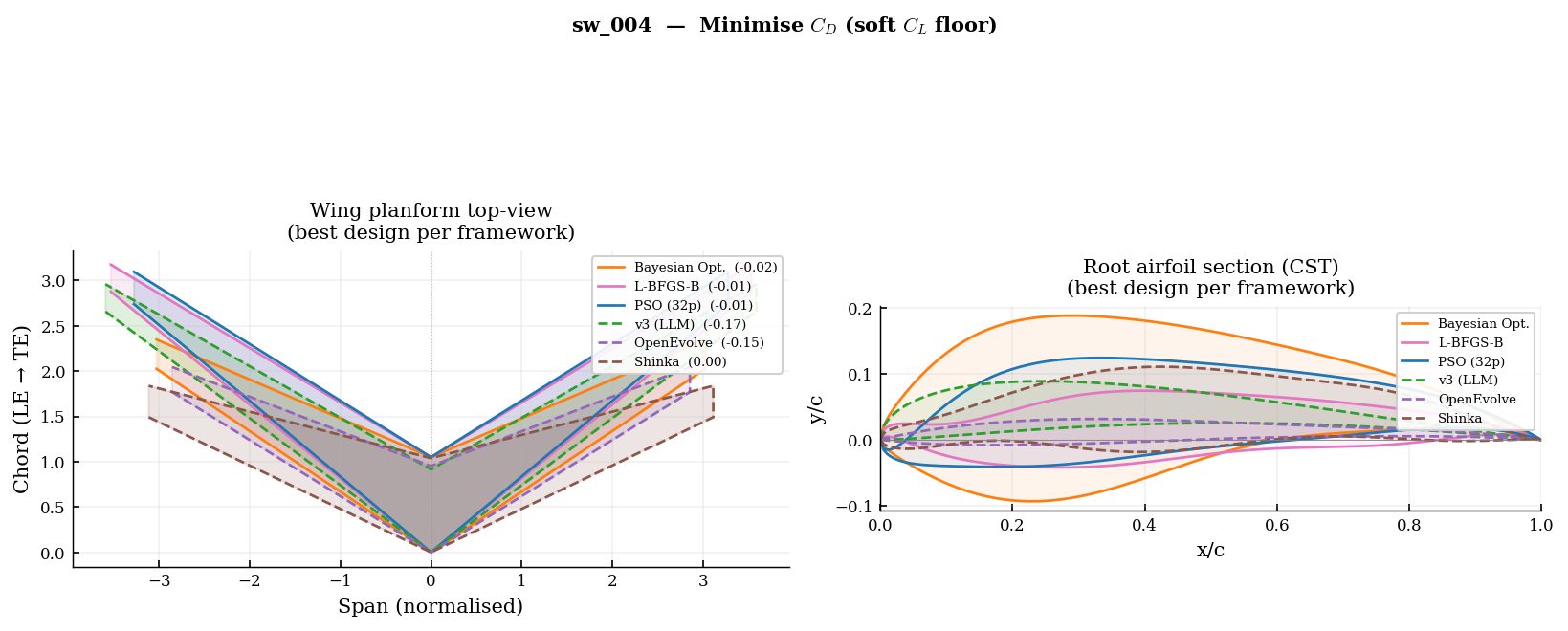}
    \caption{\texttt{SuperWing} best design overlay per method for single objective drag minimization}
    \label{fig:placeholder}
\end{figure}

\paragraph{Multi-point Task}

\paragraph{Operating points}
\[ M_0 \in \{0.75,\; 0.80,\; 0.86,\; 0.90\}
\]

\paragraph{Objective}
\[
\min_{\mathbf{x}} \mathbf{f}(\mathbf{x})
=
\min_{\mathbf{x}}
\frac{1}{K}\sum_{k \in K}
\begin{pmatrix}
-M_k\,\dfrac{C_L(\mathbf{x};\,\alpha_0^{(k)},M_k)}
            {C_D(\mathbf{x};\,\alpha_0^{(k)},M_k)} \\[10pt]
\lambda\,\bigl(M_k^2\,C_L - M_k\,C_L^*\bigr)^2
\end{pmatrix}^{\!\top},
\]
where $C_L^* = 0.55$, $\lambda = 1.0$, $K = \{0.75,\,0.80,\,0.86,\,0.90\}$, and where $\alpha_0^{(k)}$ is solved over
$\alpha \in [2^{\circ}, 12^{\circ}]$ to approximate $C_L^* = 0.55$.

\begin{figure}[h!]
    \centering
    \includegraphics[width=0.75\linewidth]{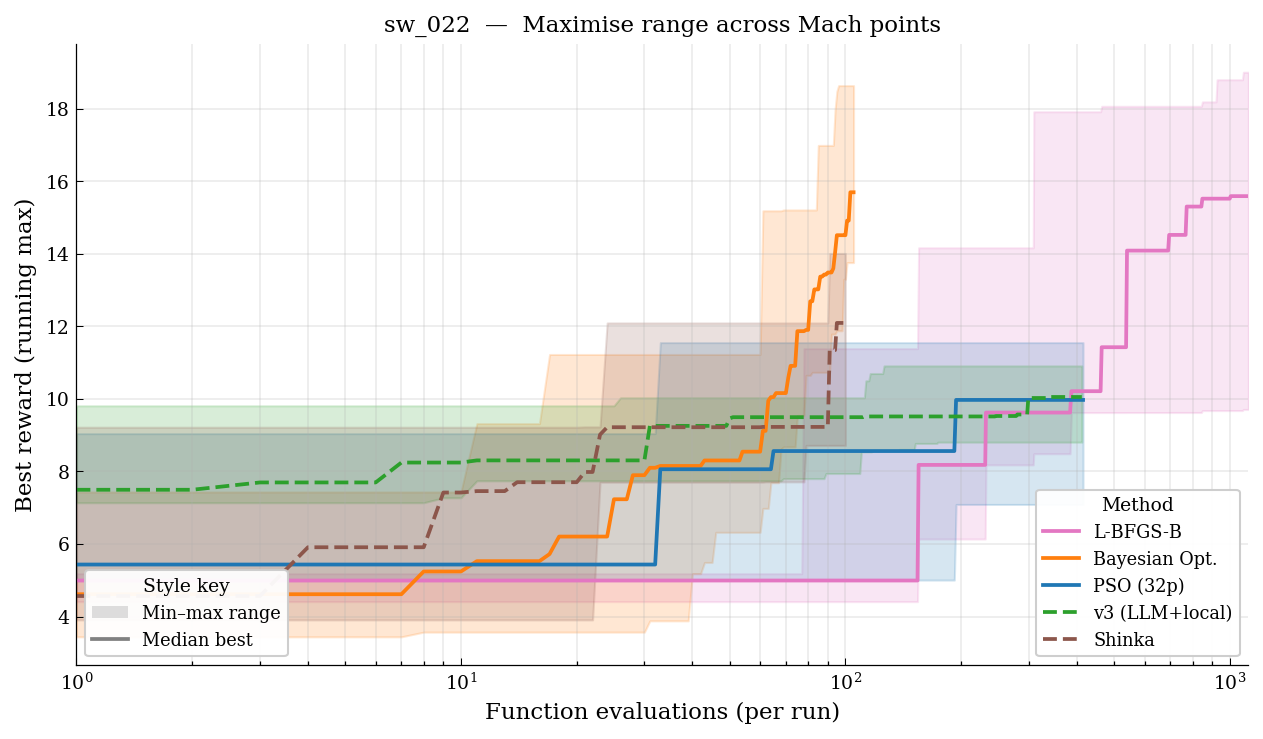}
    \caption{Optimizer trajectories for Multi point range maximization problem for \texttt{SuperWing}}
    \label{fig:placeholder}
\end{figure}

\begin{figure}[h!]
    \centering
    \includegraphics[width=1\linewidth]{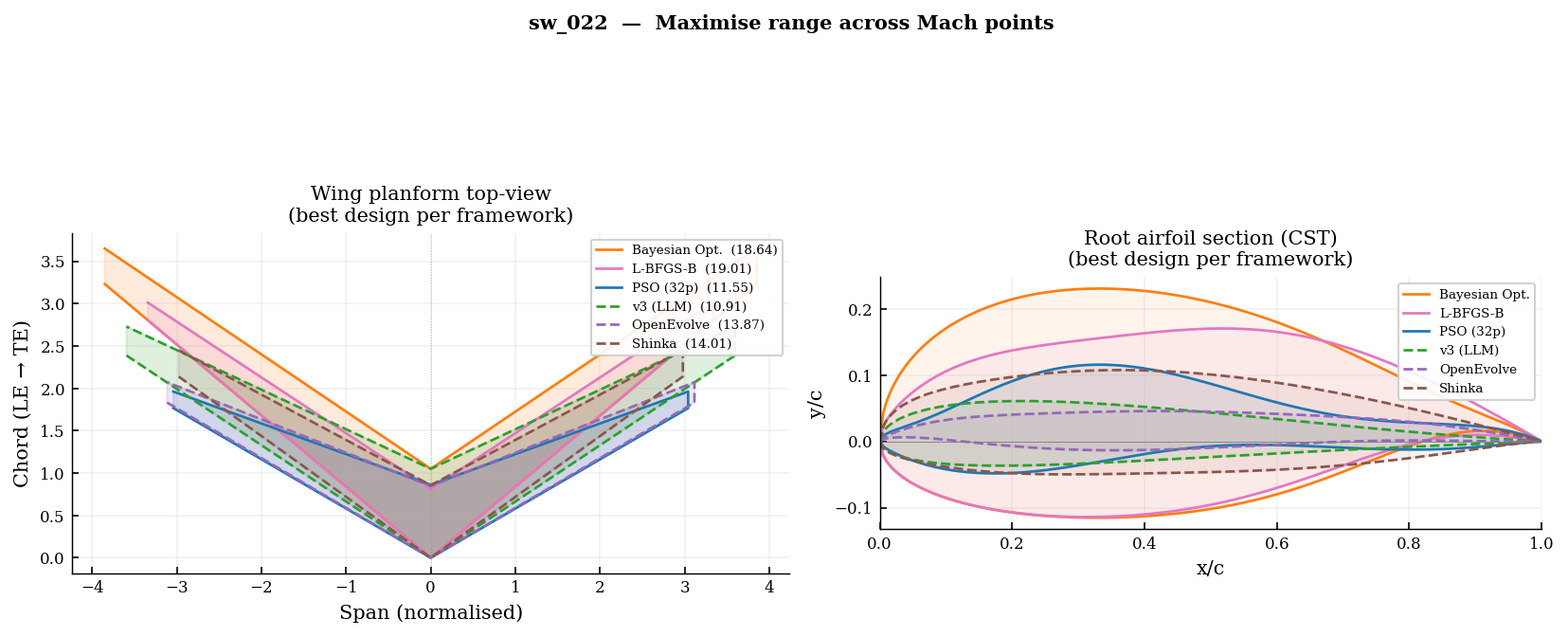}
    \caption{Optimum bets design per method for \texttt{SuperWing} multi-point range maximization problem}
    \label{fig:placeholder}
\end{figure}

\paragraph{Multi-point Task B}

\paragraph{Operating points}
\[
M_0 = 0.80,
\]
\[
C_L^* \in \{0.30,\; 0.40,\; 0.50,\; 0.60\},\]

\[\alpha_0^{(j)} \in [2^{\circ},\; 12^{\circ}]\ \text{(bisection)}
\]

\paragraph{Objective}
\[
\min_{\mathbf{x}} \mathbf{f}(\mathbf{x})
=
\min_{\mathbf{x}}
\sum_{j} w_j
\begin{pmatrix}
-M_0\,\dfrac{C_L(\mathbf{x};\,\alpha_0^{(j)},M_0)}
            {C_D(\mathbf{x};\,\alpha_0^{(j)},M_0)} \\[10pt]
\lambda\,\bigl(M_0^2\,C_L - M_0\,C_L^{*(j)}\bigr)^2
\end{pmatrix}^{\!\top},
\]
where weights $w_j = C_L^{*(j)}\!/\sum_j C_L^{*(j)}$,
$\lambda = 1.0$.

\begin{figure}[h!]
    \centering
    \includegraphics[width=0.75\linewidth]{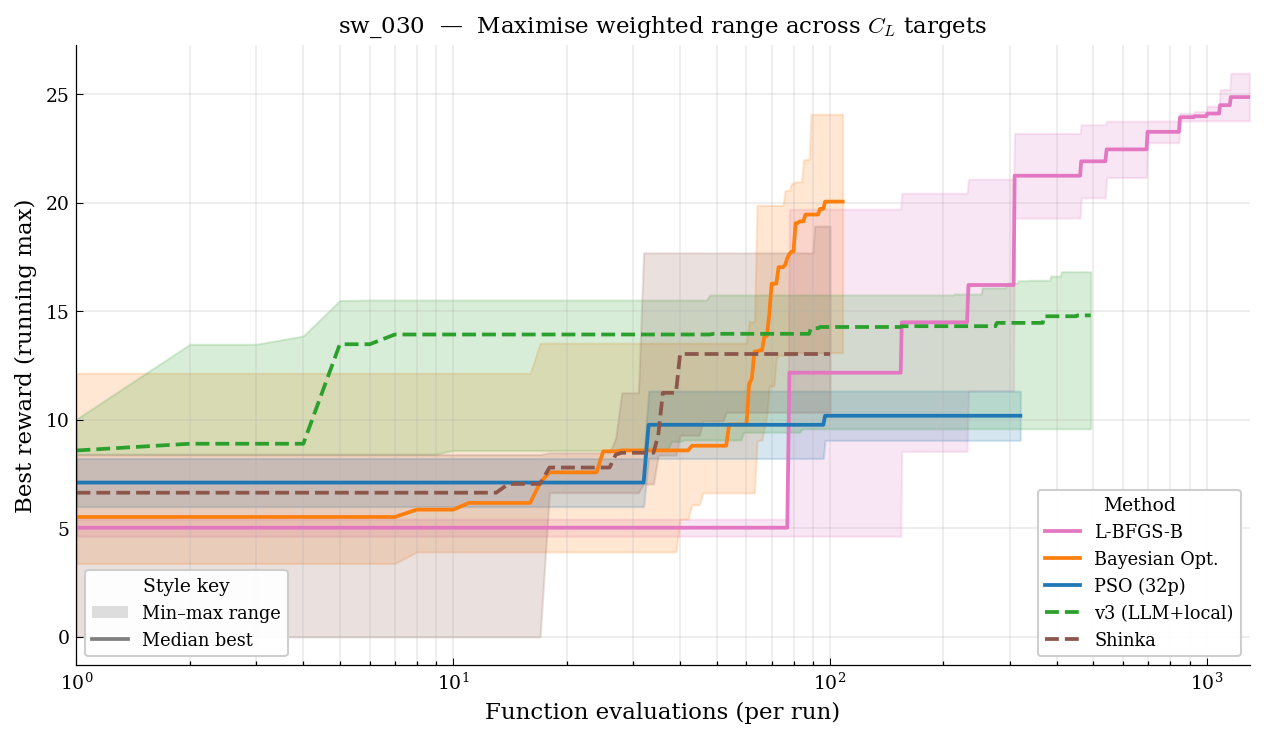}
    \caption{ShapeBench \texttt{SuperWing} problem 30: optimization methods and results}
    \label{fig:placeholder}
\end{figure}

\begin{figure}[h!]
    \centering
    \includegraphics[width=1\linewidth]{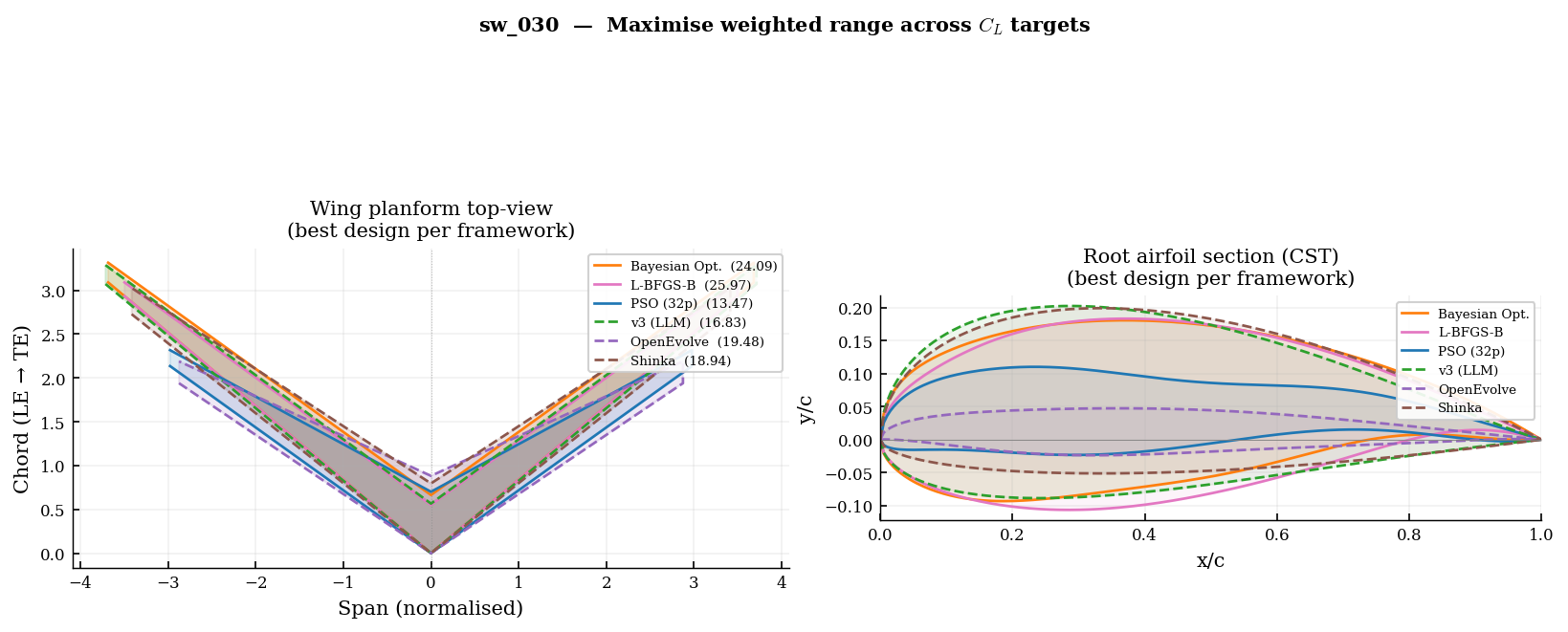}
    \caption{ShapeBench \texttt{SuperWing} problem 30: best design overlay}
    \label{fig:placeholder}
\end{figure}

\begin{figure}[h!]
    \centering
    \includegraphics[width=0.75\linewidth]{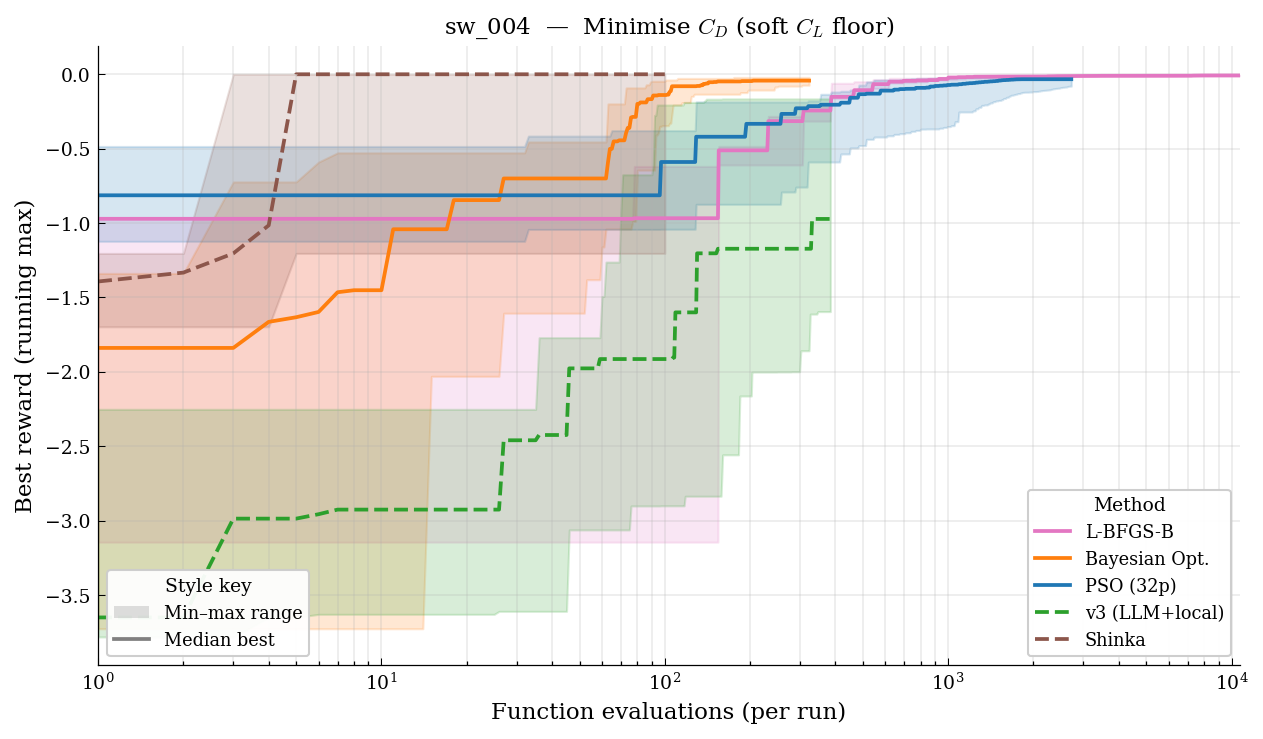}
    \caption{Enter Caption}
    \label{fig:placeholder}
\end{figure}

\newpage 

\subsection{3D Collaborative Combat Aircraft (CCA) Design} \label{app: cca tasks}

\paragraph{Design variables} We use the following 3D single-duct drone parametrization from nTop \cite{ntop2025}. 

Let
\[
\mathbf{x}
=
\begin{pmatrix}
\theta_d \\ B_w \\ \alpha_1 \\ \alpha_2 \\ p_w \\ x_r \\
l_i \\ n_{\mathrm{NACA}} \\ \theta_{ft} \\ \theta_{at} \\
h_{ta} \\ h_{ba} \\ b \\ \delta_r \\ c_r \\ c_t
\end{pmatrix}
\in \mathbb{R}^{16}
\]

where

\[
0.25 \leq \theta_d \leq 15\ \text{deg}, \qquad
25 \leq B_w \leq 1000\ \text{mm}
\]
\[
0 \leq \alpha_1 \leq 45\ \text{deg}, \qquad
0 \leq \alpha_2 \leq 10\ \text{deg}
\]
\[
0.22 \leq p_w \leq 0.51, \qquad
4500 \leq x_r \leq 7500\ \text{mm}
\]
\[
0.2 \leq l_i \leq 0.6, \qquad
n_{\mathrm{NACA}} \in \{1412,\,12,\,2408,\,4412\}
\]
\[
0 \leq \theta_{ft} \leq 10\ \text{deg}, \qquad
12 \leq \theta_{at} \leq 32.5\ \text{deg}
\]
\[
36 \leq h_{ta} \leq 220\ \text{mm}, \qquad
38 \leq h_{ba} \leq 208\ \text{mm}
\]
\[
6500 \leq b \leq 20000\ \text{mm}, \qquad
992 \leq \delta_r \leq 1770\ \text{mm}
\]
\[
1431 \leq c_r \leq 2700\ \text{mm}, \qquad
800 \leq c_t \leq 1200\ \text{mm}
\]

\subsubsection{Single-objective Task}

\paragraph{Operating point}
The aerodynamic coefficients are evaluated at a fixed operating point
\[
(\alpha_0, M_0, \mathrm{Re}_0),
\]
chosen apriori within
\[
\alpha_0 \in [0^\circ,3^\circ, 10^\circ],
\qquad
M_0 \in [0.35,0.5],
\qquad
\mathrm{Re}_0 \in [6.5\times 10^6,\,10^7].
\]

\paragraph{Objective}
Maximize the lift-to-drag ratio
\[
\max_{\mathbf{x}} \frac{C_L(x;\alpha_0,M_0,\mathrm{Re}_0)}
{C_D(x;\alpha_0,M_0,\mathrm{Re}_0)},
\]
where $C_L(\cdot)$ and $C_D(\cdot)$ denote the lift and drag coefficients.

\begin{figure}[h!]
    \centering
    \includegraphics[width=1\linewidth]{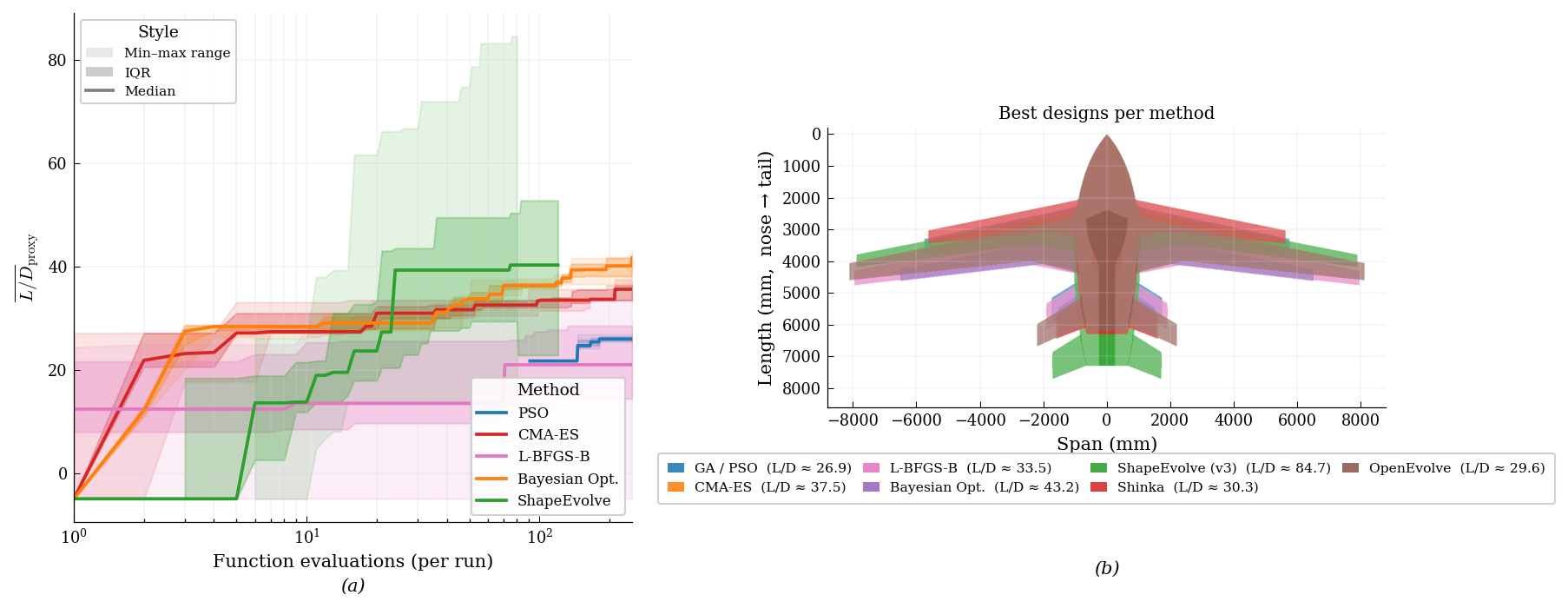}
    \caption{CCA evaluation for optimization of lift-to-drag ratio. (a) optimization method performance (b) best shape overlay per method}
    \label{fig:placeholder}
\end{figure}

\subsection{3D Car Design}\label{sec:3D_Car_Design}

\subsubsection{Single-objective Task}

\paragraph{Operating point}

The following fixed flow conditions are used
\[
  \rho_\infty = 1.25 \ \text{kg/m}^3,
\]
\[
  U_\infty = 40.0 \ \text{m/s},
\]
\[
  A_{\mathrm{ref}} = 2.37 \ \text{m}^2,
\]
\[
  q_\infty = \tfrac{1}{2}\,\rho_\infty\,U_\infty^2 = 1000 \ \text{Pa},
\]

\paragraph{Objective}

Minimize the drag coefficient $C_D$, 
\begin{equation}
    C_D = \frac{F_{\mathrm{drag,p}} + F_{\mathrm{drag,f}}}{q_\infty,A_{\mathrm{ref}}},
\end{equation}
where $F_{\mathrm{drag,p}} = \sum_i p_i\,A_i\,(n_x)_i$ is the pressure component of drag and $F_{\mathrm{drag,f}} = \sum_i \tau_{wx,i}\,A_i$ is the friction component of drag; each is summed over all surface cells of the mesh. $p$ is the (predicted) pressure, $A$ is the cell face area, $(n_x)_i$ is the $x$-component of the outward surface normal, and $\tau_{wx,i}$ is $x$-component of the the wall shear stress.

\paragraph{Design space bounds} (20 design parameters):

uniform scale factor $s_{\mathrm{car}}$:
\[
  0.80 \le s_{\mathrm{car}} \le 1.20,
\]
global linear shape parameters ($d_1 = \text{car width},\ d_2 = \text{car length},\ d_3 = \text{front bumper length}$):
\[
  -0.10 \le \Delta d_i \le 0.10 \ \text{m}
  \qquad \text{for } i = 1, 2, 3
\]
surface angles (ramp, trunk lid, diffuser, greenhouse, front hood, air intake angles) for $j = 1, \ldots, 6$:
\[
  -8.0^{\circ} \le \theta_j \le 8.0^{\circ}
  \qquad \text{for } j = 1, \ldots, 6
\]
$x$/$z$-displacements (windscreen, side mirrors, rear window, trunk lid for $k = 1, \ldots, 8$):
\[
  -0.05 \le \delta_k \le 0.05 \ \text{m}
  \qquad \text{for } k = 1, \ldots, 8,
\]
tire diameter:
\[
  -0.013 \le \Delta r_{\mathrm{tire}} \le 0.013 \ \text{m},
\]
tire width:
\[
  -0.015 \le \Delta w_{\mathrm{tire}} \le 0.015 \ \text{m}.
\]

The baseline design is the undeformed \texttt{DrivAerStar} geometry, which corresponds to $s_{\mathrm{car}} = 1.0$ and all offset/angle parameters set to $0$. No additional constraints beyond the box bounds are used in the problem definition.

\paragraph{Results}

\textit{Computational cost per $1{,}000$ evaluations per run} ($n$ = independent runs); $1$ core unless otherwise noted: 
\begin{itemize}
    \item L-BFGS-B ($n = 20$): $1.4$--$3.0$ CPU-hours 
    \item Bayesian Opt ($n = 20$): $13$--$28$ CPU-hours
    \item PSO ($n = 20$): $\approx 2.7$ CPU-hours
    \item ShapeEvolve ($n = 19$): $4.3$--$5.1$ CPU-hours
\end{itemize}

\begin{figure}
    \centering
    \includegraphics[width=0.95\linewidth]{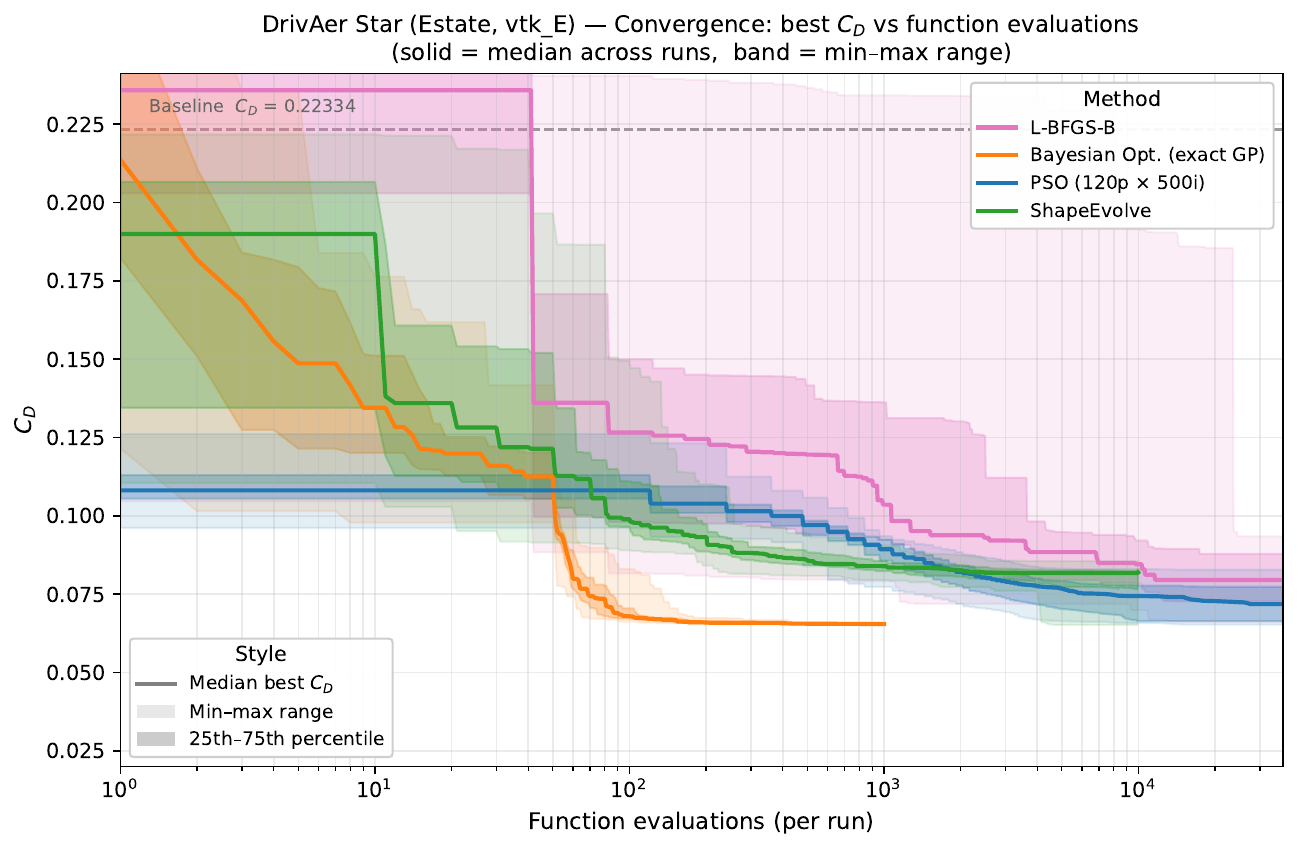}
    \caption{Convergence plot (objective vs. evaluations) for minimized $C_D$ task for 3D car design via \texttt{DrivAerStar}, configuration E.}
    \label{fig:DrivAer_Star_convergence_cd_vs_evals_vtk_E}
\end{figure}

Figure \ref{fig:DrivAer_Star_convergence_cd_vs_evals_vtk_E} shows the convergence plot for the drag minimization task on the \texttt{DrivAerStar} Estateback (configuration E) body. All four tested methods converge to a similar best $C_D \approx 0.065$.

Bayesian optimization converges with the least number of evaluations, at \textit{O}($1{,}000$); they are also nearly deterministic at the end, displaying a very narrow min/max spread. The other three methods (L-BFGS-B, PSO, and ShapeEvolve) all eventually reach the same best design, but each takes at least $5{,}000$ evaluations (with L-BFGS-B actually needing \textit{O}$(10{,}000)$). These methods also display significant min/max spread, especially PSO and ShapeEvolve, suggesting difficulty for the seeds in these methods to reliable find the optimal basin.

\paragraph{Discussion about surrogate exploitation and physical plausibility}

\begin{figure}
    \centering
    \includegraphics[width=0.95\linewidth]{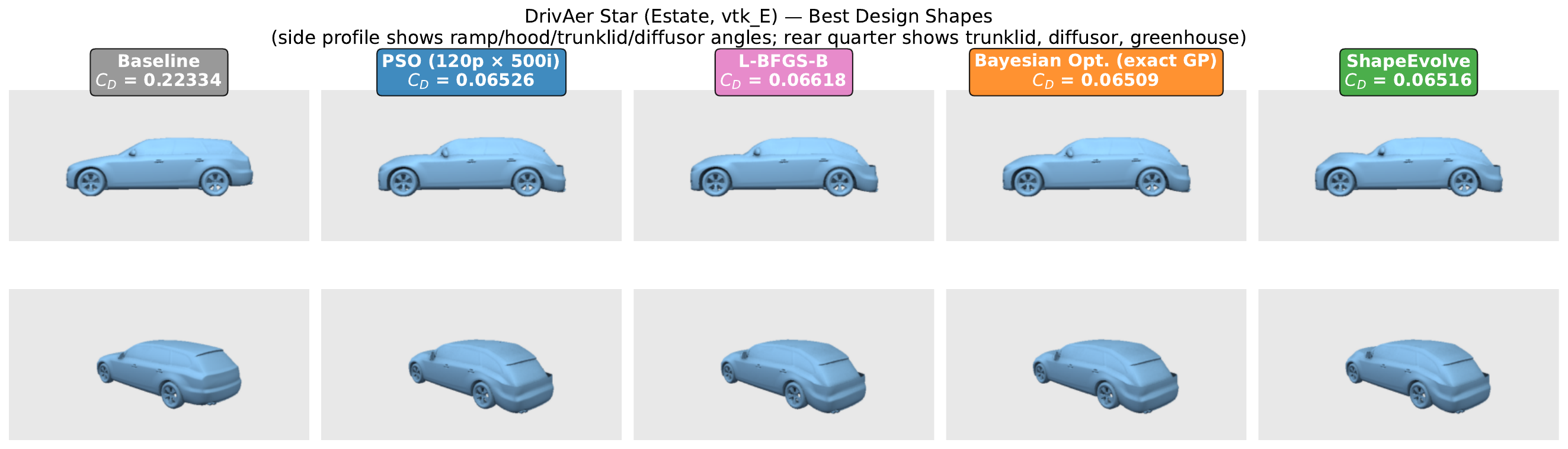}
    \caption{Best designs for the baseline and for each method (2D side views and 3D isometric views) for minimized $C_D$ task for 3D car design via \texttt{DrivAerStar}, configuration E.}
    \label{fig:DrivAer_Star_3d_panel_best_designs_vtk_E}
\end{figure}

The views of the best design found by each of the four methods in figure \ref{fig:DrivAer_Star_3d_panel_best_designs_vtk_E} confirm that each method converges to the same optimal design. However, a few observed features serve as indications of surrogate exploitation, suggesting that these results should be taken with caution. Of the $20$ total parameters in DrivAerStar, $13$ have been pushed to their bounds for the converged best designs across all methods, indicating that this convergence pattern is a landscape property of the surrogate rather than the quirk of any single optimization method. Geometrically, near-total collapse in the rear ground-clearance is observed, specifically caused by the \texttt{car\_size} and \texttt{diffusor\_angle} parameters being pushed to their bounds. 

It is possible to run CFD simulations of the final designs as ground-truth validation. However, no native CFD option exists within \texttt{DrivAerStar}, as the surrogate (\texttt{Transolver}-created using \texttt{DrivAerStar} data) is the sole evaluation source. One approach to investigate this limitation is performing tightened-bounds ablation studies to isolate the effect of surrogate exploitation, in conjunction with LLM-based physical plausibility diagnostics (which is a contributed tool within ShapeBench and discussed further in Appendix~\ref{app: diagnostic suite}). Results from both ablation and diagnostics consistently indicate high risk of surrogate exploitation across two levels of tightened parameter bounds, which suggests that this is an inherent property of the surrogate landscape rather than a boundary artifact of any particular run. Specifically, the pattern is consistent with structural surrogate exploitation -- boundary saturation increases (i.e., even more of the parameters are pushed to their boundary values) as bounds are tightened, consistent with the surrogate having no credible interior optimum.

\subsection{Mixed-variable Aircraft Design}

\paragraph{Central Reference Aircraft System (CERAS).} CERAS, based on an Airbus A320 aircraft, is considered as a constrained mixed-variable optimization problem. In the formulation reported in \citep{Saves_2022}, the problem involves $6$ continuous design variables, $2$ discrete variables, and $2$ categorical variables, for a total of $10$ design variables. Using the continuous-relaxation strategy adopted in the paper, the corresponding relaxed design space has dimension $12$.

Let the decision vector be written as
\[
w = (x,z,c),
\]
where $x \in \mathbb{R}^{6}$ collects the continuous variables, $z$ the two discrete variables, and $c$ the two categorical variables. The CERAS optimization problem can then be written as
\begin{equation}
\begin{aligned}
\min_{w=(x,z,c)} \quad & \mathrm{FuelMass}(w) \\
\text{s.t.} \quad & 0.05 < \mathrm{StaticMargin}(w) < 0.1.
\end{aligned}
\end{equation}

More explicitly, the continuous variables are bounded as
\[
x = \bigl(x_{\mathrm{MAC}},\, AR_{\mathrm{wing}},\, AR_{\mathrm{VT}},\, AR_{\mathrm{HT}},\, \lambda_{\mathrm{wing}},\, \Lambda_{\mathrm{wing}}\bigr),
\]
with
\[
x_{\mathrm{MAC}} \in [16,18] \ \text{m},
\qquad
AR_{\mathrm{wing}} \in [5,11],
\qquad
AR_{\mathrm{VT}} \in [1.5,6],
\]
\[
AR_{\mathrm{HT}} \in [1.5,6],
\qquad
\lambda_{\mathrm{wing}} \in [0,1],
\qquad
\Lambda_{\mathrm{wing}} \in [20,30]^\circ.
\]

The discrete variables are
\[
z = \bigl(h_{\mathrm{cruise}},\, N_{\mathrm{eng}}\bigr),
\]
with
\[
h_{\mathrm{cruise}} \in \{30\text{k},32\text{k},34\text{k},36\text{k}\}\ \text{ft},
\qquad
N_{\mathrm{eng}} \in \{2,3,4\}.
\]

The categorical variables are
\[
c = \bigl(c_{\mathrm{tail}},\, c_{\mathrm{engpos}}\bigr),
\]
with
\[
c_{\mathrm{tail}} \in \{\text{T-tail},\ \text{no T-tail}\},
\qquad
c_{\mathrm{engpos}} \in \{\text{front engines},\ \text{rear engines}\}.
\]

Table~\ref{tab:ceras_problem} summarizes the optimization variables and constraints.

\begin{table}[h!]
\centering
\caption{Definition of the CERAS optimization problem from \citep{Saves_2022}}
\label{tab:ceras_problem}
\begin{tabular}{llll}
\hline
\textbf{Function/variable} & \textbf{Nature} & \textbf{Quantity} & \textbf{Range} \\
\hline
Minimize Fuel mass & cont & 1 &  \\
\hline
x position of MAC & cont & 1 & $[16,18]$ m \\
Wing aspect ratio & cont & 1 & $[5,11]$ \\
Vertical tail aspect ratio & cont & 1 & $[1.5,6]$ \\
Horizontal tail aspect ratio & cont & 1 & $[1.5,6]$ \\
Wing taper aspect ratio & cont & 1 & $[0,1]$ \\
Angle for swept wing & cont & 1 & $[20,30]^\circ$ \\
\hline
Total continuous variables &  & 6 &  \\
\hline
Cruise altitude & discrete & 1 & $\{30\text{k},32\text{k},34\text{k},36\text{k}\}$ ft \\
Number of engines & discrete & 1 & $\{2,3,4\}$ \\
\hline
Total discrete variables &  & 2 &  \\
\hline
Tail geometry & cat & 2 levels & \{T-tail, no T-tail\} \\
Engine position & cat & 2 levels & \{front engines, rear engines\} \\
\hline
Total categorical variables &  & 2 &  \\
Total relaxed variables &  & 12 &  \\
\hline
$0.05 < \mathrm{StaticMargin} < 0.1$ & cont & 2 &  \\
\hline
Total constraints &  & 2 &  \\
\hline
\end{tabular}
\end{table}

\begin{figure}[h!]
    \centering
    \includegraphics[width=0.75\linewidth]{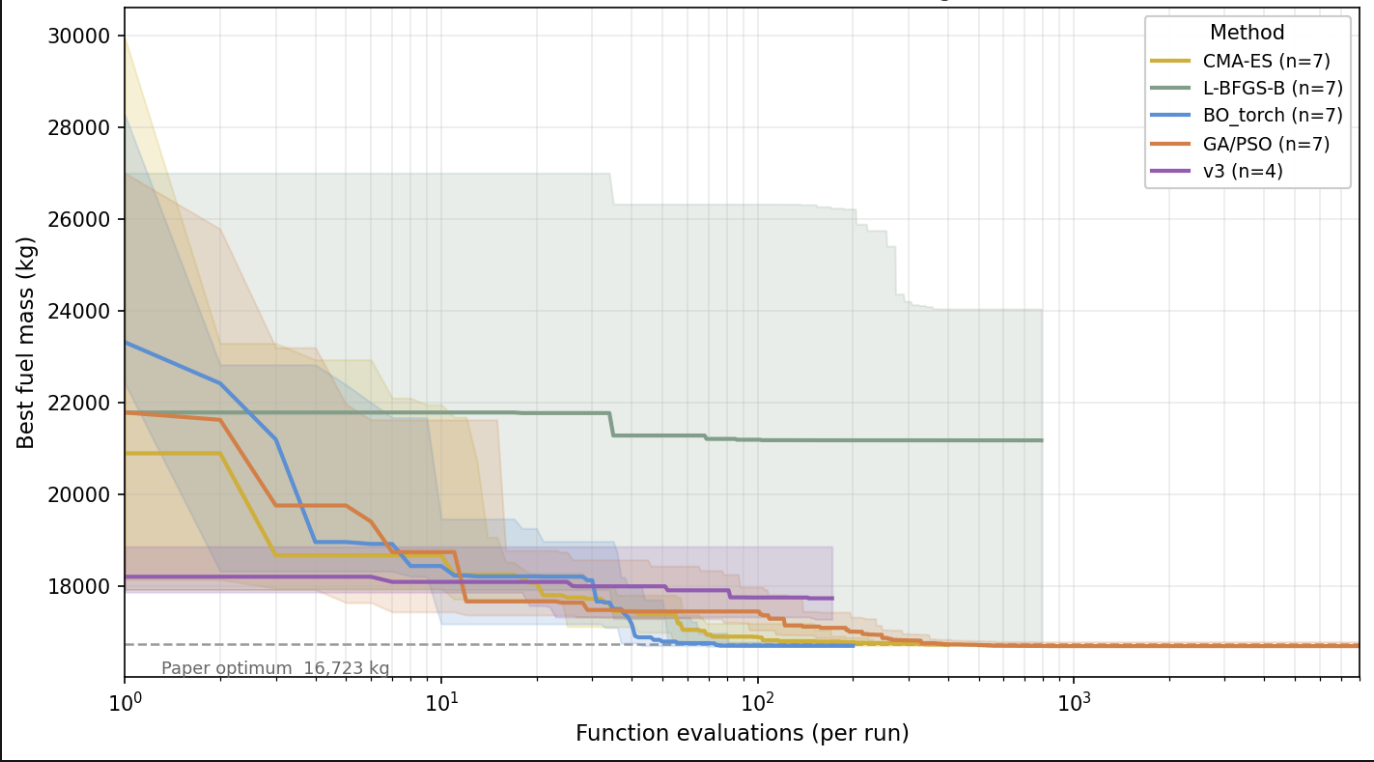}
    \caption{CERAS fuelmass objective optimization results}
    \label{fig:placeholder}
\end{figure}

\begin{figure}[h!]
    \centering
    \includegraphics[width=0.75\linewidth]{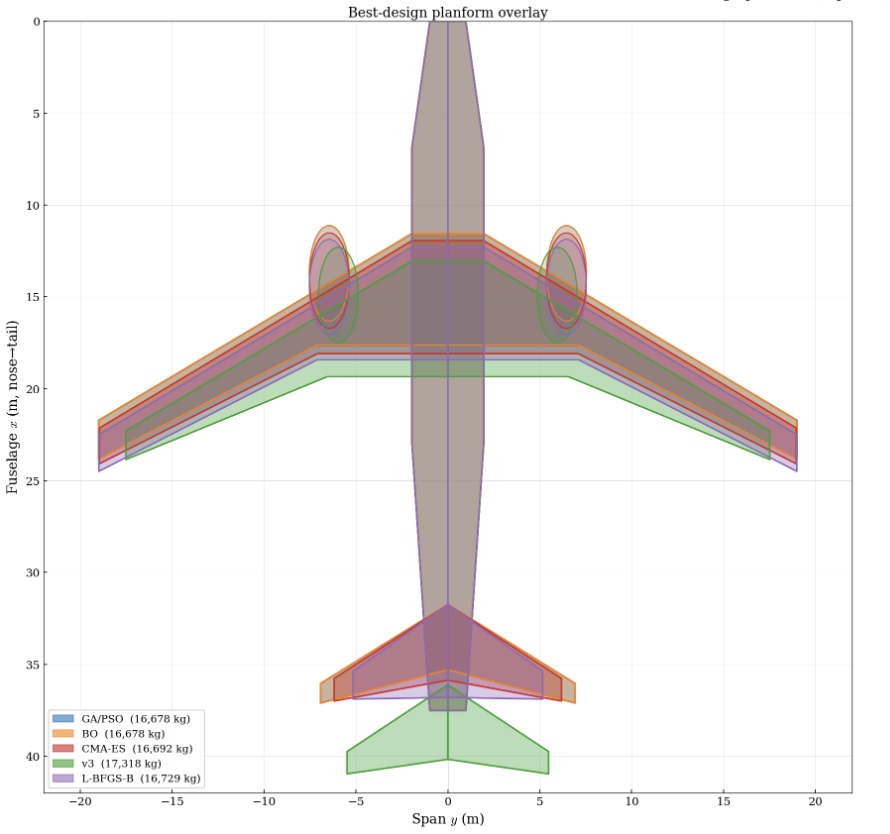}
    \caption{Best designs per method overlayed for CERAS fuelmass case}
    \label{fig:placeholder}
\end{figure}

\paragraph{Supersonic Transport Aircraft.}
The benchmark problem considers the conceptual design of a supersonic transport aircraft. The aircraft cruises at Mach $1.5$ and $500000$ feet, and is composed of a fuselage, a main wing, a tail section, and a possible canard. The optimization includes both continuous and discrete design variables in order to capture major configuration choices. In particular, three boolean discrete design variables are used to represent whether the aircraft has a cranked wing, a T-tail, and a canard. In addition, six continuous variables are used to define the aircraft geometry. As stated in the paper, the objective is to maximize the lift-to-drag ratio at cruise, while constraining the design variables to remain within their prescribed bounds.

Let the decision vector be written as
\[
w = (x,z),
\]
where $x \in \mathbb{R}^{6}$ collects the continuous variables and $z$ the three discrete boolean variables. The optimization problem can then be written as
\begin{equation}
\begin{aligned}
\min_{w=(x,z)} \quad & -\left(\frac{L}{D}\right)(w) \\
\text{s.t.} \quad & S^{lb}_{c,i} \leq S_{c,i} \leq S^{ub}_{c,i}, \\
& S_{d,i} \in \{S_d\}.
\end{aligned}
\end{equation}

More explicitly, the discrete variables are
\[
z = \bigl(
\texttt{cranked},\,
\texttt{T-tail},\,
\texttt{canard}
\bigr),
\].

The continuous variables are
\[
x = \bigl(
\Lambda_{\mathrm{in}},\,
\Lambda_{\mathrm{out}},\,
p_{\mathrm{canard}},\,
p_{\mathrm{wing}},\,
c_{\mathrm{break}},\,
s_{\mathrm{break}}
\bigr),
\]
with
\[
\Lambda_{\mathrm{in}} \in [10,50]^\circ,
\qquad
\Lambda_{\mathrm{out}} \in [10,70]^\circ,
\qquad
p_{\mathrm{canard}} \in [0.1,0.4]\%,
\]
\[
p_{\mathrm{wing}} \in [0.4,0.7]\%,
\qquad
c_{\mathrm{break}} \in [0.1,0.9]\%,
\qquad
s_{\mathrm{break}} \in [0.1,0.7]\%.
\]

Table~\ref{tab:supersonic_transport_problem} summarizes the optimization variables.

\begin{table}[h!]
\centering
\caption{Definition of the supersonic transport aircraft optimization problem.}
\label{tab:supersonic_transport_problem}
\begin{tabular}{llll}
\hline
\textbf{Function/variable} & \textbf{Nature} & \textbf{Quantity} & \textbf{Range} \\
\hline
Minimize $-\;L/D$ & cont & 1 & \\
\hline
Cranked wing  & discrete & 1 & $\{\text{cranked, not cranked}\}$ \\
 Tail geometry  & discrete & 1 & $\{\text{T-tail, no T-tail}\}$ \\
Canard & discrete & 1 & $\{\text{canard, no canard}\}$ \\
\hline
Total discrete variables &  & 3 & \\
\hline
inboard sweep angle $\Lambda_{\mathrm{in}}$ & continuous & 1 & $[10,50]^\circ$ \\
outboard sweep angle $\Lambda_{\mathrm{out}}$ & continuous & 1 & $[10,70]^\circ$ \\
canard location $p_{\mathrm{canard}}$ & continuous & 1 & $[0.1,0.4]\%$ \\
wing location $p_{\mathrm{wing}}$ & continuous & 1 & $[0.4,0.7]\%$ \\
break chord percent $c_{\mathrm{break}}$ & continuous & 1 & $[0.1,0.9]\%$ \\
break percent $s_{\mathrm{break}}$ & continuous & 1 & $[0.1,0.7]\%$ \\
\hline
Total continuous variables &  & 6 & \\
\hline
Variable lower/upper bound constraints & mixed & 9 & within prescribed bounds \\
Discrete membership constraints & discrete & 3 & boolean/discrete set membership \\
\hline
\end{tabular}
\end{table}

\section{Optimization Method Descriptions and Implementation Details} \label{Optimization_Method_Descriptions_and_Implementation_Details}

In this section, we provide more detailed descriptions of the optimization methods. Detailed implementations are available on ShapeBench's repository: \url{https://github.com/ShapeBench/ShapeBench}.

\subsection{Adjoint (IPOPT): Exact/analytical gradient-based}

When available from the problem setup and the code framework capabilities, exact analytical gradients can be computed using automatic differentiation. In PDE-constrained optimization, the classical adjoint method \citep{giles2000introduction, jameson1988aerodynamic} involves solving an adjoint (dual) PDE. Here, the automatic differentiation achieves the same result (i.e., gradient of the desired objective with respect to the design variables) at the cost of one forward pass and without an explicit dual solve.

The process starting from the design parameters and ending with the objective is represented symbolically using the CasADi\citep{andersson2018casadi} framework via AeroSandbox\citep{sharpe2021aerosandbox}, as depicted here:
\begin{equation*}
    \begin{aligned}
    \underbrace{\theta \in \mathbb{R}^{17}}_{\text{Kulfan params}} \xrightarrow{\text{CST(closed-form)}} \text{airfoil geometry} \xrightarrow{\text{\texttt{NeuralFoil}}} \text{coefficients } C_D, C_L, C_M \xrightarrow{} \text{objective}.
    \end{aligned}
\end{equation*}
Note: while technically \texttt{NeuralFoil} has $18$ free parameters, the trailing edge thickness is fixed at a value of $0$ for the adjoint problem; thus, the adjoint optimizes over $\mathbb{R}^{17}$.

Each step in the process is differentiable:
\begin{itemize}
    \item Airfoil geometric representation: The Kulfan class-shape transformation (CST)\citep{kulfan2008universal} provides analytical polynomial representations of the airfoil shape.
    \item \texttt{NeuralFoil}: The neural network is expressed using CasADi-compatible operations. By making use of \texttt{asb.numpy}, all operations are routed through CasADi's symbolic backend.
    \item Objective: When the objective is algebraic, such as with the multi-point weighted mean for \texttt{NeuralFoil} used in \ref{sec:2D_Airfoil_Design_Multi_point_task}, the overall problem is then differentiable end-to-end.
\end{itemize}
CasADi builds a computation graph of this process and then computes exact first-order (Jacobian) and second-order (Hessian) derivative information of the Lagrangian with automatic differentiation. The Jacobian and Hessian matrices information is then fed into IPOPT (Interior Point OPTimizer)\citep{wachter2006implementation}, which uses an interior-point barrier method to handle general nonlinear problems. Strict handling of constraints is applied, for equality constraints via Lagrange multipliers and for inequality constraints via the interior-point barrier method. With this combined CasADi + IPOPT framework, optimization problems can be iteratively solved, with $O(100)$ evaluations (each requiring one forward and one backward pass via automatic differentiation) for typical convergence and with quadratic convergence near the solution. For comparison, quasi-Newton methods typically require $O(1000)$ evaluations and can achieve at best superlinear convergence.

IPOPT options used are:
\begin{itemize}
    \item \texttt{"ipopt.mu\_strategy" =  "monotone"}: decreases the barrier parameter conservatively for robustness
    \item \texttt{"ipopt.start\_with\_resto"= "yes"}: begins with a feasibility restoration phase when the initial point is infeasible
\end{itemize}

\subsection{L-BFGS-B: Approximate gradient-based}

When analytical gradient values are unavailable for optimization, non-analytical numerical approximation methods can be utilized instead. The limited-memory Broyden-Fletcher-Goldfarb-Shanno with Bound constraints (L-BFGS-B) method \citep{zhu1997algorithm} is a memory-efficient, widely-used quasi-Newton method, described below.

For dimension $j \in \{1, \dots, d\}$, the gradients are approximated with 3-point finite differences:
\begin{equation*}
    \frac{\partial f}{\partial x_j} \approx \frac{f(\mathbf{x} + \epsilon \mathbf{e}_j) - f(\mathbf{x} - \epsilon \mathbf{e}_j)}{2\epsilon},  
  \quad \epsilon = 10^{-4},
\end{equation*}
where $\mathbf{e}_j$ is the unit vector in dimension $j$. 

The Hessian matrix $H$ is approximated in order to provide a gradient descent direction. Instead of the full $d \times d$ Hessian $H$, where $d$ is the dimension of the problem, the L-BFGS-B method gives
\begin{equation*}
    \mathbf{x}^{t+1} = \mathbf{x}^t - \alpha^t H_m^{-1} \nabla f(\mathbf{x}^t),
\end{equation*}
where $H_m^{-1}$ is the approximate inverse Hessian built from $m$ previous gradient/position pairs and $\alpha^t$ is the step size which is computed using a line search algorithm. $H_m^{-1}$ is calculated with the two-loop recursion method, and costs $O(md)$ per step; the scipy default value of $m = 10$ is applied. Bound constraints ($\text{lb}_j \leq x_j \leq \text{ub}_j$) are enforced via the gradient projection approach.

The solver terminates when one of the following convergence criteria is met:
\begin{itemize}
    \item $|\Delta f| / \max(|f|, 1) < 10^{-9}$ (function tolerance \texttt{ftol})
    \item $\|\nabla f\|_\infty < 10^{-6}$ (gradient tolerance \texttt{gtol})
    \item Number of iterations reaches \texttt{maxiter=200}
\end{itemize}

Because of the local nature of the method, multiple independent restarts are run per seed, with each having a different starting point sampled uniformly from $[\mathbf{lb}, \mathbf{ub}]$. Unless otherwise specified in the discussion for each case, $3$ restarts per seed are used. The best result across all restarts is kept.

\subsection{Particle Swarm Optimization (PSO): Population-based metaheuristic (swarm intelligence)}

Procedure for PSO \citep{kennedy1995particle}: Set the initial positions and velocities of the $N$ total particles in the swarm, then update the the positions and velocities for each particle for each iteration up until the $T$ total iterations. The initializations and the update steps are each defined below. The user sets the values of $N$ and $T$.

Indices:
\begin{itemize}
    \item $i \in {1, \dots,N}$: particle index 
    \item $j \in {1,\dots,d}$: parameter dimension index, where $d$ is the number of parameters/the dimension of the problem
    \item $t \in {0, \dots, T-1}$: iteration index
\end{itemize}

State variables:
\begin{itemize}
    \item $x_{i,j}^t$: particle position
    \item $v_{i,j}^t$: particle velocity
    \item $p_{i,j}^t$: personal best position of particle (continuously updated)
    \item $g_j^t$: global best position (of whichever particle has achieved the highest reward across the swarm)
\end{itemize}

Initial positions and velocity:
\begin{itemize}
    \item $v_{i,j}^{t = 0}$ is sampled from $U[-0.1,\Delta_j,\ +0.1,\Delta_j], \text{ with } \Delta_j = \text{ub}_j - \text{lb}_j$
    \item $x_{i,j}^{t = 0} = U[\mathbf{lb}, \mathbf{ub}]$; (given by a random uniform distribution from the feasible box)
    \item $p_{i,j}^{t = 0} = x_{i,j}^{t = 0}, \qquad g_j^{t = 0} = x_{i^{\text{best}},j}^{t = 0} \text{for arbitrary } i_{\text{best}} \in {1, \dots, N}$
\end{itemize}

Updates:
\begin{itemize}
    \item $v_{i,j}^{t+1} = \underbrace{w^t \cdot v_{i,j}^t}_{\text{inertia}} + \underbrace{c_1^t \cdot r_{1,i,j}^t \cdot (p_{i,j}^t - x_{i,j}^t)}_{\text{cognitive (personal best pull)}} + \underbrace{c_2^t \cdot r_{2,i,j}^t \cdot (g_j^t - x_{i,j}^t)}_{\text{social (global best pull)}}$
    \item $x_{i,j}^{t+1} = \text{clip}\left(x_{i,j}^t + v_{i,j}^{t+1} ,\ \mathbf{lb}_j,\ \mathbf{ub}_j\right)$
    \item After evaluating all $N$ particles at $t+1$:
    \begin{itemize}
        \item $p_{i,j}^{t+1} = \begin{cases}
        x_{i,j}^{t+1} & \text{if } f(x_i^{t+1}) > f_{\text{pbest},i} \\ p_{i,j}^t & \text{otherwise} 
        \end{cases}$
        \item $g_j^{t+1} = x_{i^,j}^{t+1}, \qquad i^{\text{best}} = \arg\max_i f(x_i^{t+1}), \quad \text{if } f(x_{i^{\text{best}}}^{t+1}) > f_{\text{gbest}}$
    \end{itemize}
\end{itemize}

Coefficient schedule (linear over $T$ iterations, same value applies to all $i$ and $j$):
\begin{itemize}
    \item inertia coefficient: $w^t = 0.8 - \frac{t}{T}(0.8 - 0.2)$
    \item cognitive coefficient (pulls each particle towards personal best): $c_1^t = 1.5 - \frac{t}{T}(1.5 - 0.5)$
    \item social coefficient (pulls each particle towards swarm's global best): $c_2^t = 0.2 + \frac{t}{T}(3.0 - 0.2)$
\end{itemize}

\subsection{Bayesian Optimization: Internal surrogate-based}

The implementation uses BoTorch \citep{balandat2020botorch} Bayesian Optimization (BO) \citep{frazier2018tutorial} framework from Meta, with exact Gaussian Process (GP).

Phase 1 (Initialization): 
\begin{itemize}
    \item Initial selection: Select $30$ random design samples, uniformly sampled from the normalized space $[0,1]^d$, where $d$ is the dimension pf the problem.
    \item Evaluation: De-normalize to the original parameter bounds, then evaluate with the simulator
\end{itemize}

Phase 2 (BO loop):
\begin{itemize}
    \item Fit the GP surrogate: standardize the observed rewards via $\tilde{f} = (f - \mu_f)/\sigma_f$, so that they have a mean of $0$ and a variance of $1$. This fits a SingleTaskGP (BoTorch's standard GP model for a single scalar output). The GP produces a posterior, whose mean is the surrogate prediction and whose variance gives a quantification of the uncertainty. The default Mat\'{e}rn-5/2 kernel, \begin{equation*}
        k(\mathbf{x}, \mathbf{x}') = \sigma_f^2\left(1 + \frac{\sqrt{5}\,r}{\ell} + \frac{5r^2}{3\ell^2}\right)\exp\left(-\frac{\sqrt{5}\,r}{\ell}\right), \quad r = \|\mathbf{x} - \mathbf{x}'\|,
    \end{equation*}
    is used, for length-scale $\ell$ and output variance $\sigma_f^2$. Maximizing the marginal log-likelihood using L-BFGS-B fits the GP, which thus sets both $\ell$ and $\sigma_f^2$.
    \item Optimize the acquisition function: Using the fitted GP, maximize the analytic Log Expected Improvement
    \begin{equation*}
        \text{LogEI}(\mathbf{x}) = \log \text{EI}(\mathbf{x}) = \log \mathbb{E}[\max(f(\mathbf{x}) - f^*, 0)],
    \end{equation*}
    where $f^*$ is the best observed value in standardized units. The expected improvement
    \begin{equation*}
        \text{EI}(\mathbf{x}) = (\mu_{\text{post}} - f^*)\,\Phi(Z) + \sigma_{\text{post}}\,\phi(Z), \quad Z = \frac{\mu_{\text{post}}(\mathbf{x}) - f^*}{\sigma_{\text{post}}(\mathbf{x})}
    \end{equation*}
    thus uses both the mean $\mu_{\text{post}}(\mathbf{x})$ and standard deviation $\sigma_{\text{post}}(\mathbf{x})$ of the posterior at $\mathbf{x}$ to balance exploitation (where the predicted mean yields high reward) and exploration (high uncertainty).

    The optimization uses gradient-based search with $10$ random restarts and $256$ quasi-random candidates for initialization, over the normalized space $[0,1]^d$. 
    
    This process selects the next candidate 
    $\mathbf{x}_{n+1}$. 
    \item Evaluate: De-normalize the selected $\mathbf{x}_{n+1}$ back to the original parameter bounds, then evaluate with the simulator.
    \item Update the observation set, then repeat the steps in phase 2 until the desired total number of evaluations is reached.
\end{itemize}

\subsection{Covariance Matrix Adaptation Evolution Strategy (CMA-ES): Population-based metaheuristic (evolution strategy)} \label{appendix: Evolution Strategy: Covariance Matrix Adaptation Evolution Strategy (CMA-ES)}

CMA-ES \citep{hansen2001completely} is a derivative-free, population-based algorithm that maintains and adapts a full multivariate Gaussian sampling distribution, $\mathcal{N}(\mathbf{m}, \sigma^2 \mathbf{C})$, over the search space for mean $\mathbf{m}$, global step size $\sigma$, and full covariance matrix $\mathbf{C}$. It learns the local covariance structure of the objective landscape unlike population-based methods with fixed-axis sampling (such as PSO); this makes CMA-ES invariant to scaling and rotation. CMA-ES serves as a standard reference solver for benchmarking of continuous black-box optimization \citep{hansen2021coco}.

The optimizer adapts $\mathbf{m}$, $\sigma$, and $\mathbf{C}$ of $\mathcal{N}(\mathbf{m}, \sigma^2 \mathbf{C})$ across generations. At each generation:
\begin{itemize}
    \item $\lambda$ offspring are sampled
    \item the best $\mu = \lfloor\lambda/2\rfloor$ offspring are then selected to update $\mathbf{m}$
    \item the covariance matrix is then updated via a rank-$\mu$ term (accumulating curvature from current selected steps) and a rank-1 term that reinforces the most recent successful search direction
\end{itemize}
The step size is adapted with cumulative step-size adaptation (CSA).

All hyperparameters used in the implementation here are set to the theory-derived defaults used in the \texttt{cma} package \citep{hansen2019pycma}. For dimension size $d = 10$ in BlendedNet, $\lambda = 10, \mu = 5$, and the initial step size $ = 0.3$ in normalized $[0,1]^d$ space.

\subsection{OpenEvolve \& ShinkaEvolve: LLM-driven} \label{appendix: LLM-driven: OpenEvolve ShinkaEvolve}

\paragraph{OpenEvolve:} OpenEvolve \cite{openevolve} is an open source evolutionary LLM coding agent with applications to algorithmic and performance optimization. The framework uses a MAP-Elites population model and applies diff-based edits as the primary mutation operator. OpenEvolve focuses on applications such as GPU Kernal optimization, symbolic regression, and function minimziation $\cite{openevolve}$

Our adaptation: As part of Shapebench we include OpenEvolve as one of three baseline LLM driven methods. We modify the functionality of OpenEvolve to extend to engineering shape optimization by modifying system prompt with domain specific guidance of aerodynamic parameter structure and enforcing parametric bounds. In addition, our adaptation extends OpenEvolve's LLM functionality from algorithmic code generation to parametric design sampling, which is compatible with shape bench and simulation based evaluations.

\paragraph{ShinkaEvolve:} ShinkaEvolve is an LLM guided evolutionary framework for algorithmic discovery {\cite{lange2025shinkaevolveopenendedsampleefficientprogram}}. The framework applies an island based hierarchical evolutionary sampling approach, code-novelty rejection, and LLM ensemble methods. 

Our adaptation: We adapt ShinkaEvolve \cite{lange2025shinkaevolveopenendedsampleefficientprogram} for ASO tasks in ShapeBench by evolving a sampling approach over aerodynamic and MDO evaluations. Whereas the original framework is focused on algorithm generation, we modify the approach to generate parametric engineering designs: we create an adapter for ShinkaEvolve that generates samples of design parameters, where each generation batch samples are evaluated through Shapebench and only the batch maximum design is stored in the ShinkaEvolve Framework. 

\subsection{ShapeEvolve: LLM-driven} \label{appendix: LLM-driven: ShapeEvolve}

ShapeEvolve is a ASO specific LLM guided evolutionary method we have included in Shape bench. The framework has two parts: (i) a designer agent that proposes a new center design based on hierarchical sampling from a population archive and simulation analysis (ii) an optimizer agent for generating sampling algorithms for the next $n$ designs. We include a population archive of both the best designs as well as a separate database for sampling algorithms generated over evaluations. Additionally, we include a reflective scratch pad specific to ASO and design generation. 

\end{document}